%% file: main_camera_ready.tex
\documentclass{article}

\usepackage[final]{neurips_2023}

\usepackage[utf8]{inputenc} 
\usepackage[T1]{fontenc}    
\usepackage{hyperref}       
\usepackage{url}            
\usepackage{booktabs}       
\usepackage{amsfonts}       
\usepackage{nicefrac}       
\usepackage{microtype}      
\usepackage{xcolor}         
\usepackage{amssymb}
\usepackage{amsmath}
\usepackage{times}
\usepackage{graphicx}
\usepackage{color}
\usepackage{multirow}
\usepackage{optidef}

\usepackage{tikz}
\usepackage{rotating}
\usepackage{bbm}
\usepackage{latexsym}
\usetikzlibrary{external}

\usepackage{tabularx}
\usepackage{amsfonts,amsthm,bm}
\usepackage{appendix}
\usepackage{mathtools}
\usepackage{algorithm,algorithmic}
\usepackage{subfig}
\usepackage{array}
\usepackage{wrapfig}
\usepackage{enumitem}
\usepackage{colortbl}
\usepackage{adjustbox}
\setitemize{itemsep=2pt,topsep=0pt,parsep=0pt,partopsep=0pt,leftmargin=*}

\PassOptionsToPackage{numbers,sort,compress}{natbib}
\input{math_commands.tex}

\usepackage[utf8]{inputenc} 
\usepackage[T1]{fontenc}    
\usepackage{hyperref}       
\usepackage{url}            
\usepackage{booktabs}       
\usepackage{amsfonts}       
\usepackage{nicefrac}       
\usepackage{microtype}      
\usepackage{xcolor}         
\usepackage{wrapfig}

\title{Correlative Information Maximization: A Biologically Plausible Approach to Supervised Deep Neural Networks without Weight Symmetry}

\author{Bariscan Bozkurt\textsuperscript{1,2, 3} \quad Cengiz Pehlevan\textsuperscript{4,5} \quad Alper T. Erdogan\textsuperscript{2, 3} \\
\textsuperscript{1} Gatsby Computational Neuroscience Unit, UCL, United Kingdom\\
\textsuperscript{2}KUIS AI Center, Koc University, Turkey \quad \textsuperscript{3}EEE Department, Koc University, Turkey \\
\textsuperscript{4}John A. Paulson School of Engineering \& Applied Sciences and Center for\\   Brain Science,
Harvard University, Cambridge, 02138 MA, USA\\
\textsuperscript{5}Kempner Institute for the Study of Natural and Artificial Intelligence\\
\texttt{\{bbozkurt15, alperdogan\}@ku.edu.tr}\quad 
\texttt{cpehlevan@seas.harvard.edu}
}

\begin{document}

\maketitle

\begin{abstract}
The backpropagation algorithm has experienced remarkable success in training large-scale artificial neural networks; however, its biological plausibility has been strongly criticized, and it remains an open question whether the brain employs supervised learning mechanisms akin to it. Here, we propose correlative information maximization between layer activations as an alternative normative approach to describe the signal propagation in biological neural networks in both forward and backward directions. This new framework addresses many concerns about the biological-plausibility of conventional artificial neural networks and the backpropagation algorithm. The coordinate descent-based optimization of the corresponding objective, combined with the mean square error loss function for fitting labeled supervision data, gives rise to a neural network structure that emulates a more biologically realistic network of multi-compartment pyramidal neurons with dendritic processing and lateral inhibitory neurons. Furthermore, our approach provides a natural resolution to the weight symmetry problem between forward and backward signal propagation paths, a significant critique against the plausibility of the conventional backpropagation algorithm. This is achieved by leveraging two alternative, yet equivalent forms of the correlative mutual information objective. These alternatives intrinsically lead to forward and backward prediction networks without weight symmetry issues, providing a compelling solution to this long-standing challenge. \looseness=-1
\end{abstract}

\section{Introduction}

How biological neural networks learn in a supervised manner has long been an open problem. The backpropagation algorithm \cite{rumelhart1986learning}, with its remarkable success in training large-scale artificial neural networks and intuitive structure, has inspired proposals for how biologically plausible neural networks can perform the necessary efficient credit-assignment for supervised learning in deep neural architectures \citep{whittington2019theories}.  Nonetheless, certain aspects of the backpropagation algorithm, combined with the oversimplified nature of artificial neurons, have been viewed as impediments to proposals rooted in this inspiration \cite{crick1989recent}. \looseness=-1


One of the primary critiques regarding the biological plausibility of the backpropagation algorithm is the existence of a parallel backward path for backpropagating error from the output towards the input, which uses the same synaptic weights as the forward path \citep{rumelhart1986learning, whittington2019theories, grossberg1987competitive}. Although such weight transport, or weight symmetry, is deemed highly unlikely based on experimental evidence \cite{crick1989recent, grossberg1987competitive}, some biologically plausible frameworks still exhibit this feature, which is justified by the symmetric structure of the Hebbian updates employed in these frameworks \cite{whittington2019theories, xie2003equivalence,scellier2017equilibrium}. 

The concerns about the simplicity of artificial neurons have been addressed by models which incorporate multi-compartment neuron models into networked architectures
 and ascribe important functions to dendritic processing in credit assignment \citep{larkum2013cellular, urbanczik2014learning, sacramento2018dendritic,golkar2022constrained}. This new perspective has enabled the development of neural networks with improved biological plausibility.

\begin{figure}[ht]
\begin{center}
\includegraphics[width=0.85\textwidth, trim={7.3cm, 7.4cm, 7.8cm, 2.1cm},clip]{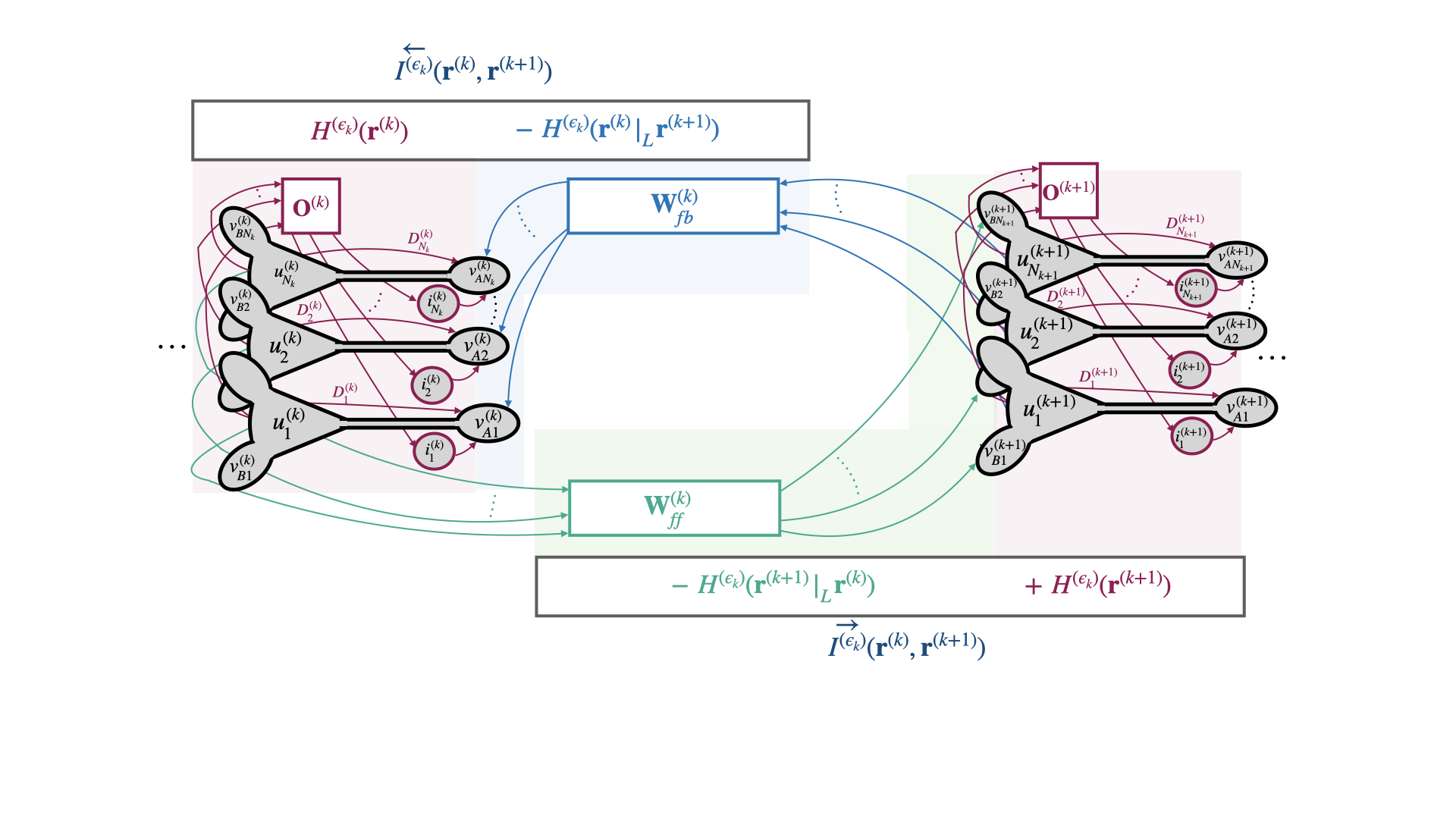}
\newline\caption{A segment of a correlative information maximization based neural network. Each layer consists of three-compartment pyramidal neurons with outputs $\rvr^{(k)}$ and membrane voltages  ($\mathbf{u}^{(k)}$-soma, $\mathbf{v}_{B}^{(k)}$-basal dendrites, $\mathbf{v}_{A}^{(k)}$-distal apical dendrites) and interneurons with outputs $\mathbf{i}^{(k)}$. The CMI  expression $ \overset{\rightarrow}{{I}^{(\epsilon_k)}}(\rvr^{(k)}, \rvr^{(k+1)})$ ($ \overset{\leftarrow}{{I}^{(\epsilon_k)}}(\rvr^{(k)}, \rvr^{(k+1)})$) defines forward (backward) prediction synapses $\mathbf{W}_{ff}^{(k)}$ ($\mathbf{W}_{fb}^{(k)}$), for minimizing $H^{(\epsilon_k)}(\mathbf{r}^{(k+1)}|_L\mathbf{r}^{(k)})$ ($H^{(\epsilon_k)}(\mathbf{r}^{(k)}|_L\mathbf{r}^{(k+1)})$) and the lateral connections $\mathbf{O}^{(k+1)}$ ($\mathbf{O}^{(k)}$) and autapses $\mathbf{D}^{(k+1)}$ ($\mathbf{D}^{(k)}$) connected to distal apical dendrites at the $k+1$-th ($k$-th) layer, for maximizing $H^{(\epsilon_k)}(\mathbf{r}^{(k+1)})$ ($H^{(\epsilon_k)}(\mathbf{r}^{(k)})$).} 
\label{fig:twolayercorinfomax}
\end{center}
\end{figure}

In this article, we propose the use of correlative information maximization (CorInfoMax) among consecutive layers of a neural network as a new supervised objective for biologically plausible models, which offers \looseness=-1
\begin{itemize}
\item \underline{a principled solution to the weight symmetry problem:}   
our proposed information theoretic criterion aims to maximize the linear dependence between the signals in two neighboring layers, naturally leading to the use of linear or affine transformations in between them. A key property of this approach is that employing two alternative expressions for the correlative mutual information (CMI) results in potentially \textit{asymmetric forward and backward prediction networks}, offering a natural solution to the weight transport problem. Consequently, predictive coding in both directions emerges as the inherent solution to the correlative information maximization principle, fostering signal transmission in both forward and top-down directions through asymmetrical connections.  While the CorInfoMax principle enhances information flow in both directions, the introduction of set membership constraints on the layer activations, such as non-negativity, through activation nonlinearities and lateral inhibitions, encourages compression of information and sparse representations \cite{bozkurt2023correlative}. \looseness=-1
\item \underline{a normative approach for deriving networks with multi-compartment neurons:} the gradient-based optimization of the CorInfoMax objective naturally leads to network models that employ multi-compartment pyramidal neuron models accompanied by interneurons as illustrated in Figure \ref{fig:twolayercorinfomax}.\looseness=-1
\end{itemize}
As derived and explained in detail in Section \ref{sec:DeepCorInfoMax}, the resulting networks incorporate lateral connections and auto-synapses (autapses) to increase the entropy of a layer, promoting utilization of all dimensions within the representation space of that layer. Meanwhile, asymmetric feedforward and feedback connections act as forward and backward predictors of layer activation signals, respectively, to reduce the conditional entropies between layers, targeting the elimination of redundancy.\looseness=-1

\subsection{Related work}
\subsubsection{Multi-compartmental  neuron model based biologically plausible approaches}
Experimentally grounded studies, such as \citep{larkum2013cellular, petreanu2009subcellular}, have 
been influential for considering a role for dendritic-processing in multi-compartmental neurons for learning and credit assignment \cite{richards2019dendritic}. 
Subsequent research has explored biologically plausible models with supervised learning functionality, such as the two-compartment neuron model by Urbanczik and Senn \citep{urbanczik2014learning} and the three-compartment pyramidal neuron model by Sacramento et al. \citep{sacramento2018dendritic}. Both models integrate non-Hebbian learning and spike-time dependent plasticity, while the latter includes SST interneurons \citep{urban2016somatostatin}. Similar frameworks have been proposed by \citep{guerguiev2017towards} and \citep{golkar2022constrained}, with the latter introducing a normative framework based on multi-compartmental  neuron structure, top-down feedback, lateral and feedforward connections, and Hebbian and non-Hebbian learning rules, emerging from the optimization of a prediction error objective with a whitening constraint on co-layer neurons.

In a similar vein to \citep{golkar2022constrained}, we propose an alternative normative framework  based on information maximization principle.  In this framework, the three-compartment structure and associated forward, top-down and lateral synaptic connections stem from the maximization of CMI between adjacent layers, without the imposition of any whitening constraint. \looseness=-1

 \subsubsection{Weight symmetry problem} 
A central concern regarding the biological plausibility of the backpropagation algorithm pertains to the weight symmetry issue: synaptic weights in the feedback path for error backpropagation are transposes of those used in the forward inference path \citep{whittington2019theories, crick1989recent, Grossberg1987CompetitiveLF}. The requirement of tied weights in backpropagation is questionable for physically distinct feedforward and feedback paths in biological systems, leading many researchers to focus on addressing the weight symmetry issue.\looseness=-1

Various strategies have been devised to address the weight symmetry issue. For example, the feedback alignment approach, which fixes randomly initialized feedback weights  and adapts feedforward weights, was offered as a plausible solution   \citep{lillicrap2016random}. Later Akrout et.al. \citep{Akrout2019Neurips} proposed its extension by updating feedback weights towards to the transpose of the feedforward weights. Along the similar lines, Amit  introduced antisymmetry through separate random initializations \citep{amit2019deep}. Liao et al. \citep{HowImportantWeightSymmetry} showed that the sign of the feedback weights (rather than their magnitude) affects the learning performance, and proposed the sign-symmetry algorithm. \looseness=-1

Intriguingly, this symmetric weight structure is also observed in biologically plausible frameworks such as predictive coding (PC) \citep{rao1999predictive, whittington2017approximation, song2020can}, equilibrium propagation (EP) \citep{EqProp, ScalingEP, laborieux2022EPholomorphic}, and similarity matching \citep{qin2021contrastive}. This phenomenon can be rationalized by the transpose symmetry of the Hebbian update with respect to inputs and outputs. The EP framework in \citep{ScalingEP} unties forward and backward connections inspired by \citep{scellier2018generalization, BP_without_weight_transport}, and only yields small performance degradation. A more recent approach by Golkar et al. \citep{golkar2022constrained} addresses this challenge by integrating two alternative forward prediction error loss function terms associated with the same network layer and leveraging presumed whitening constraints to eliminate shared feedback coefficients.

In existing predictive coding-based schemes such as \citep{rao1999predictive, whittington2017approximation, song2020can}, the loss function contains only forward prediction error terms. The feedback connection with symmetric weights, which backpropagates forward prediction error, emerges due to the gradient-based optimization of the PC loss. In contrast,  our framework's crucial contribution is the adoption of two alternative expressions for the correlative mutual information between consecutive network layers as the central normative approach. Utilizing these two alternatives naturally leads to both forward and backward prediction paths with asymmetric weights, promoting information flow in both feedforward and top-down directions. Unlike the work of \citep{golkar2022constrained}, our method circumvents the need for layer whitening constraints and additional forward prediction terms
to achieve asymmetric weights.

 \subsubsection{Correlative information maximization}
 Information maximization has been proposed as a governing or guiding principle in several machine learning and neuroscience frameworks for different tasks:  
(i) The propagation of information within a self-organized network as pioneered by Linsker \citep{linsker1988self}. (ii) Extracting hidden features or factors associated with observations by maximizing information between the input and its internal representation such as independent component analysis (ICA-InfoMax) approach by \citep{bell1995information}. In the neuroscience domain, the motivation has been to provide normative explanations to the behaviour of cortical activities evidenced by experimental work, such as orientation and visual stimuli length selectivity of primary visual cortex neurons \citep{HubelReceptive, bell1997independent}. The same idea has been recently extended in the machine learning field by the Deep Infomax approach where the goal is to transfer maximum information from the input of a deep network to its final layer, while satisfying prior distribution constraints on the output representations \citep{hjelm2018learning}. (iii) Matching representations corresponding to two alternative augmentations or modalities of the same input in the context of self-supervised learning \citep{becker1992self}. \looseness=-1

Correlative mutual information maximization has been recently proposed as an alternative for Shannon Mutual Information (SMI), due to its desirable properties \cite{erdogan2022icassp}:  (i) maximization of CMI is equivalent to maximizing \underline{linear} dependence, which may be more relevant than establishing arbitrary nonlinear dependence in certain applications \citep{ozsoy2022selfsupervised}, (ii) it is based only on the second order statistics, making it relatively easier to optimize. We additionally note that criteria based on correlation are intrinsically linked to local learning rules, leading to biologically plausible implementations, \cite{oja1982simplified,lipshutz2021biologically}.  Erdogan \citep{erdogan2022icassp} proposed the use of CorInfoMax for solving blind source separation (BSS) problem to retrieve potentially correlated components from their mixtures. Ozsoy et al. \citep{ozsoy2022selfsupervised} proposed maximizing the CMI between the representations of two different augmentations of the same input as a self-supervised learning approach. More recently, Bozkurt et al. \citep{bozkurt2023correlative} introduced an unsupervised framework to generate biologically plausible neural networks for the BSS problem with infinitely many domain selections using the CMI objective. 

In this article, we suggest employing the CorInfoMax principle for biologically plausible  supervised learning. The key difference compared to the unsupervised framework presented in \citep{bozkurt2023correlative} is the utilization of two alternative forms of mutual information. This leads to a bidirectional information flow that enables error backpropagation without encountering the weight symmetry issue.\looseness=-1

\section{Deep correlative information maximization}
\label{sec:DeepCorInfoMax}
\subsection{Network data model}
\label{sec:DCINetworkModel}
We assume a dataset with $\displaystyle L$ input data points $\rvx[t] \in \R^m, t = 1, \ldots, L$, and let $\displaystyle \rvy_T[t] \in \R^n$ be the corresponding labels. We consider a network with $\displaystyle P - 1$ hidden layers whose activities are denoted by $\displaystyle \rvr^{(k)}\in\R^{N_k}, k = 1, \ldots, P - 1$. For notational simplicity, we also denote  input and output of the network by $\displaystyle \rvr^{(0)}$ and $\displaystyle \rvr^{(P)}$, i.e., $\displaystyle \rvr^{(0)}[t] = \rvx[t]$ and $\displaystyle \rvr^{(P)}[t] = \hat{\rvy}[t]$. We consider polytopic constraints for the hidden and output layer activities, i.e., $\displaystyle \rvr^{(k)} \in \Pcal^{(k)}$, where $\displaystyle \Pcal^{(k)}$ is the presumed polytopic domain for the $k$-th layer \citep{bozkurt2023correlative, tatli2021tsp}. We note that the polytopic assumptions are plausible as the activations of neurons in practice are bounded. In particular, we will make the specific assumption that $\Pcal^{(k)}=\mathcal{B}_{\infty,+}=\{\rvr: \mathbf{0}\preccurlyeq \rvr \preccurlyeq \mathbf{1}\}$, i.e., (normalized) activations lie in a nonnegative unit-hypercube. Such nonnegativity constraints have been connected to disentangling behavior \citep{plumbley2003algorithms,pehlevan2017blind,whittington2023disentanglement}, however, we  consider extensions in the form of alternative polytopic sets corresponding to different feature priors \cite{bozkurt2023correlative} (see Appendix \ref{sec:polytopicrepresentations}).  More broadly, the corresponding label $\rvy_T$ can be, one-hot encoded label vectors for a classification problem, or discrete or continuous valued vectors for a regression problem.\looseness=-1


\subsection{Correlative information maximization based signal propagation}
\label{sec:DCIMCorInfoMax}
Our proposed CorInfoMax framework represents a principled approach where both the structure of the network and its internal dynamics as well as the learning rules governing adaptation of its parameters are not predetermined. Instead,  these elements emerge naturally from an explicit optimization process. As the optimization objective, we propose the maximization of {\it correlative mutual information} (see Appendix \ref{sec:prelim} between two consecutive network layers.  As derived in future sections, the proposed objective facilitates information flow—input-to-output and vice versa, while the presumed domains for the hidden and output layers inherently induce information compression and feature shaping. 

In Sections \ref{sec:criterion} and \ref{sec:criterionsamp}, we outline the correlative mutual information-based objective and its implementation based on samples, respectively. Section \ref{sec:criterionnn} demonstrates that the optimization of this objective through gradient ascent naturally results in recurrent neural networks with multi-compartment neurons. Finally, Section \ref{sec:learningDynamics} explains how the optimization of the same criterion leads to biologically plausible learning dynamics for the resulting network structure.

\subsubsection{Stochastic CorInfoMax based supervised criterion}
\label{sec:criterion}
We propose the total correlative mutual information among consecutive layers, augmented with the mean-square-error (MSE) training loss, as the stochastic objective to be maximized: 
\begin{eqnarray}
\label{eq:stobjective1}
J(\rvr^{(1)}, \ldots,\rvr^{(P)} )=\sum_{k=0}^{P-1} I^{(\epsilon_k)}(\rvr^{(k)}, \rvr^{(k+1)})-\frac{\beta}{2} E(\|\mathbf{y}_T-\mathbf{r}^{(P)}\|_2^2),
\end{eqnarray}
where, as defined in \citep{erdogan2022icassp, ozsoy2022selfsupervised} and in Appendix \ref{sec:prelim},
\begin{align}
    \overset{\rightarrow}{{I}^{(\epsilon_k)}}(\rvr^{(k)}, \rvr^{(k+1)}) &= \frac{1}{2} \log \det \left({\rmR}_{\rvr^{(k+1)}} + \epsilon_k \mI\right)- \frac{1}{2} \log \det \left({\mR}_{\overset{\rightarrow}{\rve}^{(k+1)}_*} + \epsilon_k \mI\right),\label{eq:sMIS1}
\end{align}
is the correlative mutual information between layers $\rvr^{(k)}$ an $\rvr^{(k+1)}$, ${\rmR}_{\rvr^{(k+1)}}=E(\rvr^{(k+1)}{\rvr^{(k+1)}}^T)$ is the autocorrelation matrix corresponding to the layer $\rvr^{(k+1)}$ activations,  and  ${{\rmR}}_{\overset{\rightarrow}{\rve}^{(k+1)}_*}
$ corresponds to the error autocorrelation matrix for the best linear regularized minimum MSE predictor of $\rvr^{(k+1)}$ from $\rvr^{(k)}$. Therefore, the mutual information objective in (\ref{eq:sMIS1}) makes a referral to the \textit{regularized {\bf forward} prediction problem} represented by the optimization
\begin{eqnarray}
    \underset{\mW_{ff}^{(k)}}{\text{minimize }} {E(\|\overset{\rightarrow}{\rve}^{(k+1)}\|_2^2)+\epsilon_k\|\mW_{ff}^{(k)}\|_F^2} \quad \text{   s.t.   } \quad {\overset{\rightarrow}{\rve}^{(k+1)}=\rvr^{(k+1)}-\mW_{ff}^{(k)}\rvr^{(k)}}, \label{eq:forwardpred}
\end{eqnarray}
and ${\rve}^{(k+1)}_*$ is the forward prediction error corresponding to the optimal forward predictor $\mW_{ff,*}^{(k)}$.

If we interpret the maximization of CMI in (\ref{eq:sMIS1}): the first term on the right side of (\ref{eq:sMIS1}) encourages the spread of $\rvr^{(k+1)}$ in its presumed domain 
$\Pcal^{(k+1)}$, while the second term incites the minimization of redundancy in $\rvr^{(k+1)}$  beyond its component predictable from $\rvr^{(k)}$.

An equal and alternative expression for the CMI can be written as (Appendix \ref{sec:prelim})
\begin{align}
            \overset{\leftarrow}{{I}^{(\epsilon_k)}}(\rvr^{(k)}, \rvr^{(k+1)}) &= \frac{1}{2} \log \det ({\rmR}_{\rvr^{(k)}}+ \epsilon_k \mI)- \frac{1}{2} \log \det \left({\mR}_{\overset{\leftarrow}{\rve}^{(k)}_*} + \epsilon_k \mI\right),\label{eq:sMIS2}
\end{align}
where ${{\mR}}_{\overset{\leftarrow}{\rve}^{(k)}_*}$ corresponds to the error autocorrelation matrix for the best linear regularized minimum MSE predictor of $\rvr^{(k)}$ from $\rvr^{(k+1)}$. The corresponding \textit{regularized {\bf backward} prediction problem} is defined by the optimization 
\begin{eqnarray}
    \underset{\mW_{fb}^{(k)}}{\text{minimize }} {E(\|\overset{\leftarrow}{\rve}^{(k)}\|_2^2)+\epsilon_k\|\mW_{fb}^{(k)}\|_F^2} \quad \text{   s.t.   } \quad \overset{\leftarrow}{\rve}^{(k)}=\rvr^{(k)}-\mW_{fb}^{(k)}\rvr^{(k+1)}. \label{eq:backwpred}
\end{eqnarray}
We observe that the two alternative yet equivalent representations of the correlative mutual information between layers $\rvr^{(k)}$ and $\rvr^{(k+1)}$ in (\ref{eq:sMIS1}) and (\ref{eq:sMIS2}) are intrinsically linked to the forward and backward prediction problems between these layers, which are represented by the optimizations in (\ref{eq:forwardpred}) and (\ref{eq:backwpred}), respectively. As we will demonstrate later, the existence of these two alternative forms for the CMI plays a crucial role in deriving a neural network architecture that overcomes the weight symmetry issue. \looseness=-1

\subsubsection{Sample-based supervised CorInfoMax criterion}
\label{sec:criterionsamp}
Our aim is to construct a biologically plausible neural network that optimizes the total CMI, equation (\ref{eq:stobjective1}), in an adaptive manner.
Here, we obtain a sample-based version of (\ref{eq:stobjective1}) as a step towards that goal.

We first define the exponentially-weighted sample auto and cross-correlation matrices as follows:
\begin{align}
\displaystyle
\hat{\rmR}_{\rvr^{(k)}}[t] = \frac{1 - \lambda_\rvr}{1 - \lambda_\rvr^t}\sum_{i=1}^{t} \lambda_\rvr^{t-i}  \rvr^{(k)}[i] {\rvr^{(k)}[i]}^T, \hspace{0.01in} \hat{\rmR}_{\rvr^{(k)}\rvr^{(k+1)}}[t] = \frac{1 - \lambda_\rvr}{1 - \lambda_\rvr^t}\sum_{i=1}^{t} \lambda_\rvr^{t-i}  \rvr^{(k)}[i] {\rvr^{(k+1)}[i]}^T, \label{eq:weightedRr}
\end{align}
for $k=0, \ldots, P$, respectively, where $0 \ll \lambda_\rvr<1$ is the forgetting factor. Next, we define two equivalent forms of the sample-based CMI, $\hat{I}^{(\epsilon)}(\rvr^{(k)}, \rvr^{(k+1)})[t]$:
\begin{align}
    \overset{\rightarrow}{\hat{I}^{(\epsilon_k)}}(\rvr^{(k)}, \rvr^{(k+1)})[t] &= \frac{1}{2} \log \det (\hat{\rmR}_{\rvr^{(k+1)}}[t] + \epsilon_k \mI)- \frac{1}{2} \log \det ({\hat{\mR}}_{\overset{\rightarrow}{\rve}^{(k+1)}_*}[t] + \epsilon_k \mI),\label{eq:MIS1}\\
        \overset{\leftarrow}{\hat{I}^{(\epsilon_k)}}(\rvr^{(k)}, \rvr^{(k+1)})[t] &= \frac{1}{2} \log \det (\hat{\rmR}_{\rvr^{(k)}}[t] + \epsilon_k \mI)- \frac{1}{2} \log \det ({\hat{\mR}}_{\overset{\leftarrow}{\rve}^{(k)}_*}[t] + \epsilon_k \mI),\label{eq:MIS2}
\end{align}
where 
${\hat{\mR}}_{\overset{\rightarrow}{\rve}^{(k+1)}_*}[t]$
is the exponentially-weighted sample autocorrelation matrix for the forward prediction error at level-$(k+1)$, $\overset{\rightarrow}{\rve}^{(k+1)_*}[t]$, corresponding to the best linear exponentially-weighted regularized least squares predictor of $\rvr^{(k+1)}[t]$ from the lower level activations $\rvr^{(k)}[t]$. Similarly,
${\hat{\mR}}_{\overset{\leftarrow}{\rve}^{(k)}}[t]$
 is the exponentially-weighted autocorrelation matrix for the backward prediction error at level-$(k)$, $\overset{\leftarrow}{\rve}^{(k)}[t]$, corresponding to the best linear exponentially-weighted regularized least squares predictor of $\rvr^{(k)}[t]$ from the higher level activations $\rvr^{(k+1)}[t]$.\looseness=-1

The sample-based CorInfoMax optimization can be written as:
\begin{maxi!}[l]<b>
{\rvr^{(k)}[t], k = 0, \ldots, P}{\sum_{k = 0}^{P-1} {\hat{I}}^{(\epsilon_k)}(\rvr^{(k)}, \rvr^{(k+1)})[t]-\frac{\beta}{2}\|\rvy_T[t]-\rvr^{(P)}[t]\|_2^2=\hat{J}(\rvr^{(1)}, \ldots, \rvr^{(P)})[t]\label{eq:ldmiobjective}}{\label{eq:UnsupervisedLDMIobjective}}{}
\addConstraint{\hspace{-1.99cm}\rvr^{(k)}[t] \in \Pcal^{(k)}, k=1, \dots, P}
\addConstraint{\rvr^{(0)}[t] = \rvx[t],}
{\label{eq:ldmioptimization}}{}
\end{maxi!}

As outlined in Appendix \ref{sec:taylorappr}, we can employ Taylor series linearization to approximate the $\log\det$ terms associated with forward and backward prediction errors in  (\ref{eq:sMIS1}) and (\ref{eq:sMIS2}) in the form
\begin{align}
    \log \det \left({\hat{\mR}}_{\overset{\rightarrow}{\rve}^{(k+1)}}[t] + \epsilon_k \mI\right)& \nonumber \\
    &\hspace*{-1.7in}\approx\frac{1}{\epsilon_k}\sum_{i=1}^t\lambda_\rvr^{t-i}\|\rvr^{(k+1)}[i]-\mathbf{W}_{ff,*}^{(k)}[t]\rvr^{(k)}[i]\|_2^2+\epsilon_k\|\mW_{ff,*}^{(k)}[t]\|_F^2+N_{k+1}\log(\epsilon_k) \label{eq:TaylorApp1} \\ 
    \log \det \left({\hat{\mR}}_{\overset{\leftarrow}{\rve}^{(k)}}[t] + \epsilon_k \mI\right)& \nonumber\\
    &\hspace*{-1.7in}\approx\frac{1}{\epsilon_k}\sum_{i=1}^t\lambda_\rvr^{t-i}\|\rvr^{(k)}[i]-\mathbf{W}_{fb,*}^{(k)}[t]\rvr^{(k+1)}[i]\|_2^2+\epsilon_k\|\mW_{fb,*}^{(k)}[t]\|_F^2+N_k\log(\epsilon_k), \label{eq:TaylorApp2}
\end{align}
where $\mathbf{W}_{ff,*}^{(k)}[t]$ is the optimal linear regularized weighted least squares forward predictor coefficients in predicting $\rvr^{(k+1)}[i]$ from $\rvr^{(k)}[i]$  for $i=1, \ldots, t$, and  $\mathbf{W}_{fb,*}^{(k)}[t]$ is the optimal linear regularized weighted least squares backward predictor coefficients in predicting $\rvr^{(k)}[i]$ from $\rvr^{(k+1)}[i]$  for $i=1, \ldots, t$. Consequently,  the optimal choices of forward and backward predictor coefficients are coupled with the optimal choices of layer activations.

In the online optimization process, we initially relax the requirement on the optimality of predictors and start with random predictor coefficient selections. During the learning process, we apply a coordinate ascent-based procedure on activation signals and predictor coefficients. Specifically,  at time  step-$t$, we consider two phases:
\begin{itemize}
\item [ 1.] First, we optimize with respect to the activations $\{\rvr^{(k)}[t],  k=1, \ldots, P\}$, 
 where we assume predictor coefficients to be fixed.  This phase yields network structure and output  dynamics,
 \item[2.] Next, we update the forward and backward predictor coefficients $\mW_{ff}^{(k)}$ and $\mW_{fb}^{(k)}$, for $k=1, \ldots, P$, to reduce the corresponding forward and backward prediction errors, respectively. This phase provides update expressions to be utilized in learning dynamics.
 \end{itemize}
 As the algorithm iterations progress, the predictor coefficients converge to the vicinity of their optimal values.\looseness=-1

For the first phase of the online optimization, we employ a projected gradient ascent-based approach for activations: for $k=1, \ldots, P-1$, the layer activation vector $\rvr^{(k)}[t]$ is included in the objective function terms ${\hat{I}}^{(\epsilon)}(\rvr^{(k - 1)}, \rvr^{(k)})[t]$ and ${\hat{I}}^{(\epsilon)}(\rvr^{(k )}, \rvr^{(k+1)})[t]$. Therefore, to calculate the gradient with respect to $\rvr^{(k)}[t]$, we can use expressions in (\ref{eq:MIS1}) and (\ref{eq:MIS2}). More specifically, we choose $\hat{J}_k(\rvr^{(k)})[t]=\overset{\rightarrow}{\hat{I}^{(\epsilon_{k-1})}}(\rvr^{(k - 1)},\rvr^{(k)})[t]+\overset{\leftarrow}{\hat{I}^{(\epsilon_k)}}(\rvr^{(k)},\rvr^{(k+1)})[t]$ for $k=1, \ldots, P-1$, to represent the components of the objective function in (\ref{eq:ldmiobjective}) involving $\rvr^{(k)}[t]$ . As described in Appendix \ref{sec:asymmetry1}, this choice is instrumental in avoiding weight transport problem. Similarly, we can write the component of the objective function in (\ref{eq:ldmiobjective}) that is dependent on the final layer activations as  $\hat{J}_P(\rvr^{(P)})[t]=\overset{\rightarrow}{\hat{I}^{(\epsilon_{P-1})}}(\rvr^{(P - 1)},\rvr^{(P)})[t]-\frac{\beta}{2}\|\rvr^{(P)}[t]-\rvy_T[t]\|_2^2$. 

Based on the derivations presented in  Appendix \ref{sec:gradients}, which directly incorporate the approximations from (\ref{eq:TaylorApp1}) and (\ref{eq:TaylorApp2}),  we can express the gradient of the objective function in  (\ref{eq:ldmiobjective}) with respect to  $\rvr^{(k)}$,  for $k=1, \ldots, P-1$ as:
\begin{eqnarray}
    \nabla_{\rvr^{(k)}}\hat{J}(\rvr^{(1)}, \ldots, \rvr^{(P)})[t]=2\gamma{\mB}_{\rvr^{(k)}}[t]\rvr^{(k)}[t]-\frac{1}{\epsilon_{k-1}}\overset{\rightarrow}{\rve}^{(k)}[t]-\frac{1}{\epsilon_k}\overset{\leftarrow}{\rve}^{(k)}[t], \label{eq:gradnk}
\end{eqnarray}
where $\gamma=\frac{1-\lambda_\rvr}{\lambda_\rvr}$,
\begin{eqnarray}
\overset{\rightarrow}{\rve}^{(k)}[t]=\rvr^{(k)}[t]-\mW^{(k-1)}_{ff}[t]\rvr^{(k-1)}[t],\quad
\overset{\leftarrow}{\rve}^{(k)}[t]=\rvr^{(k)}[t]-\mW^{(k)}_{fb}[t]\rvr^{(k+1)}[t], \label{eq:prederk}
\end{eqnarray}
and ${\mB}_{\rvr^{(k)}}[t]=(\hat{\rmR}_{\rvr^{(k)}}[t] + \epsilon_{k-1} \mI)^{-1}\approx (\hat{\rmR}_{\rvr^{(k)}}[t] + \epsilon_{k} \mI)^{-1}$. 
Similarly, for $k=P$, we have
\begin{eqnarray}
    \nabla_{\rvr^{(P)}}\hat{J}(\rvr^{(1)}, \ldots, \rvr^{(P)})[t]=\gamma{\mB}_{\rvr^{(P)}}[t]\rvr^{(P)}[t]-\frac{1}{\epsilon_{P-1}}\overset{\rightarrow}{\rve}^{(P)}[t]-\beta(\rvr^{(P)}[t]-\vy_T[t]).
\end{eqnarray}


\subsection{Neural network formulation based on information maximization}
\label{sec:criterionnn}
In this section, we develop a biologically plausible neural network grounded on the correlative information maximization-based network propagation model outlined in Section \ref{sec:DCIMCorInfoMax}.  To achieve this, we employ projected gradient ascent optimization for determining layer activations ${\rvr^{(1)}[t], \rvr^{(2)}[t], \ldots, \rvr^{(P)}[t]}$, which shape the network structure and dynamics, as well as updating the corresponding synapses that govern the learning dynamics.


\subsubsection{Network structure and neural dynamics}
\label{sec:networkStructure_OutputDynamics}
In this section, we show that the projected gradient ascent solution to the optimization in (\ref{eq:UnsupervisedLDMIobjective}) defines a multilayer recurrent neural network.  To this end, we introduce the intermediate variable $\rvu^{(k)}$ as the updated layer-$k$ activations prior to the projection onto the domain set $\Pcal^{(k)}$.
Utilizing the gradient expressions in (\ref{eq:gradnk})-(\ref{eq:prederk}), we can express the network dynamics for layers $k=1, \ldots, P-1$ as follows (see Appendix \ref{sec:derivation_netdynamics} for details): \looseness=-1
\begin{align}
    \tau_{\rvu}\frac{d \rvu^{(k)}[t;s]}{ds}&=-g_{lk}\rvu^{(k)}[t;s]+\frac{1}{\epsilon_k}\mM^{(k)}[t]\vr^{(k)}[t;s]-\frac{1}{\epsilon_{k-1}}\overset{\rightarrow}{\rve}_u^{(k)}[t;s]-\frac{1}{\epsilon_{k}}\overset{\leftarrow}{\rve}_u^{(k)}[t;s],\label{eq:nd1}\\
   \overset{\rightarrow}{\rve}_u^{(k)}[t;s]&=\rvu^{(k)}[t;s]-\mW^{(k-1)}_{ff}[t]\rvr^{(k-1)}[t;s],\quad \overset{\leftarrow}{\rve}_u^{(k)}[t;s]=\rvu^{(k)}[t;s]-\mW^{(k)}_{fb}[t]\rvr^{(k+1)}[t;s],\label{eq:nd2}\\
\rvr^{(k)}[t;s]&=\sigma_+(\rvu^{(k)}[t;s]),\label{eq:nd3}
\end{align} 
where $t$ is the discrete data index, $s$ is the continuous time index corresponding to network
dynamics, $\tau_{\rvu}$ is the update time constant, $\mM^{(k)}[t]=\epsilon_k(2\gamma \mB_{\rvr^{(k)}}[t]+g_{lk}\mI)$, and $\sigma_+$ represents the elementwise clipped-ReLU function corresponding to the projection onto the nonnegative unit-hypercube $\mathcal{B}_{\infty,+}$, defined as $\sigma_+(u)=\min(1,\max(u,0))$.  

To reinterpret the dynamics in (\ref{eq:nd1}) to (\ref{eq:nd3}) as a multi-compartmental neural network, for $k=1, \ldots, P-1$, we define the signals: \looseness=-1
\begin{eqnarray}
    \rvv^{(k)}_A[t;s]= \mM^{(k)}[t]\vr^{(k)}[t;s]+\mW^{(k)}_{fb}[t]\rvr^{(k+1)}[t;s], \label{eq:vahidden} \quad \rvv^{(k)}_B[t;s]= \mW^{(k-1)}_{ff}[t]\rvr^{(k-1)}[t;s],
\end{eqnarray}
which allow us to rewrite the network activation dynamics (\ref{eq:nd1}) to (\ref{eq:nd3}) as:
\begin{align}
    &\tau_{\rvu}\frac{d \rvu^{(k)}[t;s]}{ds}=-g_{lk}\rvu^{(k)}[t;s]+g_{A,k}(\rvv^{(k)}_A[t;s]-\rvu^{(k)}[t;s])+g_{B,k}(\rvv^{(k)}_B[t;s]-\rvu^{(k)}[t;s]), \label{eq:hiddynamics1}\\
  &\rvr^{(k)}[t;s]=\sigma_+(\rvu^{(k)}[t;s]), \label{eq:hiddynamics2}
  \end{align}
  where $g_{A,k}=\frac{1}{\epsilon_{k-1}}$ and $g_{B,k}=\frac{1}{\epsilon_k}$.
  Similarly, for the output layer, we employ the same expressions as (\ref{eq:hiddynamics1}) and (\ref{eq:hiddynamics2}) with $k=P$, except that in this case we have:
    \begin{eqnarray}
      \rvv^{(P)}_A[t;s]= \mM^{(P)}[t]\vr^{(k)}[t;s] -( \rvr^{(P)}[t;s]-\vy_T[t] )\label{eq:vaoutput}, \quad
      \rvv^{(P)}_B[t;s]= \mW^{(P-1)}_{ff}[t]\rvr^{(P-1)}[t;s], \label{eq:vboutput}
  \end{eqnarray}
  where $g_{B,P}=\frac{1}{\epsilon_{P-1}}$, $g_{A,P}=\beta$ and $\mM^{(P)}[t]=\beta^{-1}(\gamma \mB_{\rvr^{(P)}}[t]+g_{lk}\mI)$.

Remarkably, the equations (\ref{eq:vahidden}) to (\ref{eq:vboutput}) reveal a biologically plausible neural network that incorporates three-compartment pyramidal neuron models, as presented in \cite{sacramento2018dendritic,golkar2022constrained}. This intricate architecture, of which two-layer segment is demonstrated in Figure \ref{fig:twolayercorinfomax}, naturally emerges from the proposed correlative information maximization framework. In this network structure: \looseness=-1

\begin{itemize}
\item $\rvu^{(k)}$ embodies the membrane potentials for neuronal somatic compartments of the neurons at layer-$k$, where $\tau_{\rvu}$ is the membrane leak time constant of soma.
\item $\rvv^{(k)}_B$ corresponds to membrane potentials for basal dendrite compartments, receiving feedforward input originating from the previous layer.
\item $\rvv^{(k)}_A$ denotes the membrane potentials for distal apical dendrite compartments, which gather top-down input from the subsequent layer and lateral inputs represented by $\mM^{(k)}[t]\vr^{(k)}$ in (\ref{eq:vahidden}) and (\ref{eq:vaoutput}). Decomposing $\mM^{(k)}$ into $\mD^{(k)}-\mO^{(k)}$, we find that $\mD^{(k)}$ mirrors autapses\citep{lubke1996frequency}, and the off-diagonal component $\mO^{(k)}$ corresponds to lateral inhibition synapses. We use $\mathbf{i}^{(k)}=-\mO^{(k)}\rvr^{(k)}$ to represent the activations of SST interneurons \citep{urban2016somatostatin} that generate lateral inhibitions  to the apical dendrites.\looseness=-1
\item Forward (backward) prediction errors manifest in the membrane voltage differences between soma and basal (distal) compartments of the pyramidal neurons.
\item  Forward (backward) prediction coefficients $\mW_{ff}^{(k)}$ ($\mW_{fb}^{(k)}$) are associated with feedforward (top-down) synapses connecting layers $(k)$ and $(k+1)$.
\item The inverse of the regularization coefficient $\epsilon_k$ is related to the conductance between soma and dendritic compartments. This is compliant with the interpretation of the $\epsilon^{-1}$ in Appendix \ref{sec:roleepsilon} as the sensitivity parameter that determines the contribution of the prediction errors to the CMI.  Conversely,  at the output layer, the augmentation constant $\beta$ corresponds to the conductance between soma and distal compartments. This relationship can be motivated by modifying the objective in (\ref{eq:ldmiobjective}) as
\begin{eqnarray}
\sum_{k=0}^{P-1}\hat{I}^{(\epsilon_k)}(\rvr^{(k)}, \rvr^{(k+1)})[t]+ \frac{1}{2}\overset{\leftarrow}{\hat{I}^{(\beta^{-1})}}(\rvr^{(P)}, \rvy_T)[t],
\end{eqnarray}
where, through the first-order approximation, the $\rvr^{(P)}[t]$ dependent portion of $\overset{\leftarrow}{\hat{I}^{(\beta^{-1})}}(\rvr^{(P)}, \rvy_T)[t]$ can be expressed as $- \beta\|\rvr^{(P)}[t]-\mW_{fb}^{(P)}\rvy_T[t]\|_2^2$. For accuracy, we enforce $\mW_{fb}^{(P)}=\mI$.\looseness=-1
\end{itemize}


\subsection{Learning dynamics}
\label{sec:learningDynamics}
Network parameters consists of feedforward $\mW^{(k)}_{ff}$, feedback $\mW^{(k)}_{fb}$ and lateral $\mB^{(k)}$ coefficients.The learning dynamics of these coefficients are elaborated below:
\begin{itemize}
    \item \underline{\it Feedforward Coefficients} are connected to the forward prediction problem defined by the optimization in (\ref{eq:forwardpred}). We can define the corresponding online optimization objective function  
    as $C_{ff}(\mW^{(k)}_{ff})=\epsilon_k\|\mW_{ff}^{(k)}\|_F^2+\|\overset{\rightarrow}{\rve}^{(k+1)}[t]\|_2^2$ for which the the partial derivative is given by 
    \begin{eqnarray}
    \frac{\partial C_{ff}(\mW^{(k)}_{ff}[t])}{\partial \mW^{(k)}_{ff}}=2\epsilon_k\mW_{ff}^{(k)}[t]-2\overset{\rightarrow}{\rve}^{(k+1)}[t]\rvr^{(k)}[t]^T. \label{eq:partialforw}
    \end{eqnarray}
    In Appendix \ref{sec:altweightupdates}, we provide a discussion on rewriting (\ref{eq:partialforw}) in terms of the membrane voltage difference between the distal apical and soma compartments of the neuron, based on the equilibrium condition for the neuronal dynamics:
        \begin{eqnarray}
-\overset{\rightarrow}{\rve}^{(k+1)}[t]\rvr^{(k)}[t]^T=g^{-1}_{B,k}(g_{A,k}\rvv^{(k)}_A[t]-(g_{lk}+g_{A_k})\rvu_*^{(k)}[t]+\mathbf{h}_*[t])\rvr^{(k)}[t]^T \label{eq:outpW2maintext},
\end{eqnarray}
where $\mathbf{h}_*[t]$ is nonzero only for neurons that are silent or firing at the maximum rate.
\item Similarly, \underline{\it Feedback Coefficients} are connected to the backward prediction problem defined by the optimization in (\ref{eq:backwpred}), and the corresponding online optimization objective function
as $C_{fb}(\mW^{(k)}_{fb})=\epsilon_k\|\mW_{ff}^{(k)}\|_F^2+\|\overset{\leftarrow}{\rve}^{(k)}[t]\|_2^2$ for which the partial derivative is given by 
    \begin{eqnarray}
    \frac{\partial C_{fb}(\mW^{(k)}_{fb}[t])}{\partial\mW^{(k)}_{fb}}=2\epsilon_k\mW_{fb}^{(k)}[t]-2\overset{\leftarrow}{\rve}^{(k)}[t]\rvr^{(k+1)}[t]^T. \label{eq:partialbackw}
    \end{eqnarray}
To compute the updates of both feedforward and feedback coefficients, we use the EP approach \cite{EqProp}, where the update terms are obtained based on the contrastive expressions of partial derivatives in (\ref{eq:partialforw}) and (\ref{eq:partialbackw}) for the nudge phase, i.e., $\beta=\beta'>0$,  and the free phase, i.e., $\beta=0$, : \looseness=-1
\begin{eqnarray}
    \delta \mW^{(k)}_{ff}[t] \propto 
   \frac{1}{\beta'}\left((\overset{\rightarrow}{\rve}^{(k+1)}[t]\rvr^{(k)}[t]^T)\Big|_{\beta=\beta'}-(\overset{\rightarrow}{\rve}^{(k+1)}[t]\rvr^{(k)}[t]^T)\Big|_{\beta=0}\right) \label{eq:forwardEPupdate},\\
    \delta \mW^{(k)}_{fb}[t] \propto
   \frac{1}{\beta'}\left((\overset{\leftarrow}{\rve}^{(k)}[t]\rvr^{(k+1)}[t]^T)\Big|_{\beta=\beta'}-(\overset{\leftarrow}{\rve}^{(k)}[t]\rvr^{(k+1)}[t]^T)\Big|_{\beta=0}\right). \label{eq:backwardEPupdate}
\end{eqnarray}
\item \underline{\it Lateral Coefficients}, $\mathbf{B}^{(k)}$ are the inverses of the $\epsilon \mathbf{I}$ perturbed correlation matrices. We can use the update rule in \cite{bozkurt2023correlative} for their learning dynamics after the nudge phase:
\begin{eqnarray}
    \mB^{(k)}[t+1]=\lambda_\rvr^{-1}(\mB^{(k)}[t]-\gamma \rvz^{(k)}[t]\rvz^{(k)}[t]^T), \text{ where }\rvz^{(k)}=\mB^{(k)}[t]\rvr^{(k)}[t]\big|_{\beta=\beta'}. \label{eq:lateralB_update}
\end{eqnarray}
\end{itemize}
 As we derived in Appendix \ref{sec:lateralWeightUpdates}, we can rewrite the update rule of the lateral weights in terms of the updates of autapses and lateral inhibition synapses as follows:

\begin{align}
\mathbf{D}_{ii}^{(k)}[t+1]&=\lambda_{\mathbf{r}}^{-1}\mathbf{D}_{ii}^{(k)}[t]-\lambda_{\mathbf{r}}^{-1}\epsilon_k2\gamma^2 (\mathbf{z}_i^{(k)}[t])^2+\epsilon_k g_{lk} (1-\lambda_{\mathbf{r}}^{-1}), \quad \forall i \in \{1, \ldots, N_k\}\label{eq:lateralD_update}\\
\mathbf{O}^{(k)}_{ij}[t+1]&=\lambda_{\mathbf{r}}^{-1}\mathbf{O}^{(k)}[t]_{ij}+\lambda_{\mathbf{r}}^{-1}\epsilon_k2\gamma^2 \rvz_i^{(k)}[t] \rvz_j^{(k)}[t],\ \  \forall i,j \in \{1, \ldots, N_k\}, \text{ where } i\neq j \label{eq:lateralO_update}
\end{align}

\section{Discussion of results}
\label{sec:discussion}

\begin{itemize}
   \item In (\ref{eq:gradrk}), we devise an update for layer activation $\rvr^{(k)}$  by employing two distinct forms of the CMI associated with $\rvr^{(k)}$: $\overset{\rightarrow}{\hat{I}^{(\epsilon_{k-1})}}(\rvr^{(k - 1)},\rvr^{(k)})[t]$, the CMI with the preceding layer, encompassing the forward prediction error for estimating $\rvr^{(k)}$, and $\overset{\leftarrow}{\hat{I}^{(\epsilon_{k})}}(\rvr^{(k)},\rvr^{(k+1)})[t]$, the CMI with the subsequent layer, incorporating the backward prediction error for estimating $\rvr^{(k)}$. Employing these alternative expressions is crucial in circumventing the weight transport problem and offering a more biologically plausible framework. For further discussion, please refer to Appendix \ref{sec:asymmetry}.\looseness=-1

    \item In the context of  the proposed correlative information maximization framework, forward and backward predictive coding naturally emerges as a crucial mechanism. By incorporating both alternative expressions of CMI, the framework focuses on minimizing both forward and backward prediction errors between adjacent layers via feedforward and feedback connections.  These connections foster bidirectional information flow, thereby enhancing the overall learning process.

   \item Figure \ref{fig:twolayercorinfomax}  depicts the interplay between the CorInfoMax objective and the corresponding network architecture. The emergence of lateral connections and autapses can be attributed to the maximization of the unconditional layer entropy component of the CMI, which allows for efficient utilization of the available representation dimensions and avoids dimensional degeneracy. Simultaneously, the minimization of conditional entropies between adjacent layers gives rise to feedforward and feedback connections, effectively reducing redundancy within representations. \looseness=-1

   \item  We employ time-contrastive learning, as in GenRec \citep{o1996biologically}, EP \citep{EqProp} and CSM \citep{qin2021contrastive}, by implementing separate phases with Hebbian and anti-Hebbian updates, governed by an assumed teaching signal.
   It has been conjectured that the teaching signal in biological networks can be modeled by the oscillations in the brain \citep{whittington2019theories, PierreContrastive,ketz2013theta}. 
   Although the oscillatory rhythms and their synchronization in the brain are elusive, they are believed to play an important role in adaptive processes such as learning and predicting upcoming events \citep{Fellsynchronization, EngelDynamic}.\looseness=-1
\end{itemize}



\section{Numerical experiments}
\label{sec:numericalexperiments}
In this section, we evaluate the performance of our CorInfoMax framework with two layer fully connected networks on image classification tasks using three popular datasets: MNIST \citep{lecunMNIST}, Fashion-MNIST \citep{xiao2017fashionmnist}, and CIFAR10 \citep{krizhevsky2009learning}. We used layer sizes of $784, 500, 10$ for both MNIST and Fashion-MNIST datasets while we used layer sizes of $3072, 1000, 10$ for CIFAR10 dataset, and the final layer size $10$ corresponds to one-hot encoded ouput vectors. Further details including full set of hyperparameters can be found in Appendix \ref{sec:suppNumericalExperiments}. We compare the effectiveness of our approach against other contrastive methods, such as EP \citep{EqProp} and CSM \citep{qin2021contrastive}, as well as explicit methods, including PC \citep{whittington2017approximation} and PC-Nudge \citep{millidge2023backpropagation}, when training multilayer perceptron (MLP) architectures. \looseness=-1

We examine two distinct constraints on the activations of CorInfoMax Networks: (i) $\mathcal{B}_{\infty,+}$, representing the nonnegative part of the unit hypercube, and (ii) $\mathcal{B}_{1,+}=\{\rvr: \rvr\succcurlyeq 0, \|\rvr\|_1\le 1\}$, denoting the nonnegative part of the unit $\ell_1$-norm ball \citep{tatli2021tsp}. Table \ref{tab:ImageClassificationTest} presents the test accuracy results for each algorithm, averaged over 10 realizations along with the corresponding standard deviations. These findings demonstrate that CorInfoMax networks can achieve comparable or superior performance in relation to the state-of-the-art methods for the selected tasks. Additional information regarding these experiments, as well as further experiments, can be found in the Appendix. Our code is available online\footnote{ \href{https://github.com/BariscanBozkurt/Supervised-CorInfoMax}{\footnotesize https://github.com/BariscanBozkurt/Supervised-CorInfoMax}}. \looseness=-1

\begin{table}[h!]
  \caption{Test accuracy results (mean $\pm$ standard deviation from $n=10$ runs) for CorInfoMax networks are compared with other biologically-plausible algorithms. The performance of CSM on the CIFAR10 dataset is taken from \citep{qin2021contrastive}, while the remaining results stem from our own simulations.\\}
  \centering
  \begin{tabular}{llll}
    \toprule
         & MNIST     & FashionMNIST & CIFAR10 \\
    \midrule
    \textbf{CorInfoMax-$\mathcal{B}_{\infty,+}$} (Appendix \ref{sec:antisparse1}) & $97.62\pm0.1$  &$88.14\pm0.3$ & $51.86\pm0.3$ \\
    \textbf{CorInfoMax-$\mathcal{B}_{1,+}$} (Appendix \ref{sec:sparse}) & $97.71\pm0.1$ & $88.09\pm0.1$ & $51.19\pm0.4$   \\
    EP     & $97.61\pm0.1$ & $88.06\pm0.7$ &  $49.28\pm0.5$  \\
    CSM     & $98.08\pm0.1$ & $88.73\pm0.2$ & $40.79^\ast$   \\
    PC     & $98.17\pm0.2$ & $89.31\pm0.4$ & -       \\
    PC-Nudge     & $97.71\pm0.1$ &  $88.49\pm0.3$ & $48.58\pm0.7$   \\
    \midrule
    Feedback Alignment (with MSE Loss) & $97.99\pm0.03$ & $88.72\pm0.5$& $50.75\pm0.4$\\
    Feedback Alignment (with CrossEntropy Loss) & $97.95\pm0.08$ & $88.38\pm0.9$& $52.37\pm0.4$\\
    BP (with MSE Loss) & $97.58\pm0.01$& $88.39\pm0.1$& $52.75\pm0.1$\\
    BP (with CrossEntropy Loss) & $98.27\pm0.03$ &$89.41\pm0.2$ & $53.96\pm0.3$\\
    \bottomrule
    \label{tab:ImageClassificationTest}
  \end{tabular}
\end{table}





\section{Discussion and Conclusion}
\label{sec:conclusion}
In this article, we have presented the correlative information maximization (CorInfoMax) framework as a biologically plausible approach to constructing supervised neural network models. Our proposed method addresses the long-standing weight symmetry issue by providing a principled solution, which results in asymmetric forward and backward prediction networks. The experimental analyses demonstrates that CorInfoMax networks provide better or on-par performance in image classification tasks compared to other biologically plausible networks while alleviating the weight symmetry problem. Furthermore, the CorInfoMax framework offers a normative approach for developing network models that incorporate multi-compartment pyramidal neuron models, aligning more closely with the experimental findings about the biological neural networks. 
The proposed framework is useful in obtaining potential insights such as the role of lateral connections in embedding space expansion and avoiding degeneracy, feedback and feedforward connections for prediction to reduce redundancy, and activation functions/interneurons to shape feature space and compress. Despite the emphasis on supervised deep neural networks in our work, it’s crucial to highlight that our approach—replacing the backpropagation algorithm, which suffers from the weight transportation problem, with a normative method devoid of such issues—is potentially extendable to unsupervised and self-supervised learning contexts.\looseness=-1

One potential limitation of our framework, shared by other supervised approaches, is the necessity for model parameter search to improve accuracy. We discuss this issue in detail in Appendix \ref{sec:ablation}. Another limitation stems from the intrinsic nature of our approach, which involves the determination of neural activities through recursive dynamics (see Appendix \ref{sec:suppNumericalExperiments}). While this aspect is fundamental to our methodology, it does result in slower computation times compared to conventional neural networks in digital hardware implementation. However, it is worth noting that our proposed network, characterized by local learning rules, holds the potential for efficient and low-power implementations on future neuromorphic hardware chips. Furthermore, our method employs the time contrastive learning technique known as Equilibrium Propagation, which necessitates two distinct phases for learning. \looseness=-1

\section{Acknowledgments and Disclosure of Funding}
This research was supported by KUIS AI Center Research Award. B. Bozkurt acknowledges the support by Gatsby PhD programme, which is supported by the Gatsby Charitable Foundation (GAT3850). C. Pehlevan  is supported by NSF Award DMS-2134157, NSF CAREER Award IIS-2239780, and a Sloan Research Fellowship. This work has been made possible in part by a gift from the Chan Zuckerberg Initiative Foundation to establish the Kempner Institute for the Study of Natural and Artificial Intelligence.
\bibliography{corinfo_bibfile}
\newpage

\appendix


\section*{Appendix}
\renewcommand{\theequation}{A.\arabic{equation}}
\setcounter{equation}{0}


\section{Preliminaries on correlative entropy and mutual information}
\label{sec:prelim}

In this section, we present the essential background on correlative entropy and information measures that will be employed in our proposed approach. 
\subsection{Correlative entropy and correlative mutual information}
Consider a random vector $\rvx \in \R^m$ with a correlation matrix $\rmR_\rvx = E(\rvx \rvx^T)$. The correlative entropy for this random vector $\rvx$ is defined as follows \citep{erdogan2022icassp, zhouyin:21}:
\begin{align}
    H^{(\epsilon)}(\rvx) = \frac{1}{2} \log \det (\rmR_\rvx + \epsilon \mI) + \frac{m}{2} \log(2 \pi \euler), \label{eq:coentdef}
\end{align}
where $\epsilon > 0$ is a small positive constant. Please note that (\ref{eq:coentdef}) is an adapted version of Shannon's differential entropy for Gaussian vectors. However, we utilize it as an entropy definition based on second-order statistics, independent of the distribution of the vector $\rvx$. Furthermore,  the joint correlative entropy of two random vectors $\rvx \in \R^m$ and $\rvy \in \R^n$ can be expressed as:
\begin{align}
    H^{(\epsilon)}(\rvx, \rvy) &= \frac{1}{2} \log \det (\rmR_{\begin{bmatrix} \rvx \\ \rvy \end{bmatrix}} + \epsilon \mI) + \frac{m + n}{2 } \log \det (2 \pi \euler) \label{eq:cor_conditional_entropy}\\
    &= \frac{1}{2} \log \det \left( \begin{bmatrix} \rmR_{\rvx} + \epsilon \mI & \rmR_{\rvx \rvy} \\ \rmR_{\rvy \rvx} & \rmR_{\rvy} + \epsilon \mI \end{bmatrix}\right) + \frac{m + n}{2} \log \det (2 \pi \euler)\nonumber\\
    &= \frac{1}{2} \log \left( \det(\rmR_\rvx + \epsilon \mI)  \det(\rmR_\rvy + \epsilon \mI - \rmR_{\rvx \rvy}^T(\rmR_\vx + \epsilon \mI)^{-1} \rmR_{\rvx\rvy}) \right)\nonumber\\
&+ \frac{m + n}{2} \log \det (2 \pi \euler) \nonumber\\
&=\frac{1}{2} \log \det (\rmR_\rvx + \epsilon \mI) + \frac{m}{2} \log(2 \pi \euler) + \frac{1}{2} \log \det (\rmR_\rve + \epsilon \mI) + \frac{n}{2} \log(2 \pi \euler), \nonumber
\end{align}
where $\rmR_{\rvx\rvy} = E(\rvx \rvy^T)$, and $\rmR_\rve = \rmR_\rvy - \rmR_{\rvx \rvy}^T(\rmR_\vx + \epsilon \mI)^{-1} \rmR_{\rvx\rvy}$ is the autocorrelation matrix of the error vector corresponding to the best linear minimum mean square estimator (MMSE) of $\rvy$ from $\rvx$, and the third equality is obtained by writing the determinant in terms of a principle submatrix and its Schur's complement \citep{kailath2000linear}. This derivation naturally leads to the expression of $H^{(\epsilon)}(\rvx, \rvy) = H^{(\epsilon)}(\rvx) + H^{(\epsilon)}(\rvy |_L \rvx)$. Note that we use the notation $|_L$ to distinguish that $H^{(\epsilon)}(\rvx | \rvy)$ requires to use $ \rmR_{\rvx| \rvy} $ that is not equal to $\rmR_\rvx - \rmR_{\rvy \rvx}^T(\rmR_\vy + \epsilon \mI)^{-1} \rmR_{\rvy\rvx}$ in general. Therefore, we can express the conditional correlative entropy definitions as \citep{erdogan2022icassp}
\begin{align}
    H^{(\epsilon)}(\rvy |_L \rvx) &= \frac{1}{2} \log \det (\rmR_\rvy - \rmR_{\rvx \rvy}^T(\rmR_\vx + \epsilon \mI)^{-1} \rmR_{\rvx\rvy} + \epsilon \mI) + \frac{n}{2} \log(2 \pi \euler) \nonumber \\
    H^{(\epsilon)}(\rvx |_L \rvy) &= \frac{1}{2} \log \det (\rmR_\rvx - \rmR_{\rvy \rvx}^T(\rmR_\vy + \epsilon \mI)^{-1} \rmR_{\rvy\rvx} + \epsilon \mI) + \frac{m}{2} \log(2 \pi \euler). \nonumber
\end{align}
Using the alternative Schur's complement in (\ref{eq:cor_conditional_entropy}), it can be also shown that $H^{(\epsilon)}(\rvx, \rvy) = H^{(\epsilon)}(\rvy) + H^{(\epsilon)}(\rvx |_L \rvy)$. Based on these definitions, the correlative mutual information (CMI) is defined as follows
\begin{align}
    I^{(\epsilon)}(\rvx, \rvy) &= H^{(\epsilon)}(\rvy) - H^{(\epsilon)}(\rvy |_L \rvx), \nonumber \\
    &= H^{(\epsilon)}(\rvx) - H^{(\epsilon)}(\rvx|_L \rvy), \label{eq:corInfoSymmetricDefinition}\\
    &= H^{(\epsilon)}(\rvx) + H^{(\epsilon)}(\rvy) - H^{(\epsilon)}(\rvx, \rvy), \nonumber
\end{align}
More explicitly, we can write the correlative mutual information using the following two alternative yet equivalent expressions,
\begin{align*}
    I^{(\epsilon)}(\rvx, \rvy) &= \frac{1}{2} \log \det (\rmR_\rvx + \epsilon \mI) - \frac{1}{2} \log \det (\rmR_\rvx - \rmR_{\rvy \rvx}^T(\rmR_\vy + \epsilon \mI)^{-1} \rmR_{\rvy\rvx} + \epsilon \mI),\\
    &= \frac{1}{2} \log \det (\rmR_\rvy + \epsilon \mI) - \frac{1}{2} \log \det (\rmR_\rvy - \rmR_{\rvx \rvy}^T(\rmR_\vx + \epsilon \mI)^{-1} \rmR_{\rvx\rvy} + \epsilon \mI).
\end{align*}
At its core, the correlative mutual information, $I^{(\epsilon)}(\rvx, \rvy)$, quantifies the degree of correlation or linear association between the random vectors $\rvx$ and $\rvy$ \citep{erdogan2022icassp}.

\subsection{On the interpretation of \texorpdfstring{$\epsilon$}{Lg} parameter}
\label{sec:roleepsilon}
The definition of correlative entropy as presented in equation (\ref{eq:coentdef}) includes a perturbation term, $\epsilon$, in the eigenvalues of the correlation matrix argument of the log-determinant. At first glance, this term appears to function as a correction factor to compensate for rank-deficient correlation matrices of degenerate random vectors. From this perspective, this adjustment serves two primary purposes:
\begin{itemize}
    \item [i.] To establish a finite lower bound for the entropy, and
    \item [ii.] To circumvent numerical optimization issues, given that the derivative of the $\log\det$ function is the inverse of its argument.
\end{itemize}
In fact, robust matrix factorization methods that rely on determinant-maximization often use the perturbation term $\epsilon \mathbf{I}$ for these reasons, as suggested by Fu et al.  \citep{fu:2016}.

Beyond these, upon examining the expression for correlative mutual information more closely,
\begin{align}
    I^{(\epsilon)}(\rvx, \rvy) &= \frac{1}{2} \log \det (\rmR_\rvx + \epsilon \mI) - \frac{1}{2} \log \det (\rmR_{\rve_{\rvx|\rvy}}  + \epsilon \mI), \label{eq:cormutinf}
\end{align}
we can draw the following insights:
\begin{itemize}
    \item The matrix $\rmR_{\rve_{\rvx|\rvy}}=\rmR_\rvx - \rmR_{\rvy \rvx}^T(\rmR_\vy + \epsilon \mI)^{-1} \rmR_{\rvy\rvx}$ corresponds to the error correlation matrix for the best linear regularized minimum mean square estimator of $\vx$ from $\vy$. This estimator is obtained as  the solution of the optimization problem
    \begin{eqnarray}
    \underset{\mW_{\rvx|\rvy}}{\text{minimize }} {E(\|{\rvx}-\mW_{\rvx|\rvy}\rvy\|_2^2)+\epsilon\|\mW_{\rvx|\rvy}\|_F^2}.
\end{eqnarray}
In this context, $\epsilon$ parameter acts as a regularizing coefficient for the linear estimation problem integral to measuring linear dependence between the two arguments of the CMI.

\item  Maximizing the CMI given by equation (\ref{eq:cormutinf}) can be accomplished by increasing the correlative entropy of  $\rvx$ while decreasing the correlative entropy of the estimation error $\rve_{\rvx|\rvy}$. Given the relationship $\mR_{\rvx}\succeq \mR_{\rve_{\rvx|\rvy}}$, we anticipate that the choice of $\epsilon$  will primarily influence the correlative entropy of $\rve_{\rvx|\rvy}$. Indeed, since $\epsilon$ is added to all the eigenvalues of $\mR_{\rve_{\rvx|\rvy}}$, reducing its eigenvalues below $\epsilon$ would have only incremental increase in the mutual information. As such, a smaller$\epsilon$ value will place greater emphasis on reducing the estimation error  $\rve_{\rvx|\rvy}$. Consequently, one can consider  $\epsilon^{-1}$  can be viewed as an indicator of the sensitivity of the CMI to the levels of estimation error $\rve_{\rvx|\rvy}$, determining how far we need to push down the estimation error values to increase the CMI.
\item The role of $\epsilon$ parameter in adjusting sensitivity to estimation errors becomes evident when we linearize the $\mR_{\rve_{\rvx|\rvy}}$ dependent (second) term in (\ref{eq:cormutinf}) using the truncated Taylor series approximation equation (\ref{eq:linearlogdet}) in Appendix \ref{sec:taylorappr} (by choosing $\mathbf{A}=\epsilon \mI$ and $\mathbf{\Delta}=\mR_{\rve_{\rvx|\rvy}}$ in (\ref{eq:linearlogdet})):
\begin{align}
    I^{(\epsilon)}(\rvx, \rvy) &\approx \frac{1}{2} \log \det (\rmR_\rvx ) - \frac{\epsilon^{-1}}{2} \text{Tr} (\rmR_{\rve_{\rvx|\rvy}}) +\frac{m}{2}\log(\epsilon). \label{eq:cormutinflin}
\end{align}
In the above expression, we have assumed that the choice of $\epsilon$ is less than the eigenvalues of $\rmR_\rvx$, so that we can approximate  $\rmR_\rvx +\epsilon \mI \approx \rmR_\rvx$, and greater than the eigenvalues of $\rmR_{\rve_{\rvx|\rvy}}$. As apparent from equation (\ref{eq:cormutinflin}), $\epsilon^{-1}$ serves as a sensitivity parameter that determines the contribution of estimation errors $\rve_{\rvx|\rvy}$  to the CMI.

\end{itemize}

\section{Linear approximation of correlative entropy}
\label{sec:taylorappr}
In this section, we provide linear approximation of the $\log\det$ terms on the rightmost terms of (\ref{eq:MIS1}) and (\ref{eq:MIS2}). 
For this purpose we utilize the the first-order Taylor series approximation of the $\log\det$ function:
\begin{align}
\log\det(\mathbf{A}+\mathbf{\Delta})&\approx\log\det(\mathbf{A})+\text{Tr}(\nabla_\mathbf{A}\log\det(\mathbf{A})^T\mathbf{\Delta})\\
 &\approx \log\det(\mathbf{A})+\text{Tr}({\mathbf{A}^{-1}}\mathbf{\Delta}), \label{eq:linearlogdet}
\end{align}
assuming $\mathbf{A}$ is Hermitian (or real symmetric). Therefore, using (\ref{eq:linearlogdet}) and choosing $\mathbf{A}=\epsilon_k \mI$ and $\mathbf{\Delta}={\hat{\mR}}_{\overset{\rightarrow}{\rve}^{(k+1)}}[t]$, we can approximate the rightmost term of (\ref{eq:MIS1}) corresponding to the correlative entropy of  forward error $\overset{\rightarrow}{\rve}^{(k+1)}$
\begin{align}
    \log \det \left({\hat{\mR}}_{\overset{\rightarrow}{\rve}^{(k+1)}}[t] + \epsilon_k \mI\right)&\approx \frac{1}{\epsilon_k}\Tr\left({\hat{\mR}}_{\overset{\rightarrow}{\rve}^{(k+1)}}[t]\right)+N_{k+1}\log(\epsilon_k)\nonumber \\
    &\hspace*{-1.7in}=\frac{1}{\epsilon_k}\sum_{i=1}^t\lambda_\rvr^{t-i}\|\rvr^{(k+1)}[i]-\mathbf{W}_{ff,*}^{(k)}[t]\rvr^{(k)}[i]\|_2^2+\epsilon_k\|\mW_{ff,*}^{(k)}[t]\|_F^2+N_{k+1}\log(\epsilon_k). \label{eq:logdettoTrff}
    \end{align}
    This approximation would be more accurate for prediction error correlation matrices with smaller eigenvalues.
    
  Similarly, using (\ref{eq:linearlogdet}) and choosing $\mathbf{A}=\epsilon_k \mI$ and $\mathbf{\Delta}={\hat{\mR}}_{\overset{\rightarrow}{\rve}^{(k)}}[t]$, we can approximate the rightmost term of (\ref{eq:MIS2}) corresponding to the correlative entropy of  forward error $\overset{\leftarrow}{\rve}^{(k)}$  
    \begin{align}
    \log \det \left({\hat{\mR}}_{\overset{\leftarrow}{\rve}^{(k)}}[t] + \epsilon_k \mI\right)&\approx \frac{1}{\epsilon_k}\Tr\left({\hat{\mR}}_{\overset{\leftarrow}{\rve}^{(k)}}[t]\right)+N_k\log(\epsilon_k)\nonumber \\
    &\hspace*{-1.7in}=\frac{1}{\epsilon_k}\sum_{i=1}^t\lambda_\rvr^{t-i}\|\rvr^{(k)}[i]-\mathbf{W}_{fb,*}^{(k)}[t]\rvr^{(k+1)}[i]\|_2^2+\epsilon_k\|\mW_{fb,*}^{(k)}[t]\|_F^2+N_k\log(\epsilon_k).\label{eq:logdettoTrfb}
\end{align}
Note that in (\ref{eq:logdettoTrff}), $\mathbf{W}_{ff,*}^{(k)}[t]$ denotes the optimal linear regularized weighted least squares forward predictor coefficients in predicting $\rvr^{(k+1)}[i]$ from $\rvr^{(k)}[i]$  for $i=1, \ldots, t$. Likewise,  $\mathbf{W}_{fb,*}^{(k)}[t]$ in (\ref{eq:logdettoTrfb}) represents the optimal linear regularized weighted least squares backward predictor coefficients in predicting $\rvr^{(k)}[i]$ from $\rvr^{(k+1)}[i]$  for $i=1, \ldots, t$. 

\section{ Background on polytopic representations}
\label{sec:polytopicrepresentations}
Convex polytopes, compact sets formed by the intersections of halfspaces ~\cite{brondsted2012introduction}, serve as constraint domains for latent representations. The choice of a particular polytope reflects the attribute assignments to these vectors. For instance, the $\ell_1$-norm-ball polytope enforces sparsity and finds extensive use in machine learning, signal processing and computational neuroscience~\cite{donoho2006most,elad2010sparse,duchi2008efficient,babatas2018algorithmic}. Conversely, the $\ell_\infty$-norm-ball polytope is prevalent in antisparse (democratic) representations, especially in bounded component analysis applications~\cite{studer2014democratic,erdogan2013class,inan2014convolutive}.

The Polytopic Matrix Factorization (PMF) extends these examples to an infinite set of polytopes, incorporating specific symmetry restrictions for identifiability~\cite{tatli2021tsp}. In the PMF paradigm, observation vectors $\{\mathbf{y}_1, \ldots, \mathbf{y}_N\} \subset \R^M$ are modeled as unknown linear transformations of latent vectors $\{\mathbf{s}_1, \ldots, \mathbf{s}_N\}$ from a selected polytope $\mathcal{P}$:
\begin{eqnarray}
    \underbrace{\left[\begin{array}{cccc} \mathbf{y}_1 & \mathbf{y}_2 & \ldots & \mathbf{y}_N\end{array}\right]}_{\mathbf{Y}}=\mathbf{H} \underbrace{\left[\begin{array}{cccc} \mathbf{s}_1 & \mathbf{s}_2 & \ldots & \mathbf{s}_N\end{array}\right]}_{\mathbf{S}}, \label{eq:MatFact}
\end{eqnarray}
where $\mathbf{H}$ denotes the unknown linear transformation. The PMF's primary objective is to deduce the factors $\mathbf{H}$ and $\mathbf{S}$ from the observation matrix $\mathbf{Y}$, by exploiting the information that the columns of $\mathbf{S}$ as "representative" samples from $\mathcal{P}$. The shape of the chosen polytope determine the latent vector features  by combining common attributes such as sparsity, nonnegativity and anti-sparsity in subvector level.    For example, the reference \cite{bozkurt2023correlative} proposes the canonical polytope description 
\begin{eqnarray*}
\displaystyle
\Pcal=\left\{\vs \in \mathbb{R}^n\ \mid \evs_i\in[-1,1] \, \forall i\in \mathcal{I}_s,\, \evs_i\in[0,1] \, \forall i\in \mathcal{I}_+, \, \left\|\vs_{\mathcal{J}_l}\right\|_1\le 1, \, \mathcal{J}_l\subseteq \sZ_n, \, l\in\sZ_L  \right\}, \label{eq:polygeneral}
\end{eqnarray*}
where $L$ is the number of mutually sparse subvector constraints, $\mathcal{I}_+$ is the index set of nonnegative elements, $\mathcal{I}_s$ is the index set of signed elements, the sets $\mathcal{J}_l, l\in \sZ_L$ are the index sets for the sparse subvectors. The availability of infinitely many polytope options provides a powerful and diverse toolkit for representing and characterizing latent vectors.

The online correlative information maximization  solution for obtaining factors in (\ref{eq:MatFact}), combined with the polytopic constraints on the columns of $\mathbf{S}$ results in  biologically plausible neural networks with local learning  rules~\cite{bozkurt2022biologicallyplausible}. These polytopic constraints manifest as piecewise linear neural activation functions, such as ReLU and clipping functions.

In a vein similar to the unsupervised problem in~\cite{bozkurt2022biologicallyplausible}, our proposed framework employs polytopic representations to characterize embedding vectors for each network layer. The inclusion of polytopic constraints influences:
\begin{itemize}
    \item The characterization of embeddings based on assigned attributes.
    \item The network's nonlinear component via piecewise activation functions.
\end{itemize}

In summary, polytopic representations offer a versatile and mathematically rigorous framework for modeling latent vectors in various applications, from machine learning to signal processing. Their ability to encapsulate diverse attributes, such as sparsity and non-negativity, provides a rich characterization of latent features.

\section{Gradient derivation for the CorInfoMax objective}
\label{sec:gradients}
In this section, we offer a derivation of the gradients for the CorInfoMax objective function $\hat{J}(\rvr^{(1)}, \ldots, \rvr^{(P)})$ in (\ref{eq:ldmiobjective}) with respect to the layer activation vector $\rvr^{(k)}$. As outlined in Section \ref{sec:criterionsamp}, the components of $\hat{J}(\rvr^{(1)}, \ldots, \rvr^{(P)})$ containing $\rvr^{(k)}$ can be expressed as:
\begin{eqnarray}
    \hat{J}_k(\rvr^{(k)})[t]=\overset{\rightarrow}{\hat{I}^{(\epsilon_{k-1})}}(\rvr^{(k - 1)},\rvr^{(k)})[t]+\overset{\leftarrow}{\hat{I}^{(\epsilon_k)}}(\rvr^{(k)},\rvr^{(k+1)})[t],
\end{eqnarray}
for $k=1, \ldots, P$, and 
\begin{eqnarray}
    \hat{J}_P(\rvr^{(P)})[t]=\overset{\rightarrow}{\hat{I}^{(\epsilon_{P-1})}}(\rvr^{(P - 1)},\rvr^{(P)})[t]-\frac{\beta}{2}\|\rvr^{(P)}[t]-\vy_T[t]\|_2^2.
\end{eqnarray}

To simplify our derivations, as explained in Appendix \ref{sec:taylorappr}, we replace the correlative entropy terms for the prediction errors (the rightmost terms of (\ref{eq:MIS1}) and (\ref{eq:MIS2})) with their linear approximations in (\ref{eq:logdettoTrff})-(\ref{eq:logdettoTrfb}). Thus, the resulting gradient with respect to $\rvr^{(k)}[t]$ can be expressed as:
\begin{eqnarray}
\nabla_{\rvr^{(k)}}\hat{J}(\rvr^{(1)}, \ldots, \rvr^{(P)})[t]&=&\nabla_{\rvr^{(k)}}\hat{J}_k(\rvr^{(k)})[t]\nonumber \\&=&\nabla_{\rvr^{(k)}}\overset{\rightarrow}{\hat{I}^{(\epsilon_{k-1})}}(\rvr^{(k - 1)},\rvr^{(k)})[t]+\nabla_{\rvr^{(k)}}\overset{\leftarrow}{\hat{I}^{(\epsilon_k)}}(\rvr^{(k)},\rvr^{(k+1)})[t] \nonumber\\
&=& \frac{1}{2}\nabla_{\rvr^{(k)}} (\log \det (\hat{\rmR}_{\rvr^{(k)}}[t]+ \epsilon_{k-1} \mI)+\log \det (\hat{\rmR}_{\rvr^{(k)}}[t]+ \epsilon_{k} \mI))\nonumber\\ 
&&-\frac{1}{\epsilon_{k-1}}\overset{\rightarrow}{\rve}^{(k)}[t]-\frac{1}{\epsilon_k}\overset{\leftarrow}{\rve}^{(k)}[t], \label{eq:gradrk}
\end{eqnarray}
where 
\begin{eqnarray}
\overset{\rightarrow}{\rve}^{(k)}[t]=\rvr^{(k)}[t]-\mW^{(k-1)}_{ff}[t]\rvr^{(k-1)}[t],\quad
\overset{\leftarrow}{\rve}^{(k)}[t]=\rvr^{(k)}[t]-\mW^{(k)}_{fb}[t]\rvr^{(k+1)}[t] \label{eq:preder}
\end{eqnarray}
are the forward and backward prediction errors at level-$k$, based on the current estimates of the corresponding predictor matrices. Following the procedure detailed in \cite{bozkurt2023correlative}, for the gradient term in (\ref{eq:gradrk}), we obtain: 
\begin{eqnarray}
   \frac{1}{2}\nabla_{\rvr^{(k)}} (\log \det (\hat{\rmR}_{\rvr^{(k)}}[t]+ \epsilon_{k-1} \mI)+\log \det (\hat{\rmR}_{\rvr^{(k)}}[t]+ \epsilon_{k} \mI))= 2\gamma{\mB}_{\rvr^{(k)}}[t]\rvr^{(k)}[t], \label{eq:gradB}
\end{eqnarray}
where ${\mB}{\rvr^{(k)}}[t]=(\hat{\rmR}{\rvr^{(k)}}[t] + \epsilon_{k-1} \mI)^{-1}\approx (\hat{\rmR}{\rvr^{(k)}}[t] + \epsilon_{k} \mI)^{-1}$ and $\gamma=\frac{1-\lambda_\rvr}{\lambda_\rvr}$. The gradient of the objective for the final layer can be written as: 
\begin{eqnarray}
    &&\hspace*{-0.7in}\nabla_{\rvr^{(P)}} \hat{J}_P(\rvr^{(P)})[t]=\gamma{\mB}_{\rvr^{(P)}}[t]\rvr^{(P)}[t]-\frac{1}{\epsilon_{P-1}}\overset{\rightarrow}{\rve}^{(P)}[t]-\beta(\rvr^{(P)}[t]-\vy_T[t]). \label{eq:gradrpap}
\end{eqnarray}
In conclusion, combining expressions (\ref{eq:gradk})-(\ref{eq:gradrpap}), we can formulate:
\begin{eqnarray}
    \nabla_{\rvr^{(k)}}\hat{J}(\rvr^{(1)}, \ldots, \rvr^{(P)})[t]=2\gamma{\mB}_{\rvr^{(k)}}[t]\rvr^{(k)}[t]-\frac{1}{\epsilon_{k-1}}\overset{\rightarrow}{\rve}^{(k)}[t]-\frac{1}{\epsilon_k}\overset{\leftarrow}{\rve}^{(k)}[t], \label{eq:gradk}
\end{eqnarray}
for $k=1, \ldots, P-1$, and
\begin{eqnarray}
    \nabla_{\rvr^{(P)}}\hat{J}(\rvr^{(1)}, \ldots, \rvr^{(P)})[t]=\gamma{\mB}_{\rvr^{(P)}}[t]\rvr^{(P)}[t]-\frac{1}{\epsilon_{P-1}}\overset{\rightarrow}{\rve}^{(P)}[t]-\beta(\rvr^{(P)}[t]-\vy_T[t]).
\end{eqnarray}

\section{Derivation of CorInfoMax network dynamics}
\label{sec:derivation_netdynamics}
In this section, we provide details about the derivation of the CorInfoMax network dynamics equations (\ref{eq:nd1})-(\ref{eq:nd3}) in Section \ref{sec:networkStructure_OutputDynamics}. 

As discussed in Section \ref{sec:criterionsamp}, network dynamics naturally emerge from the application of the projected gradient ascent algorithm to obtain solution of the optimization problem  (\ref{eq:UnsupervisedLDMIobjective}).  Therefore, we update layer activations $\rvr^{(k)}[t]$ using $\nabla_{\rvr^{(k)}}\hat{J}(\rvr^{(1)}, \ldots, \rvr^{(P)})$ and then project the result to the presumed domain $\Pcal^{(k)}=\mathcal{B}_{\infty,+}$.

Following the approach in \cite{rozell:2008}, we define pre-projection signal $\rvu^{(k)}$, which is updated with gradient, and then project it onto $\Pcal^{(k)}=\mathcal{B}_\infty$. We use continuous dynamic update in the form
\begin{align}
\tau_u \frac{d \rvu^{(k)}[t;s]}{ds}&=\nabla_\rvr^{(k)}\hat{J}(\rvr^{(1)}, \ldots, \rvr^{(P)})[t;s] \label{eq:dynamics_1}\\
\rvr^{(k)}[t;s]&=\sigma_{+}(\rvu^{(k)}[t;s]).
\end{align}
In above equations, $t$ is the discrete data index, referring to the index of input ($\rvx[t]$) and label ($\rvy_T[t]$) samples, as defined in Section \ref{sec:DCINetworkModel}. Whereas, $s$ is the continuous time index corresponding to network dynamics. Furthermore, $\sigma_+$ represents the elementwise clipped-ReLU function corresponding to the projection onto the nonnegative unit-hypercube $\mathcal{B}_{\infty,+}$, defined as $\sigma_+(u)=\min(1,\max(u,0))$.

We can plug in the gradient expression in (\ref{eq:gradnk}) in (\ref{eq:dynamics_1}) to obtain

\begin{align}
\tau_u \frac{d \rvu^{(k)}[t;s]}{ds}&=2\gamma{\mB}_{\rvr^{(k)}}[t]\rvr^{(k)}[t;s]-\frac{1}{\epsilon_{k-1}}\overset{\rightarrow}{\rve}^{(k)}[t;s]-\frac{1}{\epsilon_k}\overset{\leftarrow}{\rve}^{(k)}[t;s] \label{eq:dynamics_2}\\
\overset{\rightarrow}{\rve}_u^{(k)}[t;s]&=\rvu^{(k)}[t;s]-\mW^{(k-1)}_{ff}[t]\rvr^{(k-1)}[t;s],\quad \overset{\leftarrow}{\rve}_u^{(k)}[t;s]=\rvu^{(k)}[t;s]-\mW^{(k)}_{fb}[t]\rvr^{(k+1)}\\
\rvr^{(k)}[t;s]&=\sigma_{+}(\rvu^{(k)}[t;s]). \label{eq:dynamics_2b}
\end{align}
In order to covert neural dynamics in (\ref{eq:dynamics_2})-(\ref{eq:dynamics_2b}) to leaky-integrate-and-fire form, we add leaky term $-g_{lk}\rvu^{(k)}[t;s]$ and modify the output feedback form to compensate this addition:
\begin{align}
\tau_u \frac{d \rvu^{(k)}[t;s]}{ds}&=-g_{lk}\rvu^{(k)}[t;s]+(2\gamma{\mB}_{\rvr^{(k)}}[t]+g_{lk}\mI)\rvr^{(k)}[t]-\frac{1}{\epsilon_{k-1}}\overset{\rightarrow}{\rve}^{(k)}[t]-\frac{1}{\epsilon_k}\overset{\leftarrow}{\rve}^{(k)}[t] \label{eq:dynamics_3}\\
\overset{\rightarrow}{\rve}_u^{(k)}[t;s]&=\rvu^{(k)}[t;s]-\mW^{(k-1)}_{ff}[t]\rvr^{(k-1)}[t;s],\quad \overset{\leftarrow}{\rve}_u^{(k)}[t;s]=\rvu^{(k)}[t;s]-\mW^{(k)}_{fb}[t]\rvr^{(k+1)}\\
\rvr^{(k)}[t;s]&=\sigma_{+}(\rvu^{(k)}[t;s]). \label{eq:dynamics_3b}
\end{align}
Finally, substituting $\mM^{(k)}[t]=\epsilon_k(2\gamma{\mB}_{\rvr^{(k)}}[t]+g_{lk}\mI)$, we can rewrite the neural dynamics as
\begin{align}
\tau_u \frac{d \rvu^{(k)}[t;s]}{ds}&=-g_{lk}\rvu^{(k)}[t;s]+\frac{1}{\epsilon_k}\mM^{(k)}[t]\rvr^{(k)}[t]-\frac{1}{\epsilon_{k-1}}\overset{\rightarrow}{\rve}^{(k)}[t]-\frac{1}{\epsilon_k}\overset{\leftarrow}{\rve}^{(k)}[t] \label{eq:dynamics_4}\\
\overset{\rightarrow}{\rve}_u^{(k)}[t;s]&=\rvu^{(k)}[t;s]-\mW^{(k-1)}_{ff}[t]\rvr^{(k-1)}[t;s],\quad \overset{\leftarrow}{\rve}_u^{(k)}[t;s]=\rvu^{(k)}[t;s]-\mW^{(k)}_{fb}[t]\rvr^{(k+1)}\\
\rvr^{(k)}[t;s]&=\sigma_{+}(\rvu^{(k)}[t;s]). \label{eq:dynamics_4b}
\end{align}

\section{Derivation of lateral weight updates in terms of autopses and lateral inhibition synapses}
\label{sec:lateralWeightUpdates}
Here, we provide the derivation of the lateral weight updates provided in (\ref{eq:lateralD_update}) and (\ref{eq:lateralO_update}). Recall that in Section \ref{sec:networkStructure_OutputDynamics}, we defined  $\mM^{(k)}[t]=\epsilon_k(2\gamma \mB_{\rvr^{(k)}}[t]+g_{lk}\mI)$. Therefore, using (\ref{eq:lateralB_update}), we can write:

\begin{align}
    \mathbf{M}^{(k)}[t+1]&=\epsilon_k(2\gamma\lambda_{\mathbf{r}}^{-1}(\mathbf{B}^{(k)}[t]-\gamma \mathbf{z}^{(k)}[t]\mathbf{z}^{(k)}[t]^T)+g_{lk}\mathbf{I})\\
    &= \lambda_{\mathbf{r}}^{-1}\epsilon_k2\gamma\mathbf{B}^{(k)}[t]-\lambda_{\mathbf{r}}^{-1}\epsilon_k2\gamma^2 \mathbf{z}^{(k)}[t]\mathbf{z}^{(k)}[t]^T+\epsilon_kg_{lk}\mathbf{I} \label{eq:lateralM_update1}
\end{align}

Using the relationship
$\mathbf{B}^{(k)}[t]=\frac{1}{\epsilon_k2\gamma}\mathbf{M}^{(k)}[t]-\frac{g_{lk}}{2\gamma}\mathbf{I}$, we can rearrange the right-hand side of (\ref{eq:lateralM_update1}) such that
\begin{eqnarray}
    \mathbf{M}^{(k)}[t+1]=\lambda_{\mathbf{r}}^{-1}\mathbf{M}^{(k)}[t]-\lambda_{\mathbf{r}}^{-1}\epsilon_k2\gamma^2 \mathbf{z}^{(k)}[t]\mathbf{z}^{(k)}[t]^T+\epsilon_k g_{lk}(1 -\lambda_{\mathbf{r}}^{-1})\mathbf{I} \label{eq:lateralM_update2}
\end{eqnarray}

Here, $\mathbf{z}^{(k)}[t] = \left(\frac{1}{\epsilon_k2\gamma}\mathbf{M}^{(k)}[t]\mathbf{r}^{(k)} - \frac{g_{lk}}{2\gamma}\mathbf{r}^{(k)}\right)\Big|_{\beta=\beta'}$. It's worth noting that in Section \ref{sec:networkStructure_OutputDynamics}, we decomposed $\rmM^{(k)}[t]$ into $\rmD^{(k)}[t]$ and $\rmO^{(k)}[t]$ as $\rmM^{(k)}[t] = \rmD^{(k)}[t] - \rmO^{(k)}[t]$, representing autapses and lateral inhibition synapses, respectively. Therefore, the update rule in (\ref{eq:lateralM_update2}) leads to the following updates: 
\begin{align}
\mathbf{D}_{ii}^{(k)}[t+1]&=\lambda_{\mathbf{r}}^{-1}\mathbf{D}_{ii}^{(k)}[t]-\lambda_{\mathbf{r}}^{-1}\epsilon_k2\gamma^2 (\mathbf{z}_i^{(k)}[t])^2+\epsilon_k g_{lk} (1-\lambda_{\mathbf{r}}^{-1}), \quad \forall i \in \{1, \ldots, N_k\}\\
\mathbf{O}^{(k)}_{ij}[t+1]&=\lambda_{\mathbf{r}}^{-1}\mathbf{O}^{(k)}[t]_{ij}+\lambda_{\mathbf{r}}^{-1}\epsilon_k2\gamma^2 \rvz_i^{(k)}[t] \rvz_j^{(k)}[t],\ \  \forall i,j \in \{1, \ldots, N_k\}, \text{ where } i\neq j 
\end{align}

\section{On the asymmetry of feedforward and feedback weights}
\label{sec:asymmetry}
In the main article, we emphasized that one of the key contributions of the proposed supervised CorInfoMax framework is to offer a natural resolution to the weight symmetry problem found in some biologically plausible neural network frameworks. In this section, we aim to expand upon this aspect and supplement the discussion provided in Section \ref{sec:discussion}.

\subsection{The importance of the choice of alternative CMI expressions}
\label{sec:asymmetry1}
In developing the update formula for the layer activation $\rvr^{(k)}$, we used the gradient (\ref{eq:gradrk}). This gradient comprises two alternative expressions of CMI: the first, $\overset{\rightarrow}{\hat{I}^{(\epsilon_{k-1})}}(\rvr^{(k - 1)},\rvr^{(k)})[t]$ represents the CMI with the preceding layer, incorporating the forward prediction error for estimating $\rvr^{(k)}$. The second,  $\overset{\leftarrow}{\hat{I}^{(\epsilon_{k})}}(\rvr^{(k)},\rvr^{(k+1)})[t]$, signifies the CMI with the subsequent layer, encompassing the backward prediction error for forecasting $\rvr^{(k)}$. These carefully selected expressions were pivotal in realizing a canonical network model that is free from the weight symmetry issue.

To underscore the significance of our selection, let's examine an alternate form of  (\ref{eq:gradrk}):
\begin{eqnarray}
\nabla_{\rvr^{(k)}}\hat{J}_k(\rvr^{(k)})[t]&=&\nabla_{\rvr^{(k)}}\overset{\rightarrow}{\hat{I}^{(\epsilon_{k-1})}}(\rvr^{(k - 1)},\rvr^{(k)})[t]+\nabla_{\rvr^{(k)}}\overset{\rightarrow}{\hat{I}^{(\epsilon_k)}}(\rvr^{(k)},\rvr^{(k+1)})[t] \nonumber\\
&&\hspace*{-1.4in}= \frac{1}{2}\nabla_{\rvr^{(k)}} (\log \det (\hat{\rmR}_{\rvr^{(k)}}[t]+ \epsilon_{k-1} \mI)
 -\frac{1}{\epsilon_{k-1}}\overset{\rightarrow}{\rve}^{(k)}[t]-\frac{1}{\epsilon_k}{\mW_{ff}^{(k)}}^T\overset{\rightarrow}{\rve}^{(k+1)}[t]. \label{eq:gradrkalt}
\end{eqnarray}
In this case, both the CMI expressions are founded on forward prediction errors. The gradient expression in (\ref{eq:gradrkalt}) incorporates both the forward prediction error $\overset{\rightarrow}{\rve}^{(k)}[t]$ and the backpropagated forward prediction error $\overset{\rightarrow}{\rve}^{(k+1)}[t]$,  weighted by ${\mW_{ff}^{(k)}}^T$. Consequently, choosing (\ref{eq:gradrkalt}) would result in a derived network model displaying symmetric feedforward and feedback weights.

This condition is typically encountered in predictive coding-based supervised schemes, such as \citep{whittington2017approximation, song2020can}, where the loss function is formulated based on feedforward prediction errors, and the symmetric feedback path arises from calculating the gradient for this loss function. Recently, Golkar et al. \citep{golkar2022constrained} proposed a loss function  that involves two alternative feedforward prediction error elements for the same branch.  By coupling this with an upper bound optimization and a whitening assumption, they manage to sidestep the weight symmetry issue. In contrast, our CorInfoMax framework provides a natural solution to the weight transport problem encountered in predictive coding networks by carefully selecting alternative CMI expressions, thus eliminating the need for the whitening assumption.

\subsection{The analytical comparison of forward and backward prediction coefficients}
\label{sec:asymmetry2}
Regarding the correlative mutual information component of the proposed stochastic objective presented in (\ref{eq:stobjective1}), the optimal solutions for both feedforward and feedback weights are found as solutions to problems (\ref{eq:forwardpred}) and (\ref{eq:backwpred}) respectively. To be more specific, these expressions can be framed as follows:
\begin{eqnarray}
\mW_{ff,*}^{(k)}&=&\mR_{\rvr^{(k)}\rvr^{(k+1)}}^T(\mR_{\rvr^{(k)}}+\epsilon_k \mathbf{I})^{-1}, \label{eq:ffmmse}\\
    \mW_{fb,*}^{(k)}&=&\mR_{\rvr^{(k)}\rvr^{(k+1)}}(\mR_{\rvr^{(k+1)}}+\epsilon_k \mathbf{I})^{-1}. \label{eq:fbmmse}
\end{eqnarray}
Consequently, the condition $\mW_{ff,}^{(k)}={\mW_{fb,}^{(k)}}^T$ does not generally hold true. Symmetry might be anticipated in very specific scenarios - such as when layer activations are signed with a zero mean, and the autocorrelation matrices meet the whiteness condition, i.e., $\mR_{\rvr^{(k)}}=\sigma_r^2\mathbf{I}$ and $\mR_{\rvr^{(k+1)}}=\sigma_r^2\mathbf{I}$. This analysis only considers the mutual information maximization component of the objective, yet it offers insight into the expected asymmetry of the forward and backward weights. 

While this analysis focuses solely on the mutual information maximization component of the objective, it still provides valuable insight into the anticipated asymmetry of the forward and backward weights.

\subsection{Empirical angle between forward and backward weights}
\label{sec:angle}
To test the level of symmetry between the forward prediction synaptic weight matrix $\mW_{ff}^{(k)}$ and the transpose of the backward prediction synaptic weight matrix $\mW_{fb}^{(k)}$,  we investigate the cosine angle between them. This measure is calculated using \citep{lillicrap2016random}:
\begin{eqnarray}
\label{eq:angle}
\Theta^{(k)}=\arccos\left(\frac{\text{Tr}\left(\mW_{ff}^{(k)}\mW_{fb}^{(k)}\right)}{\|\mW_{ff}^{(k)}\|_F\|\mW_{fb}^{(k)}\|_F}\right).
\end{eqnarray}
This metric serves as an indicator of alignment - with a cosine angle of $\Theta^{(k)}=0$ degrees signifying perfect symmetry, and $\Theta^{(k)}=90$ degrees indicating orthogonality and, thus, a significant degree of asymmetry. To provide a clearer illustration of the asymmetry between feedforward and feedback weights for the proposed CorInfoMax networks, we carried out a numerical evaluation of this metric in the context of the experiments documented in Table \ref{tab:ImageClassificationTest}.  It's worth noting that in these experiments, the feedforward and feedback weights were independently initialized with random values.

\begin{figure}[ht!]
\centering
\includegraphics[width=9.0cm, trim=0.0cm 0cm 0cm 0.0cm,clip]{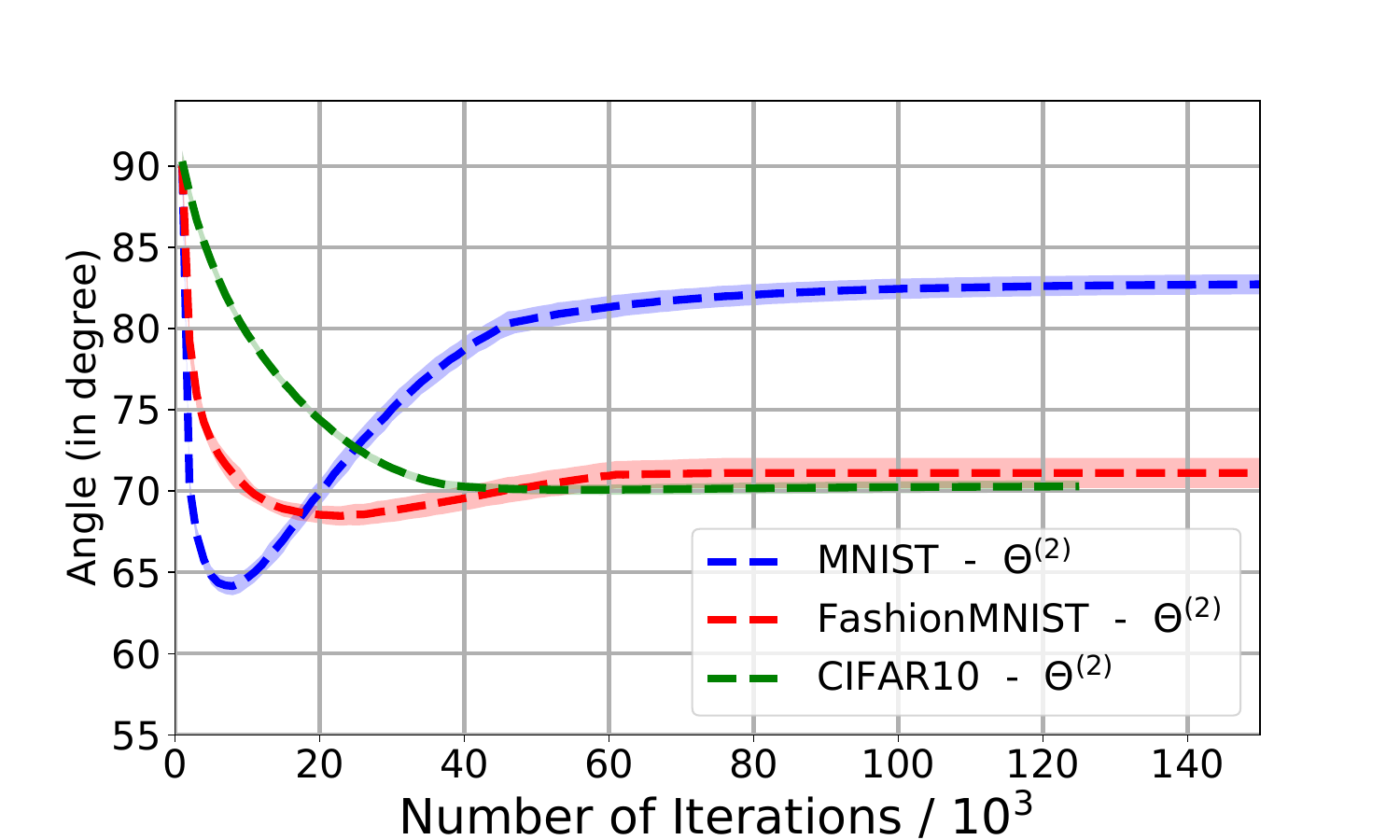} 
	\caption{The angle between the feedforward and the transpose of the feedback weights between hidden and output layers (averaged over $n=10$ runs associated with the corresponding $\pm$ std envelopes) as a function of weight update iterations for CorInfoMax-$\mathcal{B}_{\infty, +}$. }
	\label{fig:CorInfoWeightAngle}
\end{figure}
Figure \ref{fig:CorInfoWeightAngle} presents the experimental alignment results as a function of iterations for the MNIST, Fashion MNIST, and CIFAR10 datasets. According to this figure, the curves initially start at $90$ degrees due to the independent random selection of initial weights. As the iterations progress, the angle does indeed lessen, however, the steady-state levels largely persist, suggesting a noteworthy degree of asymmetry between the feedforward and feedback weights.


\section{Alternative expression for the feedforward weight updates}
\label{sec:altweightupdates}
In Section \ref{sec:learningDynamics}, we derived the partial derivative of the forward prediction cost with respect to the feedforward weights as follows:
    \begin{eqnarray}
    \frac{\partial C_{ff}(\mW^{(k)}_{ff}[t])}{\partial \mW^{(k)}_{ff}}=2\epsilon_k\mW_{ff}^{(k)}[t]-2\overset{\rightarrow}{\rve}^{(k+1)}[t]\rvr^{(k)}[t]^T. \label{appeq:partialforw}
    \end{eqnarray}
In this equation, $\overset{\rightarrow}{\rve}^{(k+1)}[t]$ symbolizes the feedforward error value at the point when system dynamics have stabilized.   In order to characterize this steady state value, we look at the optimal value of the optimization problem  employed to derive the system dynamics:

\begin{eqnarray}
    \underset{\rvr^{(k)}[t]}{\text{maximize }} & \Bigg( \Bigg.\frac{1}{2}(\log \det (\hat{\rmR}_{\rvr^{(k)}}[t]+ \epsilon_{k-1} \mI)+\log \det (\hat{\rmR}_{\rvr^{(k)}}[t]+ \epsilon_{k} \mI))\nonumber\\  &-\frac{1}{2\epsilon_{k-1}}\left\|\overset{\rightarrow}{\rve}^{(k)}[t]\right\|_2^2-\frac{1}{2\epsilon_k}\left\|\overset{\leftarrow}{\rve}^{(k)}[t]\right\|_2^2\Bigg. \Bigg)\label{eq:objsysdynamics}\\
    \text{subject to} &  \mathbf{0}\preccurlyeq \rvr^{(k)}[t]\preccurlyeq \mathbf{1}.\nonumber
\end{eqnarray}
We formulate the corresponding Lagrangian function for this optimization as
\begin{eqnarray}
L(\rvr^{(k)}[t], \mathbf{q}_1[t],\mathbf{q}_2[t])=O(\rvr^{(k)}[t])+\mathbf{q}_1[t]^T(\rvr^{(k)}[t])+\mathbf{q}_2[t]^T(\mathbf{1}-\rvr^{(k)}[t])
\end{eqnarray}
where $O(\rvr^{(k)}[t])$  corresponds to the objective in(\ref{eq:objsysdynamics}), and $\mathbf{q}_1[t],\mathbf{q}_2[t]\succcurlyeq \mathbf{0}$. Applying the KKT optimality conditions \citep{boyd2004convex}, we find
\begin{eqnarray}
    \nabla_{\rvr^{(k)}}L(\rvr_*^{(k)}[t], \mathbf{q}_{1*}[t],\mathbf{q}_{2*}[t])=\mathbf{0},
\end{eqnarray}
at the optimal point $(\rvr_*^{(k)}[t], \mathbf{q}_{1*},\mathbf{q}_{2*}[t])$. This results in
\begin{eqnarray}
    2{g_{B,k}}\overset{\rightarrow}{\rve}_*^{(k+1)}[t]+{2}\left(g_{A,k}\rvv^{(k)}_A[t]-(g_{lk}+g_{A_k})\rvr_*^{(k)}[t]\right)+\mathbf{q}_{1*}[t]+\mathbf{q}_{2*}[t]=\mathbf{0},
\end{eqnarray}
where  $\mathbf{q}_{1*}[t]$ and $\mathbf{q}_{2*}[t]$ are entirely null except at the indices where $\rvr^{(k)}_*[t]$ equals $0$ or $1$, owing to the nonnegative-clipping operation of $\sigma_+(\rvu^{(k)}_*[t])$.

Thus, by defining $\mathbf{q}_*[t]=\mathbf{q}_{1*}[t]+\mathbf{q}_{2*}[t]$,  the outer product term in (\ref{appeq:partialforw}) used in the feedforward weight update can be equivalently expressed as
    \begin{eqnarray}
-2\overset{\rightarrow}{\rve}^{(k+1)}[t]\rvr^{(k)}[t]^T=\frac{2}{g_{B,k}}\left(g_{A,k}\rvv^{(k)}_A[t]-(g_{lk}+g_{A_k})\rvr_*^{(k)}[t]+\frac{1}{2}\mathbf{q}_*[t]\right)\rvr^{(k)}[t]^T. \label{eq:outpW}
\end{eqnarray}
Based on $\rvr^{(k)}_*[t]=\sigma_+(\rvu^{(k)}_*[t])$, we can express the relationship between $\rvr^{(k)}_*[t]$ and $\rvu^{(k)}_*[t]$ as $\rvr^{(k)}_*[t]=\rvu^{(k)}_*[t]+\mathbf{d}[t]$, where $\mathbf{d}[t]$ has non-zero values only at the indices $\rvu^{(k)}$ is negative or exceeds $1$. Consequently, we can rewrite the outer product term in (\ref{eq:outpW}) as
    \begin{eqnarray}
-2\overset{\rightarrow}{\rve}^{(k+1)}[t]\rvr^{(k)}[t]^T=\frac{2}{g_{B,k}}\left(g_{A,k}\rvv^{(k)}_A[t]-(g_{lk}+g_{A_k})\rvu_*^{(k)}[t]+\mathbf{h}_*[t]\right)\rvr^{(k)}[t]^T \label{eq:outpW2},
\end{eqnarray}
where $\mathbf{h}_*[t]=\frac{1}{2}\mathbf{q}_*[t]-(g_{lk}+g_{A_k})\mathbf{d}[t]$.

Ultimately, following a similar approach as in \cite{golkar2022constrained}, we infer that the update expression for the feedforward term can be derived from the difference between the soma's membrane voltages and the distal apical compartments. This conclusion aligns with experimental observations that emphasize the dependence of basal synaptic plasticity on apical calcium plateau potentials \cite{larkum2013cellular}.

\section{Employing sparsity assumption on neuronal activities}
\label{sec:sparseNetwork}
In this section, we elaborate on the structure of the CorInfoMax network and the corresponding neuronal dynamics  that emerge with the selection of the activation domain  $\displaystyle \Pcal^{(k)}=\mathcal{B}_{1,+}=\{\rvr: \|\rvr\|_1 \leq 1, \vzero \preccurlyeq \rvr \}$.  It's important to note that  this domain corresponds to the intersection of the $\ell_1$-norm ball and the nonnegative orthant.

To derive the network dynamics corresponding to $\rvr^{(k)}[t]$, we consider the following constrained optimization similar to (\ref{eq:objsysdynamics}),

\begin{eqnarray}
    \underset{\rvr^{(k)}[t]}{\text{maximize }} & \Bigg( \Bigg.\frac{1}{2}(\log \det (\hat{\rmR}_{\rvr^{(k)}}[t]+ \epsilon_{k-1} \mI)+\log \det (\hat{\rmR}_{\rvr^{(k)}}[t]+ \epsilon_{k} \mI))\nonumber\\  &-\frac{1}{2\epsilon_{k-1}}\left\|\overset{\rightarrow}{\rve}^{(k)}[t]\right\|_2^2-\frac{1}{2\epsilon_k}\left\|\overset{\leftarrow}{\rve}^{(k)}[t]\right\|_2^2\Bigg. \Bigg)\label{eq:objsysdynamicsSparse}\\
    \text{subject to} &  \|\rvr^{(k)}[t]\|_1 \le 1,\nonumber\\
    &   \mathbf{0}\preccurlyeq \rvr^{(k)}[t].\nonumber
\end{eqnarray}
We can write down the Lagrangian min-max problem corresponding to this optimization as 
\begin{eqnarray}
\underset{{q}^{(k)}[t] \ge 0}{\text{minimize }} 
 \underset{\rvr^{(k)}[t] \succcurlyeq \vzero}{\text{maximize }}  L(\rvr^{(k)}[t], {q}^{(k)}[t])=O(\rvr^{(k)}[t])+{q}^{(k)}[t](1 - \|\rvr^{(k)}[t]\|_1).
\end{eqnarray}
Here $O(\rvr^{(k)}[t])$  signifies the objective in (\ref{eq:objsysdynamicsSparse}).  By applying the proximal gradient update \citep{parikh2014proximal} for $\displaystyle \rvr^{(k)}[t]$ using the gradient expression in (\ref{eq:gradrk}), we can represent the output dynamics for layer-$k$ as
  
\begin{eqnarray}
    \tau_{\rvu}\frac{d \rvu^{(k)}[t;s]}{ds}&=&-g_{lk}\rvu^{(k)}[t;s]+\frac{1}{\epsilon_k}\mM^{(k)}[t]\vr^{(k)}[t;s]-\frac{1}{\epsilon_{k-1}}\overset{\rightarrow}{\rve}_u^{(k)}[t;s]-\frac{1}{\epsilon_{k}}\overset{\leftarrow}{\rve}_u^{(k)}[t;s], \nonumber\\
   \overset{\rightarrow}{\rve}_u^{(k)}[t;s]&=&\rvu^{(k)}[t;s]-\mW^{(k-1)}_{ff}[t]\rvr^{(k-1)}[t;s], \nonumber\\
\overset{\leftarrow}{\rve}_u^{(k)}[t;s]&=&\rvu^{(k)}[t;s]-\mW^{(k)}_{fb}[t]\rvr^{(k+1)}[t;s],\nonumber\\
\rvr^{(k)}[t;s]&=& \text{ReLU}(\rvu^{(k)}[t;s] - q^{(k)}[t;s] \vone). \nonumber
\end{eqnarray} 
In this formulation, we introduce the intermediate variable  $\displaystyle \rvu^{(k)}$, where $\text{ReLU}$ represents the element-wise rectified linear unit. We then define the apical and basal potentials as
\begin{eqnarray}
    \rvv^{(k)}_A[t;s]&=& \mM^{(k)}[t]\vr^{(k)}[t;s]+\mW^{(k)}_{fb}[t]\rvr^{(k+1)}[t;s] -\frac{1}{g_{A,k}} {q}^{(k)}[t;s]\vone \nonumber, \\
    \rvv^{(k)}_B[t;s]&=& \mW^{(k-1)}_{ff}[t]\rvr^{(k-1)}[t;s]. \nonumber
\end{eqnarray}
With these potentials in place, we can restate the output dynamics in a format similar to (\ref{eq:hiddynamics1}) and (\ref{eq:hiddynamics2}),
\begin{align}
    &\tau_{\rvu}\frac{d \rvu^{(k)}[t;s]}{ds}=-g_{lk}\rvu^{(k)}[t;s]+g_{A,k}(\rvv^{(k)}_A[t;s]-\rvu^{(k)}[t;s])+g_{B,k}(\rvv^{(k)}_B[t;s]-\rvu^{(k)}[t;s]), \label{eq:hiddynamicsSparse1}
    \\
  &\rvr^{(k)}[t;s]= \text{ReLU}(\rvu^{(k)}[t;s]). \label{eq:hiddynamicsSparse2}
  \end{align}
For the Lagrangian variable  $q^{(k)}[t;s]$,  we can derive the update based on the dual minimization as follows,
\begin{align}
    \frac{d a^{(k)}[t;s]}{ds} = -a^{(k)}[t;s] + \sum_{j = 1}^{N_k} \rvr_j^{(k)}[t;s] - 1 + q^{(k)}[t;s], \quad q^{(k)}[t;s] = \text{ReLU}(a^{(k)}[t;s]). \label{eq:InhibitoryDynamics}
\end{align}
In the above formulation, the Lagrangian variable $\displaystyle q^{(k)}$ equates to an inhibitory inter-neuron that receives input from all neurons of layer-$k$ and generates an inhibitory signal to the apical compartments of all these neurons.


\section{Supplementary on numerical experiments}
\label{sec:suppNumericalExperiments}

In the forthcoming subsections, we delve into the specifics of our experimental setup, which covers everything from hyperparameters to an in-depth analysis. We utilize four image classification tasks to evalutate the effectiveness of our proposed framework:  MNIST \citep{lecunMNIST}, Fashion MNIST \citep{xiao2017fashionmnist}, CIFAR10, and CIFAR100 \citep{krizhevsky2009learning}. These image datasets are all acquired from the Pytorch library \citep{paszke2017automatic}.

The MNIST dataset includes $60000$ grayscale training images of hand-drawn digits and $10000$ test images, each with a dimension of $28 \times 28$. The Fashion MNIST dataset mirrors the MNIST in terms of format, size, and quantity of training and test images. However, instead of digits, it features images of clothing items, divided into $10$ categories.

On a larger scale, we have the CIFAR10 and CIFAR100 datasets, which contain $32\times32$ RGB images. CIFAR10 consists of $50000$ training images and $10000$ test images, each associated with one of $10$ object labels. CIFAR100, while maintaining the same quantity of training and test images as CIFAR10, is distinguished by its $100$ object categories.

We initially map the image pixels to the range $[0, 1]$ by rescaling them by $255$. For the CIFAR10 and CIFAR100 datasets, normalization is performed using the mean and standard deviation values cited in the published codes of \citep{ScalingEP, laborieux2022EPholomorphic}. To accommodate our training of MLP architectures, the images are flattened prior to being fed into the neural networks. Lastly, we have opted not to employ any augmentation in our numerical experiments.

\subsection{On the computation of neural dynamics}
The neural dynamics equations defined by (\ref{eq:hiddynamics1})-(\ref{eq:hiddynamics2}) implicitely determines the input-output relations of the CorInfoMax-$\mathcal{B}_{\infty, +}$ network for each segments. We obtain the solution of these equations by applying the discrete-time method that is provided in Algorithm \ref{alg:CorInfoMaxneuraldynamiciterationsAntisparse}. Within this algorithm, $\mu_\rvu[s]$ refers to the learning rate of the neural dynamics during the gradual time scale indicated by index $s$. Additionally, $s_{\text{max}}$ stands for the highest count of iterations in the loop-driven calculations of neural dynamics.

\begin{algorithm}[H]
\begin{algorithmic}[1]
\STATE Initialize $\rvr^{(k)}[t;1]$, $\rvu^{(k)}[t;1]$, $\mu_\rvu[1]$, $s_{\text{max}}$, and $s = 1$
\WHILE{$s < s_{\text{max}}$}
\FOR{$k = 1, \ldots, P$}
\STATE $\tau_{\rvu}\frac{d \rvu^{(k)}[t;s]}{ds}=-g_{lk}\rvu^{(k)}[t;s]+g_{A,k}(\rvv^{(k)}_A[t;s]-\rvu^{(k)}[t;s])+g_{B,k}(\rvv^{(k)}_B[t;s]-\rvu^{(k)}[t;s])$
\STATE $\rvu^{(k)}[t;s+1] = \rvu^{(k)}[t;s] + \mu_\rvu[s] \tau_{\rvu}\frac{d \rvu^{(k)}[t;s]}{ds}$
\STATE $\rvr^{(k)}[t;s + 1]=\sigma_+(\rvu^{(k)}[t;s + 1])$
\ENDFOR
\STATE $s = s +1$, and adjust $\mu_\rvu[s]$ if necessary.
\ENDWHILE
\end{algorithmic}
\caption{CorInfoMax neural dynamic iterations: $\Pcal^{(k)} = \mathcal{B}_{\infty, +}$}
\label{alg:CorInfoMaxneuraldynamiciterationsAntisparse}
\end{algorithm}

Similarly, the equations defined by (\ref{eq:hiddynamicsSparse1}), (\ref{eq:hiddynamicsSparse2}), and (\ref{eq:InhibitoryDynamics} implicitely determines the input-output relations of the CorInfoMax-$\mathcal{B}_{1,+}$ network for each segment with an additional inhibition signal $q^{(k)}$. In an analogous manner, we use the following discrete-time method outlined in Algorithm \ref{alg:CorInfoMaxneuraldynamiciterationsSparse} to obtain the solution of these equations. In contrast to Algorithm \ref{alg:CorInfoMaxneuraldynamiciterationsAntisparse}, Algorithm \ref{alg:CorInfoMaxneuraldynamiciterationsSparse} introduces an extra hyperparameter denoted as $\mu_a[s]$. This hyperparameter governs the learning rate of supplementary inhibitory neurons $q^{(k)}$. Detailed discussions concerning the specific values of each hyperparameter related to the neural dynamics for individual tasks are presented in the subsequent subsections.

\begin{algorithm}[H]
\begin{algorithmic}[1]
\STATE Initialize $\rvr^{(k)}[t;1]$, $\rvu^{(k)}[t;1]$, $a^{(k)}[t;1]$, $q^{(k)}[t;1]$, $\mu_\rvu[1]$, $\mu_a[1]$, $s_{\text{max}}$, and $s = 1$
\WHILE{$s < s_{\text{max}}$}
\FOR{$k = 1, \ldots, P$}
\STATE $\tau_{\rvu}\frac{d \rvu^{(k)}[t;s]}{ds}=-g_{lk}\rvu^{(k)}[t;s]+g_{A,k}(\rvv^{(k)}_A[t;s]-\rvu^{(k)}[t;s])+g_{B,k}(\rvv^{(k)}_B[t;s]-\rvu^{(k)}[t;s])$
\STATE $\frac{d a^{(k)}[t;s]}{ds} = -a^{(k)}[t;s] + \sum_{j = 1}^{N_k} \rvr_j^{(k)}[t;s] - 1 + q^{(k)}[t;s]$
\STATE $\rvu^{(k)}[t;s+1] = \rvu^{(k)}[t;s] + \mu_\rvu[s] \tau_{\rvu}\frac{d \rvu^{(k)}[t;s]}{ds}$
\STATE $\rvr^{(k)}[t;s + 1]=\text{ReLU}(\rvu^{(k)}[t;s + 1])$
\STATE $a^{(k)}[t; s + 1] = a^{(k)}[t;s]) + \mu_a[s]\frac{d a^{(k)}[t;s]}{ds}$
\STATE $q^{(k)}[t;s + 1] = \text{ReLU}(a^{(k)}[t; s + 1])$
\ENDFOR
\STATE $s = s +1$, and adjust $\mu_\rvu[s]$ and $\mu_a[s]$ if necessary.
\ENDWHILE
\end{algorithmic}
\caption{CorInfoMax neural dynamic iterations: $\Pcal^{(k)} = \mathcal{B}_{1, +}$}
\label{alg:CorInfoMaxneuraldynamiciterationsSparse}
\end{algorithm}

\subsection{Learning Dynamics}

We outline the comprehensive learning dynamics of our proposed framework in Algorithm \ref{alg:CorInfoMaxLearningDynamicsAlgo}. As we mentioned earlier, we adopt the Equilibrium Propagation technique \cite{EqProp}, leading our approach through two distinct phases of neural dynamics prior to weight updates: i) the free phase, and ii) the nudged phase. Initially, we execute the free phase by setting $\beta = 0$. Subsequently, we proceed with the nudged phase, activating the neural dynamics with $\beta = \beta' > 0$. The adjustments to the feedforward and feedback weights are then determined by contrasting neural activities between the nudged and free phases. Concerning the update equation for the forward weights in (\ref{eq:forwardEPupdate}), the error vector $\overset{\rightarrow}{\rve}^{(k+1)}[t]$ evaluated at $(\overset{\rightarrow}{\rve}^{(k+1)}[t]\rvr^{(k)}[t]^T)\big|_{\beta=\beta'}$ corresponds to $\overset{\rightarrow}{\rve}^{(k)}[t]_{\beta=\beta'}=\rvr^{(k)}[t]_{\beta=\beta'}-\mW^{(k-1)}_{ff}[t]\rvr^{(k-1)}[t]_{\beta=\beta'}$. Similarly, for the update expression of the backward weights in (\ref{eq:backwardEPupdate}), the error vector $\overset{\leftarrow}{\rve}^{(k)}[t]$ at $(\overset{\leftarrow}{\rve}^{(k)}[t]\rvr^{(k+1)}[t]^T)\big|_{\beta=\beta'}$ corresponds to $\overset{\leftarrow}{\rve}^{(k)}[t]\big|_{\beta=\beta'}=\rvr^{(k)}[t]\big|_{\beta=\beta'}-\mW^{(k)}_{fb}[t]\rvr^{(k+1)}[t]\big|_{\beta=\beta'}$. In Algorithm \ref{alg:CorInfoMaxLearningDynamicsAlgo}, the learning rates for the forward and backward weights are denoted as $\mu_{\text{ff}}$ and $\mu_{\text{fb}}$, respectively. Lateral weight updates adhere to the learning rule outlined in \cite{bozkurt2023correlative}, and we rewrite them in terms of autapses and lateral inhibition synapses (Appendix \ref{sec:lateralWeightUpdates}). These updates are performed based on the neural activities subsequent to the nudged phase.
\begin{algorithm}[H]
\begin{algorithmic}[1]
\STATE Initialize network parameters: $\mW^{(k)}_{ff}[1]$, $\mW^{(k)}_{fb}[1]$, $\mB^{(k)}[1]$, $\epsilon_k$, (for each $k$), $\lambda_\rvr$, $g_{lk}$
\STATE Initialize hyperparameters: $\beta'$, $T_{\text{free}}$, $T_{\text{nudged}}$, $\mu_{\text{ff}}$, $\mu_{\text{fb}}$, $\mu_\rvu$, $\mu_a$ (if $\Pcal = \mathcal{B}_{1,+}$)
\FOR{t = 1, 2, \ldots}
\STATE \textbf{Run neural dynamics with $\beta = 0$ for} $s_{\text{max}} = T_{\text{free}}$ to collect $\rvr^{(k)}[t]\big|_{\beta=0}$ ( $k = 1, \ldots P$)
\STATE \textbf{Run neural dynamics with $\beta = \beta'$ for} $s_{\text{max}} = T_{\text{nudged}}$ to collect $\rvr^{(k)}[t]\big|_{\beta=\beta'}$ ( $k = 1, \ldots P$)
\STATE Update synaptic weights with (\ref{eq:forwardEPupdate}) and (\ref{eq:backwardEPupdate}) for each $k$:
\begin{align*}
    \mW^{(k)}_{ff}[t + 1] &= \mW^{(k)}_{ff}[t] + \mu_{\text{ff}}
   \frac{1}{\beta'}\left((\overset{\rightarrow}{\rve}^{(k+1)}[t]\rvr^{(k)}[t]^T)\bigg|_{\beta=\beta'}-(\overset{\rightarrow}{\rve}^{(k+1)}[t]\rvr^{(k)}[t]^T)\bigg|_{\beta=0}\right)\\
   \mW^{(k)}_{fb}[t + 1] &= \mW^{(k)}_{fb}[t] + \mu_{\text{fb}}
    \frac{1}{\beta'}\left((\overset{\leftarrow}{\rve}^{(k)}[t]\rvr^{(k+1)}[t]^T)\bigg|_{\beta=\beta'}-(\overset{\leftarrow}{\rve}^{(k)}[t]\rvr^{(k+1)}[t]^T)\bigg|_{\beta=0}\right)
\end{align*}
\STATE Update lateral weights for each $k$ (see Appendix \ref{sec:lateralWeightUpdates}):
\begin{align*}
    \mathbf{D}_{ii}^{(k)}[t+1] &= \lambda_\mathbf{r}^{-1} \mathbf{D}_{ii}^{(k)}[t] - \lambda_\mathbf{r}^{-1} \epsilon_k2\gamma^2(\mathbf{z}_i^{(k)}[t])^2 + \epsilon_k g_{lk}(1 - \lambda_\rvr^{-1}), \quad \forall i \in \{1, \ldots, N_k\}\\
    \mathbf{O}_{ij}^{(k)}[t+1] &= \lambda_\mathbf{r}^{-1} \mathbf{O}^{(k)}_{ij}[t] + \lambda_\mathbf{r}^{-1} \epsilon_k 2 \gamma^2 \mathbf{z}_i^{(k)}[t] \mathbf{z}_j^{(k)}[t], \ \  \forall i,j \in \{1, \ldots, N_k\}, \text{ where } i\neq j
\end{align*}
\ENDFOR
\end{algorithmic}
\caption{CorInfoMax network learning dynamics}
\label{alg:CorInfoMaxLearningDynamicsAlgo}
\end{algorithm}

\subsection{Description of hyperparameters}
In the upcoming sections, we present the hyperparameters and their effects in our experiments. Therefore, Table \ref{hyperparamDescriptions} describes notation for the hyperparameters used in CorInfoMax neural dynamics and learning updates.

\begin{table}[h!]
  \caption{Description of the hyperparameter notations.\newline}
  \label{hyperparamDescriptions}
  \centering
  \begin{tabular}{lll}
    \toprule
    Hyperparameter     & Description   \\
    \midrule
    Architecture & An array containing the dimension of each layer. \\
    $T_{\text{free}}$     & Number of neural dynamics iterations for the free phase.   \\
    $T_{\text{nudged}}$     & Number of neural dynamics iterations for the nudged phase.        \\
    $\mu_{\text{ff}}$     & An array containing the learning rates for feedforward weights.      \\
    $\mu_{\text{fb}}$     & An array containing the learning rates for feedback weights.         \\
    $\lambda_\rvr$     & Forgetting factor for sample the auto correlation matrices in (\ref{eq:weightedRr}).       \\
    $\epsilon_k$     & Perturbation coefficient for autocorrelation matrices in  (\ref{eq:sMIS1}) and (\ref{eq:sMIS2}).      \\
    $\beta'$     & Nudging parameter for the nudged neural dynamics.  \\
    $g_{lk}$     & Leak coefficient for the neural dynamics in (\ref{eq:hiddynamics2}) and (\ref{eq:hiddynamicsSparse2}).       \\
    $\mu_{\rvu}$     & Learning rate for the neural dynamics in Algorithm \ref{alg:CorInfoMaxneuraldynamiciterationsAntisparse} and \ref{alg:CorInfoMaxneuraldynamiciterationsSparse}.        \\
    $\mu_{a}$     & Learning rate for the inhibition neuron in Algorithm \ref{alg:CorInfoMaxneuraldynamiciterationsSparse}.        \\
    lr decay     & Learning rate decay that multiplies the $\mu_{\text{ff}}$ and  $\mu_{\text{fb}}$ after each epoch.       \\
    \bottomrule
  \end{tabular}
\end{table}

\subsection{Two layer CorInfoMax-\texorpdfstring{$\mathcal{B}_{\infty,+}$}{Lg} network}
\label{sec:antisparse1}
In this section, we provide supplementary experimental results for the CorInfoMax-$\mathcal{B}_{\infty,+}$ network with a single hidden layer. 

\subsubsection{Network architecture}
\begin{figure}[h]
\includegraphics[width=0.98\textwidth, trim={5.5cm, 12.0cm, 1.8cm, 7.0cm},clip]{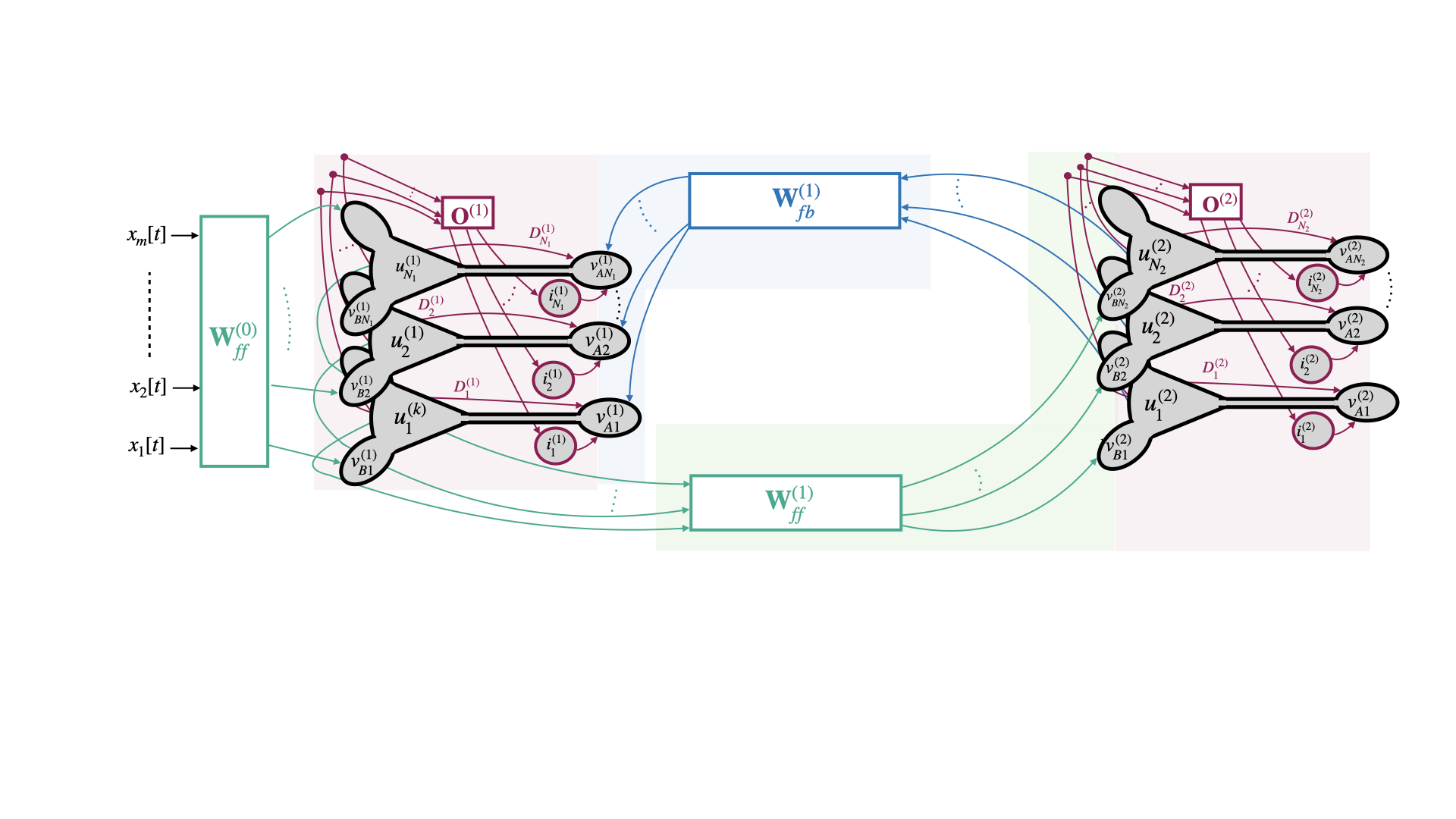}
\newline\caption{Two layer correlative information maximization based neural network with non-negative antisparsity constraint: $\mathbf{0}\preccurlyeq \rvr^{(l)}\preccurlyeq \mathbf{1}, l=1,2$.} 
\label{fig:twolayercorinfomaxNNAntiSparse}
\end{figure}

Figure \ref{fig:twolayercorinfomaxNNAntiSparse} provides a depiction of a CorInfoMax network with a single hidden layer. In this instance, both the hidden and output layers have the same constraint set $\mathcal{P}=\mathcal{B}_{\infty,+}$.


\subsubsection{Hyperparameters}
Table \ref{CorInfoOneHL_Hyperparams} provides the hyperparameters to train $2$-layer CorInfoMax networks on MNIST, Fashion MNIST, and CIFAR10, for which the corresponding test accuracy results are provided in Section \ref{sec:numericalexperiments} and in Appendix \ref{sec:CorInfoMaxBinf2layerTestFigures}.
\begin{table}[h!]
  \caption{Hyperparameters used to train CorInfoMax-$\mathcal{B}_{\infty, +}$ networks with single hidden layer. (In the row stating the lr decay, ep. and O/W means \textit{epoch} and \textit{otherwise}, respectively.)\newline}
  \label{CorInfoOneHL_Hyperparams}
  \centering
  \begin{tabular}{llll}
    \toprule
    Hyperparameter     & MNIST     & Fashion-MNIST & CIFAR10  \\
    \midrule
    Batch size & $20$ & $20$ & $20$  \\
    Architecture & $[784, 500, 10]$ & $[784, 500, 10]$ & $[3072, 1000, 10]$ \\
    $T_{\text{free}}$ & $30$ & $30$ & $30$ \\
    $T_{\text{nudged}}$ & $10$ & $10$ & $10$ \\
    $\mu_{\text{ff}}$ & $[1.0, 0.7]$ & $[0.3, 0.22]$ & $[0.08, 0.04]$\\
    $\mu_{\text{fb}}$ & $[-, 0.15]$ & $[-, 0.07]$ & $[-, 0.04]$ \\
    $\lambda_\rvr$ & $1 - 10^{-5}$ & $1 - 10^{-5}$ & $1 - 5 \times 10^{-5}$ \\
    $\epsilon_k$ & $0.15 \quad \forall k$ & $0.15 \quad \forall k$ & $0.15 \quad \forall k$  \\
    $\beta'$ & $1.0$ & $1.0$ & $1.0$  \\
    $g_{lk}$ & $0.5$ & $0.3$ & $0.1$  \\
    $\mu_{\rvu}$  & $\text{max}\{\frac{0.05}{s \times 10^{-2} + 1}, 10^{-3} \}$ & $\text{max}\{\frac{0.07}{s \times 10^{-2} + 1}, 10^{-3} \}$ & $0.05$ \\
    lr decay &  $\left\{\begin{array}{cc}
                        0.95 & \text{ep.}< 15 \\
                        0.9 & \mbox{O/W}. \end{array}\right.$ & 
                        $\left\{\begin{array}{cc}
                        0.95 & \text{ep.}< 20 \\
                        0.9 & 20 \leq \text{ep.}< 25\\
                        0.8 & \mbox{O/W}. \end{array}\right.$&
                         $\left\{\begin{array}{cc}
                        0.95 & \text{ep.}< 15 \\
                        0.9 & \mbox{O/W}. \end{array}\right.$\\
    \bottomrule
  \end{tabular}
\end{table}

\newpage
\subsubsection{Test accuracy results}
\label{sec:CorInfoMaxBinf2layerTestFigures}
\begin{figure}[h!]
\centering
\includegraphics[trim = {0cm 0cm 0cm 1.0cm},clip,width=0.9\textwidth]{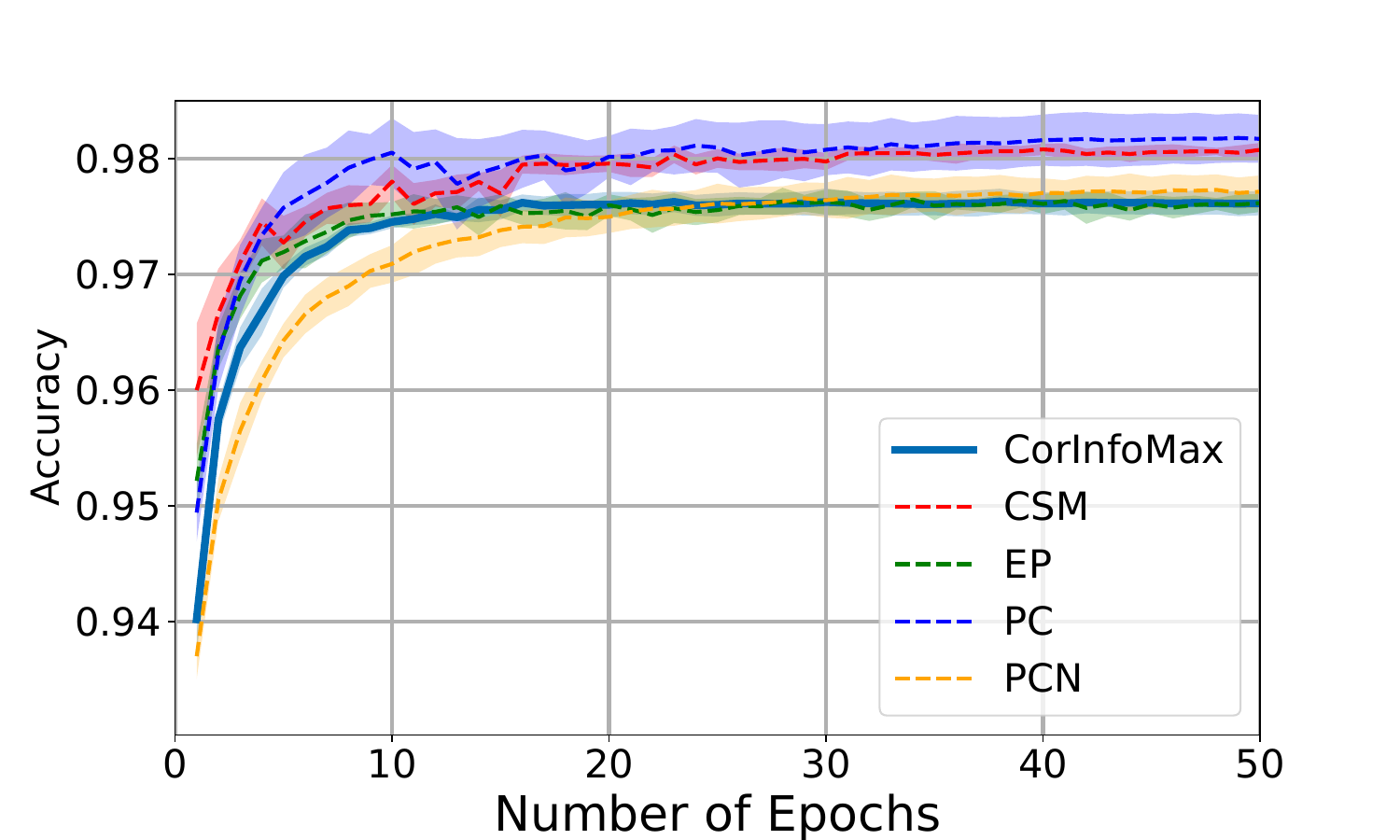}
\newline\caption{Test accuracy convergences of CorInfoMax-$\mathcal{B}_{\infty,+}$ networks and other (CSM, EP, PC, and PC-Nudge) algorithms as a function of epochs (averaged over $n=10$ runs associated with the corresponding $\pm$ std envelopes) for the MNIST dataset.}
\label{fig:MNISTTestComparison}
\end{figure}

Figure \ref{fig:MNISTTestComparison} compares  the test accuracy performance of CorInfoMax-$\mathcal{B}_{\infty,+}$ network (as a function of training epochs)  with CSM, EP, PC, and PC-Nudge algorithms for the MNIST dataset.

\begin{figure}[h!]
\centering
\includegraphics[trim = {0cm 0cm 0cm 1.0cm},clip,width=0.9\textwidth]{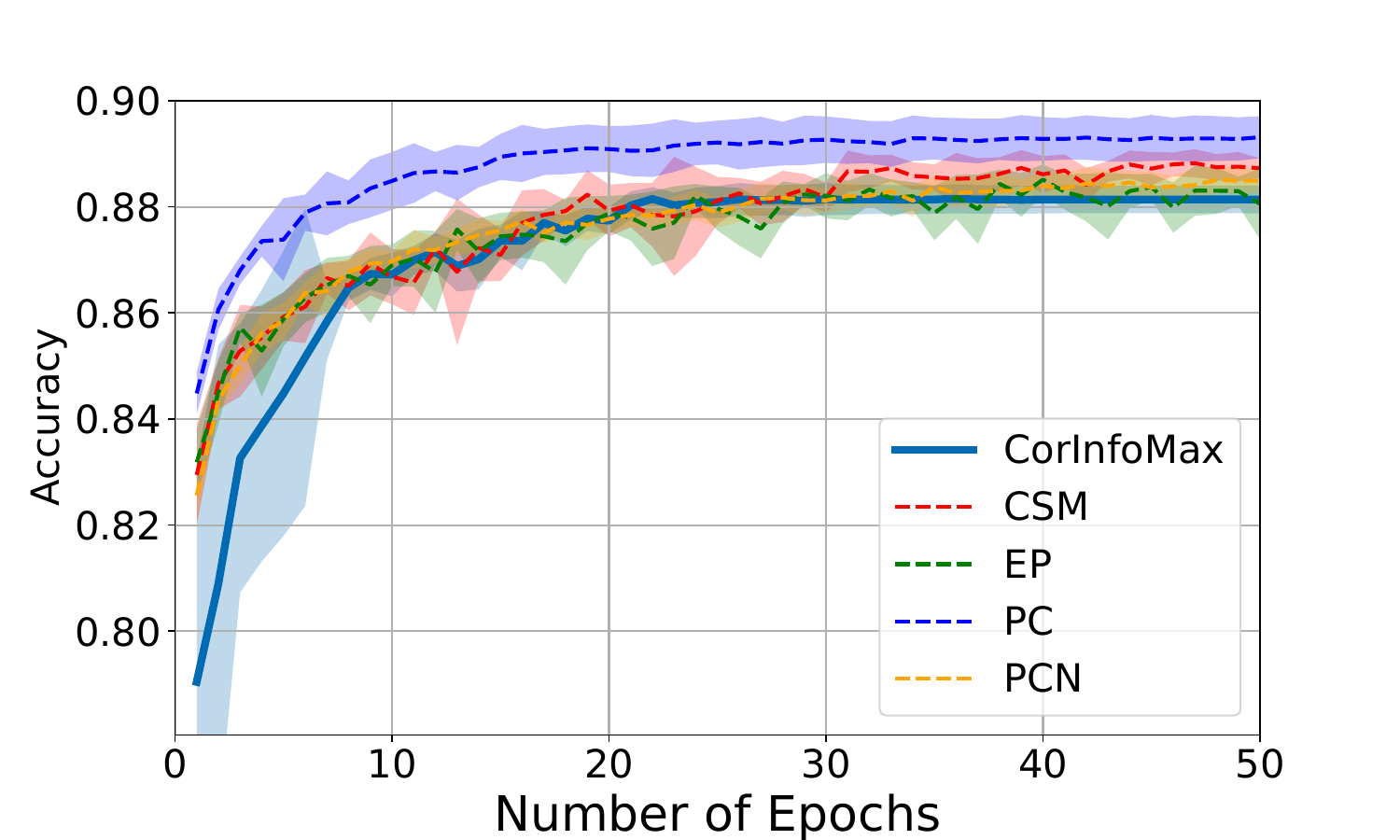}
\newline\caption{Test accuracy convergences of CorInfoMax-$\mathcal{B}_{\infty,+}$ networks and other (CSM, EP, PC, and PC-Nudge) algorithms as a function of epochs (averaged over $n=10$ runs associated with the corresponding $\pm$ std envelopes) for the Fashion MNIST dataset.}
\label{fig:FashionMNISTTestComparison}
\end{figure}

Figure \ref{fig:FashionMNISTTestComparison} compares  the test accuracy performance of CorInfoMax-$\mathcal{B}_{\infty,+}$ network (as a function of training epochs)  with CSM, EP, PC, and PC-Nudge algorithms for the Fashion MNIST dataset.

\begin{figure}[h!]
\centering
\includegraphics[trim = {0cm 0cm 0cm 1.0cm},clip,width=0.9\textwidth]{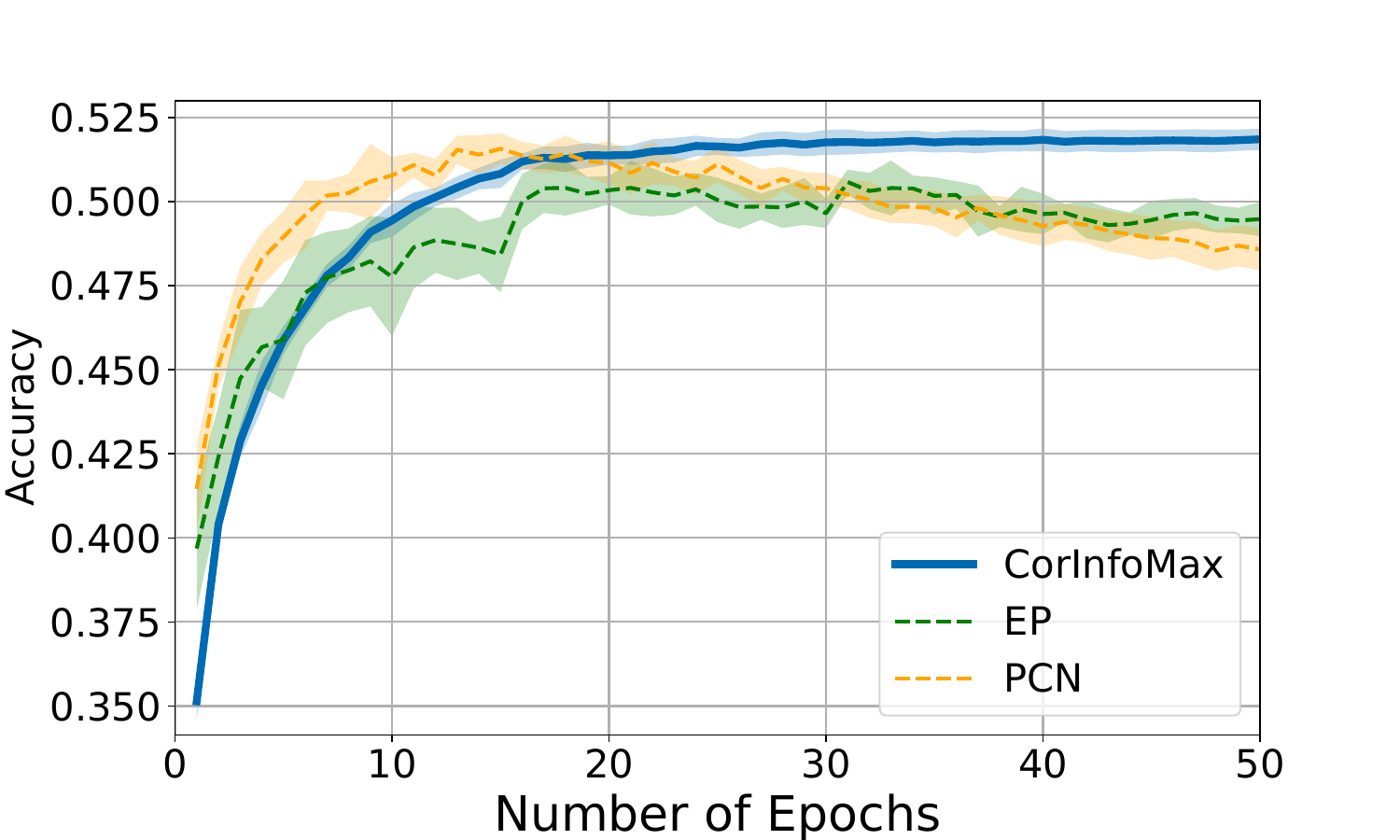}
\newline\caption{Test accuracy convergences of CorInfoMax-$\mathcal{B}_{\infty,+}$ networks and other (CSM, EP, PC, and PC-Nudge) algorithms as a function of epochs (averaged over $n=10$ runs associated with the corresponding $\pm$ std envelopes) for the CIFAR10 dataset.}
\label{fig:CIFARTestComparison}
\end{figure}

\newpage
Figure \ref{fig:CIFARTestComparison} compares  the test accuracy performance of CorInfoMax-$\mathcal{B}_{\infty,+}$ network (as a function of training epochs)  with CSM, EP, PC, and PC-Nudge algorithms for the CIFAR10 dataset.

\newpage
\subsection{Three layer CorInfoMax-\texorpdfstring{$\mathcal{B}_{\infty,+}$}{Lg} network}
\label{sec:antisparse3layer}
In this section, we consider the extension of the CorInfoMax-$\mathcal{B}_{\infty,+}$ network in Figure \ref{fig:twolayercorinfomaxNNAntiSparse}, obtained by inserting an additional hidden layer.

\subsubsection{Hyperparameters}
\begin{table}[h!]
  \caption{Hyperparameters used to train  CorInfoMax-$\mathcal{B}_{\infty,+}$ networks with two hidden layers.(In the row stating the lr decay, ep. and O/W means \textit{epoch} and \textit{otherwise}, respectively.)\newline}
  \label{CorInfoTwoHL_Hyperparams}
  \centering
  \begin{tabular}{llll}
    \toprule
    Hyperparameter     & MNIST   & CIFAR10 & CIFAR100 \\
    \midrule
    Batch size & $20$ & $20$ & $20$ \\
    Architecture & $[784, 500, 500, 10]$ & $[3072, 1000, 500, 10]$ & $[3072, 2000 , 1000, 100]$\\
    $T_{\text{free}}$ &  $30$ & $30$ & $50$ \\
    $T_{\text{nudged}}$ & $10$  & $10$ & $20$\\
    $\mu_{\text{ff}}$ & $[1.1, 0.75, 0.6]$ & $[0.11, 0.06, 0.035]$& $[0.18, 0.15, 0.09]$ \\
    $\mu_{\text{fb}}$ & $[-, 0.17, 0.07]$ & $[-, 0.045, 0.015]$ & $[-, 0.08, 0.06]$\\
    $\lambda_\rvr$ & $1 - 10^{-5}$ & $1 - 10^{-5}$ & $1 - 10^{-5}$ \\
    $\epsilon_k$ & $0.15 \quad \forall k$ & $0.15 \quad \forall k$ & $0.15 \quad \forall k$\\
    $\beta'$ & $1.0$ & $1.0$ & $1.0$\\
    $g_{lk}$ & $0.5$  & $0.1$ & $0.1$\\
    $\mu_{\rvu}$ & 0.05 & $0.05$ & $0.06$\\
    lr decay & $\left\{\begin{array}{cc}
                        0.95 & \text{ep.}< 15 \\
                        0.9 & \mbox{O/W}. \end{array}\right.$ &
                        $\left\{\begin{array}{cc}
                        0.95 & \text{ep.}< 15 \\
                        0.9 & \mbox{O/W}. \end{array}\right.$&
                        $\left\{\begin{array}{cc}
                        0.99 & \text{ep.}< 20 \\
                        0.9 & \mbox{O/W}. \end{array}\right.$\\
    \bottomrule
  \end{tabular}
\end{table}

\vspace{0.4in}
\subsubsection{Test accuracy results}
Table \ref{tab:ImageClassificationTestDeeperLayers} displays the test accuracy results for the 3-layer CorInfoMax-$\mathcal{B}_{\infty,+}$ networks on MNIST, CIFAR10 and CIFAR100 datasets in comparison with Feedback Alignment and Backpropagation algorihtms.
\begin{table}[h!]
  \caption{Test accuracy ($n=10$ runs $\pm$ stddev) of CorInfoMax-$\mathcal{B}_{\infty, +}$ networks with two hidden layers. \newline}
    \begin{adjustbox}{width=\columnwidth,center}
  \centering
  \begin{tabular}{llll}
    \toprule
         & MNIST  & CIFAR10 & \hspace{0.75cm}CIFAR100 \\
         \cmidrule(r){2-4}
         & Top-1 (\%) & Top-1 (\%) & Top-1 (\%) / Top5 (\%) \\
    \midrule
    \textbf{CorInfoMax}-$\mathcal{B}_{\infty, +}$ & $97.58\pm0.1$  & $50.97\pm0.4$& $20.84\pm0.4$ / $37.86\pm0.8$ \\
    \midrule
    Feedback Alignment (with MSE Loss) & $98.18\pm0.0$ & $50.26\pm1.4$ & - / -\\
    Feedback Alignment (with CrossEntropy Loss) & $97.96\pm0.2$ & $51.64\pm0.6$ & - / -\\
    BP (with MSE Loss) & $97.74\pm0.1$ & $55.49\pm0.4$ & $26.56\pm0.2$ / $40.64\pm0.4$\\
    BP (with CrossEntropy Loss) & $98.28\pm0.08$& $56.14\pm0.3$& $28.93\pm0.3$ / $54.13\pm0.4$\\
    \bottomrule
    \label{tab:ImageClassificationTestDeeperLayers}
  \end{tabular}
  \end{adjustbox}
\end{table}
\newpage

\begin{figure}[h!]
\subfloat[]
{\includegraphics[trim = {0cm 0cm 0cm 1.0cm},clip,width=0.325\textwidth]{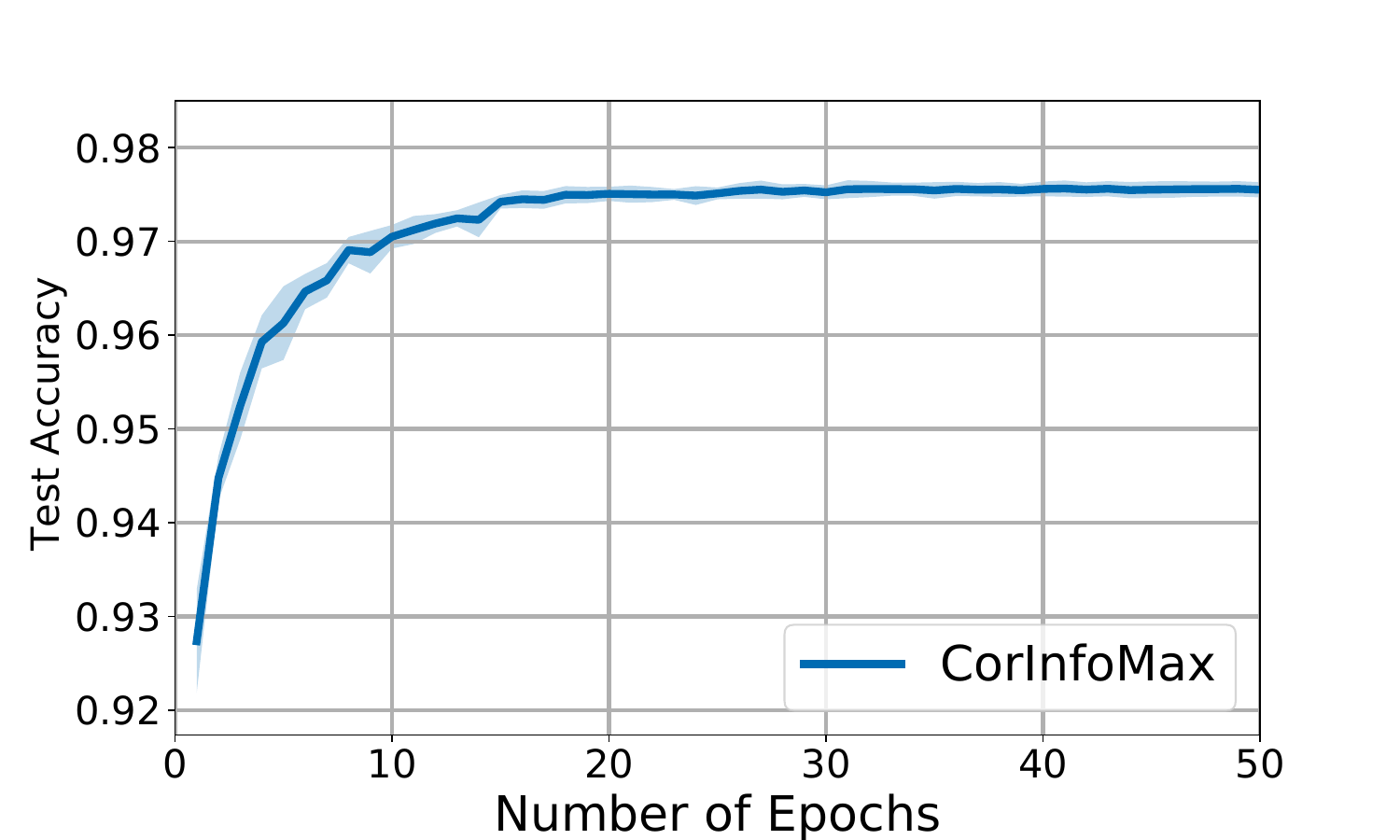}}
\hspace{0.01in}
\subfloat[]
{\includegraphics[trim = {0cm 0cm 0cm 1.0cm},clip,width=0.325\textwidth]{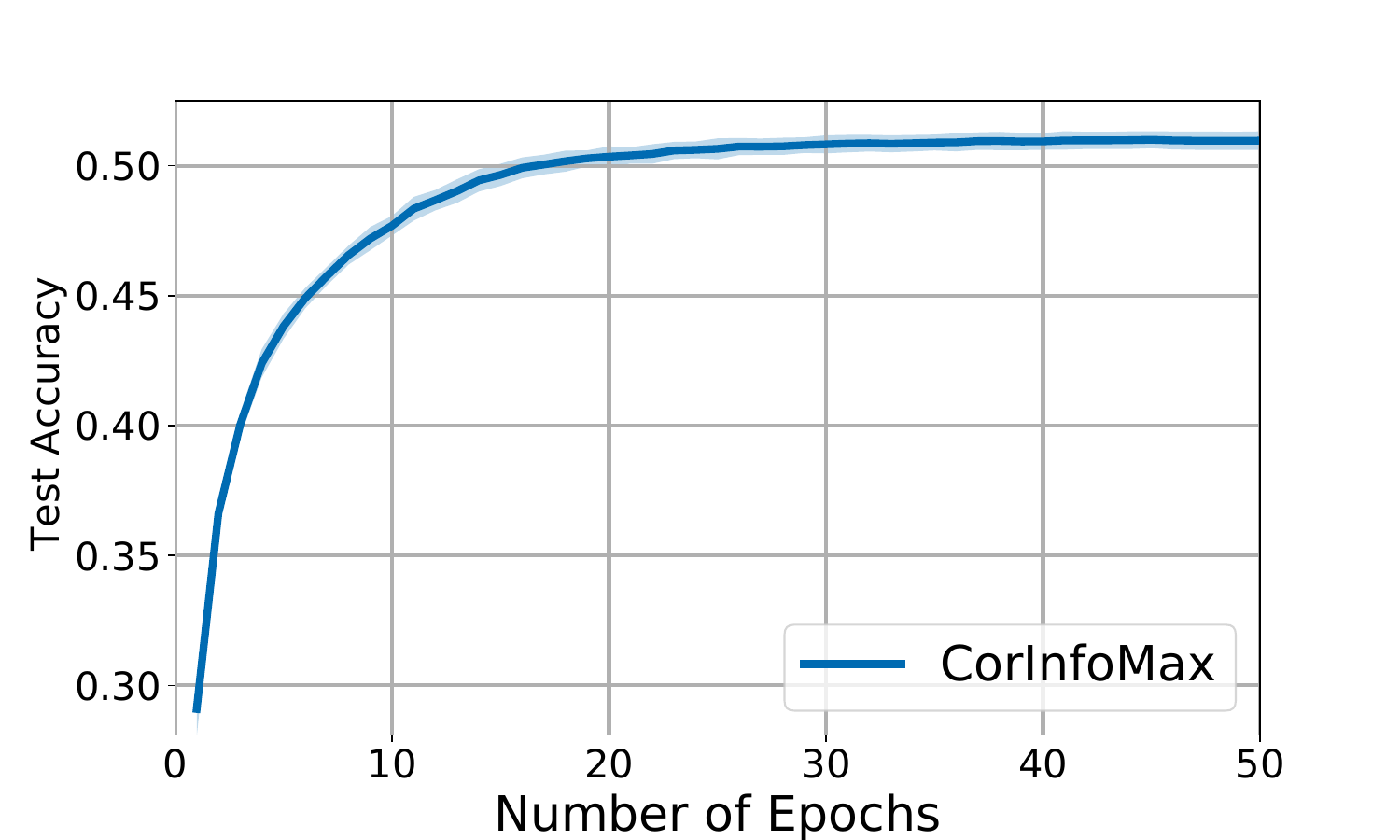}}
\hspace{0.01in}
\subfloat[]
{\includegraphics[trim = {0cm 0cm 0cm 1.0cm},clip,width=0.325\textwidth]{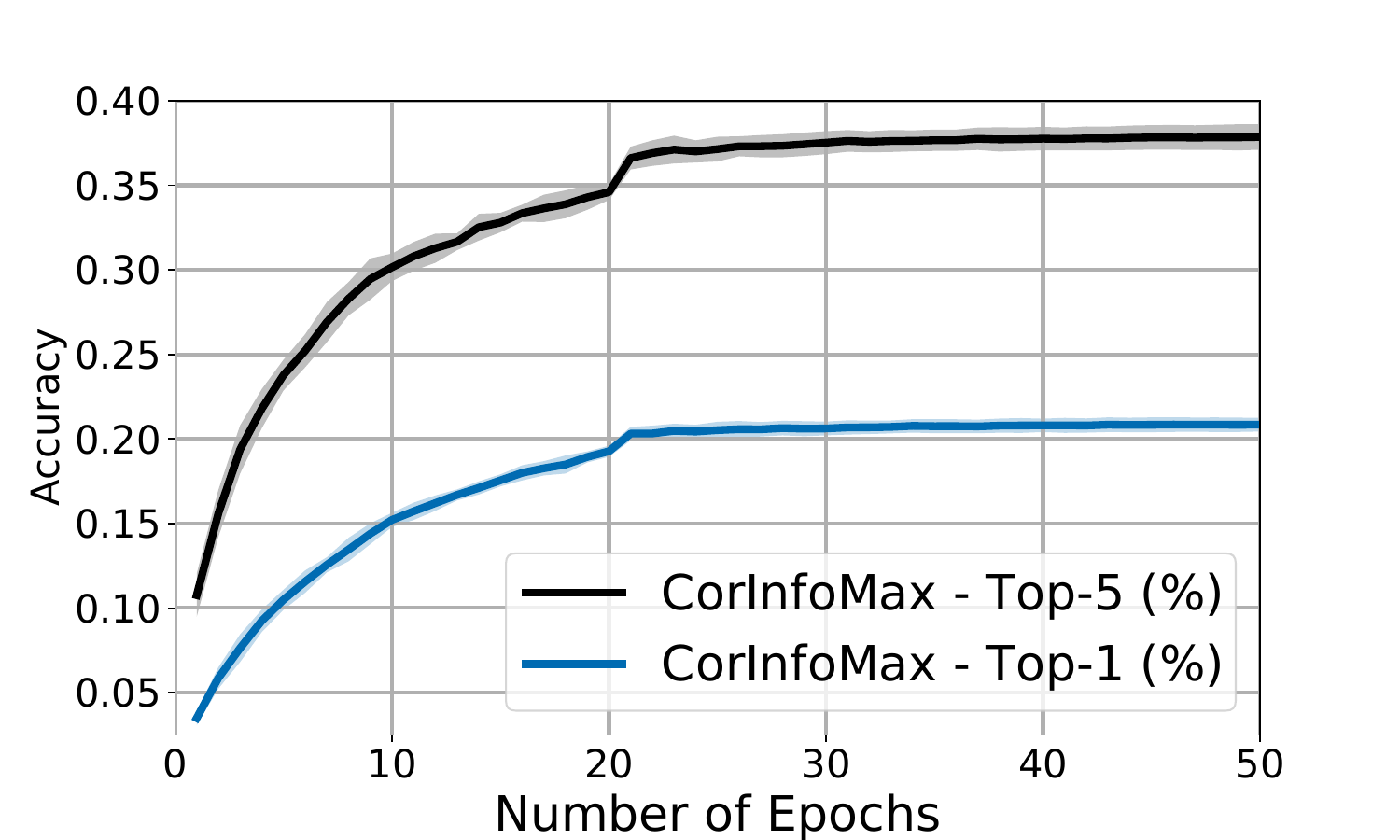}}
\newline\caption{Test accuracy convergence of CorInfoMax-$\mathcal{B}_{\infty,+}$ network with two hidden layers as a function of epochs (averaged over $n=10$ runs associated with the corresponding $\pm$ std envelopes) for the (a) MNIST dataset, (b) CIFAR10 dataset, and (c) CIFAR100 dataset (Top-1(\%) and Top-5(\%) test accuracy convergences)}
\label{fig:MNIST_CIFAR10_CIFAR100_3LayersConvergence}
\end{figure}

Figure \ref{fig:MNIST_CIFAR10_CIFAR100_3LayersConvergence} reports the test accuracy performance of  CorInfoMax-$\mathcal{B}_{\infty,+}$ network with two hidden layers as a function of training epochs for the MNIST, CIFAR10 and CIFAR100 image classification tasks.

\subsubsection{Angle measurement results}
\begin{figure}[h!]
\centering

\subfloat[b][]{
\includegraphics[trim = {0cm 0cm 0cm 0.0cm},clip,width=0.47\textwidth]{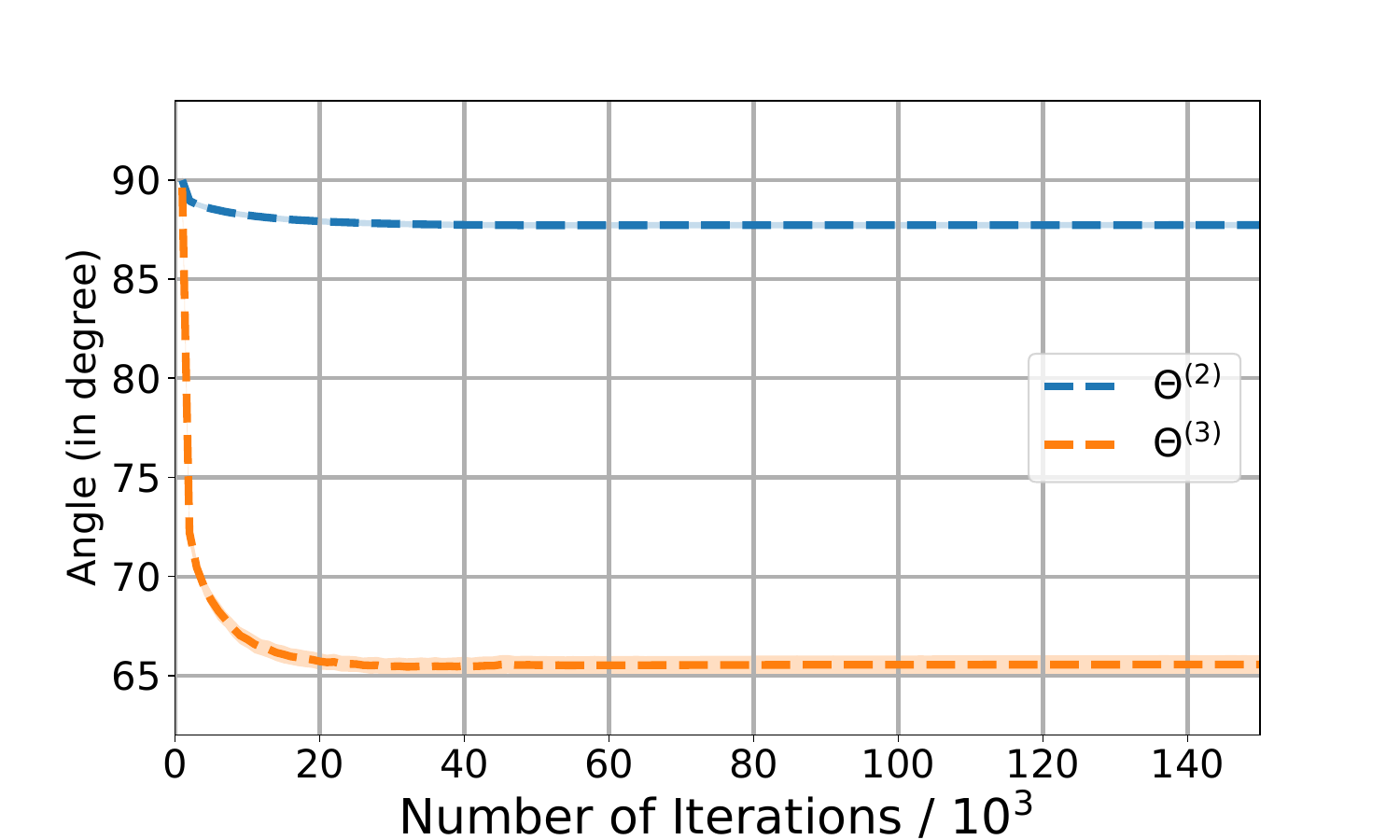}
\label{fig:3LayerMNISTAngle}
}

\subfloat[c][]{
\includegraphics[trim = {0cm 0cm 0cm 0.0cm},clip,width=0.47\textwidth]{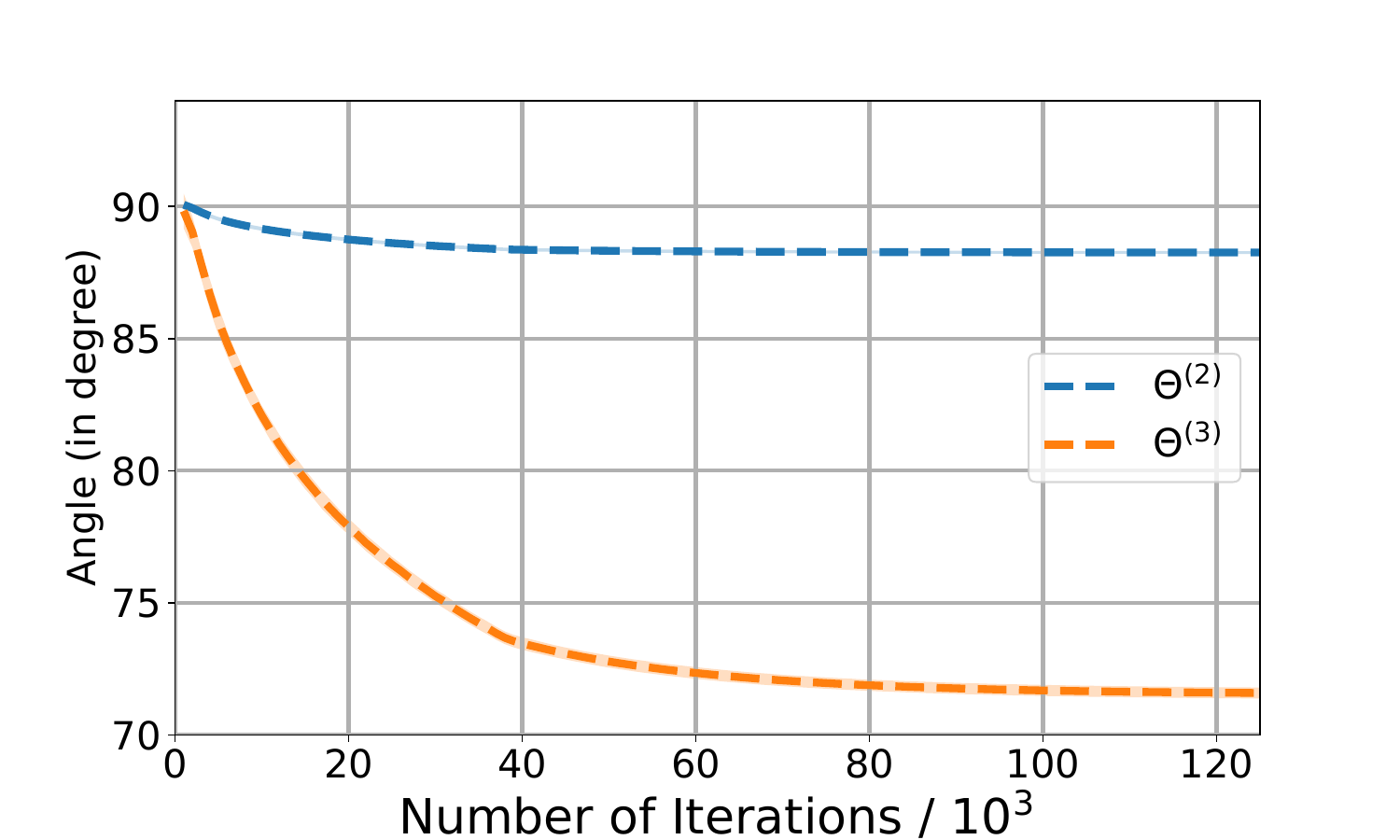}
\label{fig:3LayersCIFAR10Angle}
}
\subfloat[d][]{
\includegraphics[trim = {0cm 0cm 0cm 0.0cm},clip,width=0.47\textwidth]{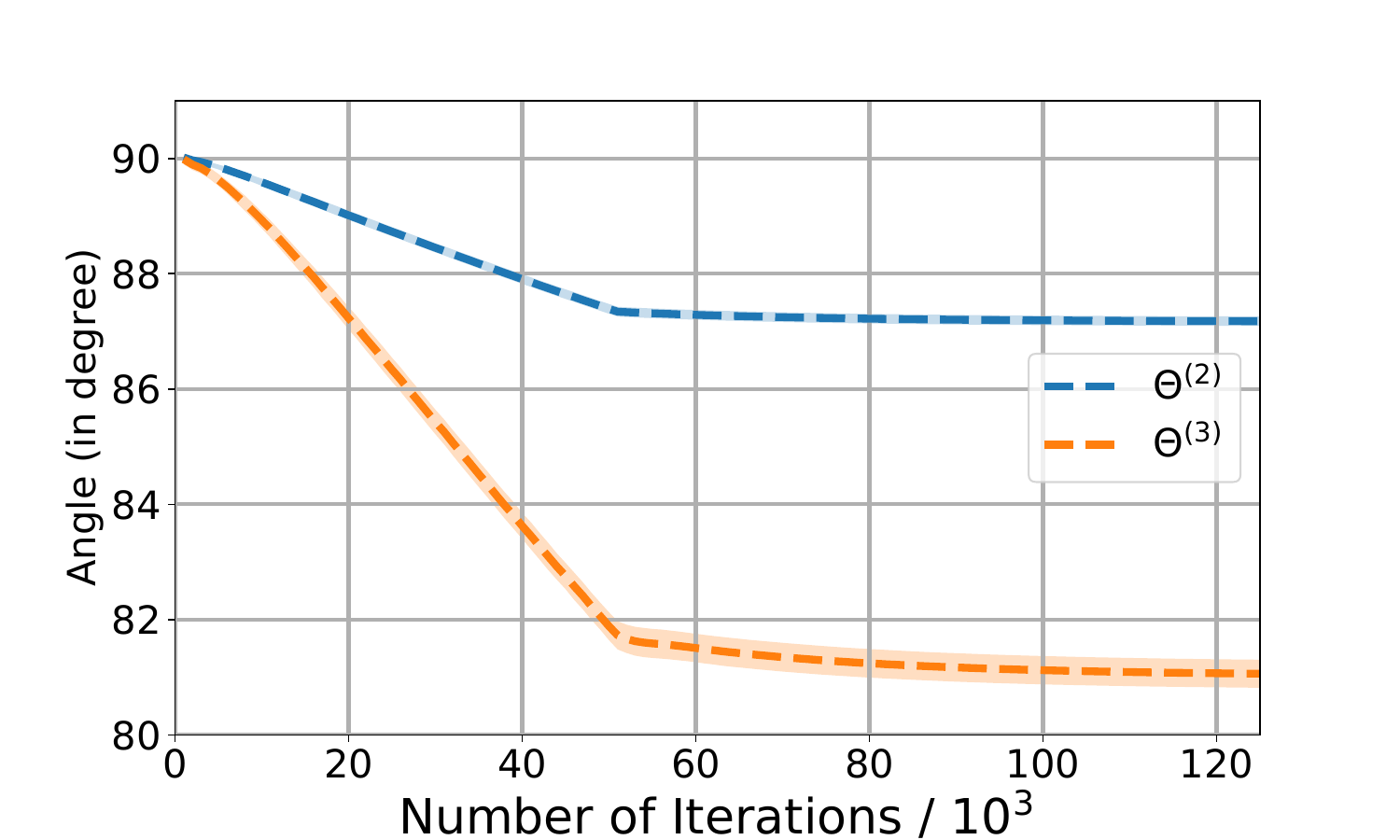}
\label{fig:CIFAR100Angle}
}
\newline\caption{The angle between the feedforward and the transpose of the feedback weights (based on (\ref{eq:angle}) for the $3$-layer CorInfoMax-$\mathcal{B}_{\infty,+}$ network (averaged over $n=10$ runs associated with the corresponding $\pm$ std envelopes) for (a) MNIST, (b) CIFAR10, and (c) CIFAR100 datasets.}
\hfill
\label{fig:ForwardBackwardAngle2}
\end{figure}

Figure \ref{fig:ForwardBackwardAngle2} provides the angle between feedforward and the transpose of the feedback weights (as defined in (\ref{eq:angle}))  results for the $3$-layer CorInfoMax-$\mathcal{B}_{\infty,+}$ network for  the MNIST, CIFAR10 and CIFAR100 datasets. These results also confirm the asymmetry between the feedforward and feedback weights  corresponding to the same segment.

\newpage

\subsection{Two layer CorInfoMax-\texorpdfstring{$\mathcal{B}_{1,+}$}{Lg} network}
\label{sec:sparse}

\subsubsection{Network architecture}

\begin{figure}[h]
\includegraphics[width=0.98\textwidth, trim={5.5cm, 12.0cm, 1.8cm, 7.0cm},clip]{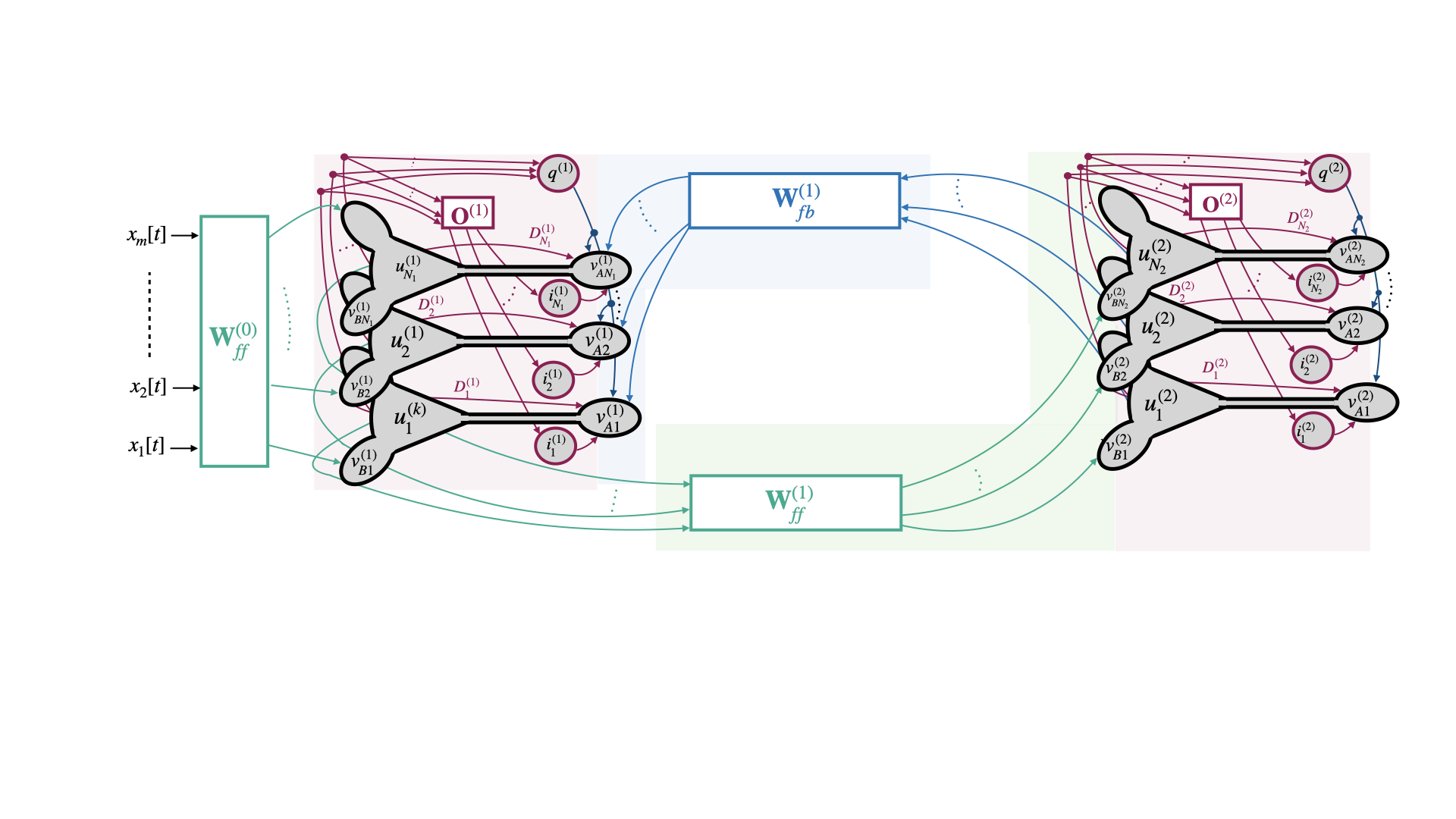}
\newline\caption{Two layer correlative information maximization based neural network with non-negative sparsity constraint: $\|\rvr^{(l)}\|_1\le 1, \text{ and } \rvr^{(l)}\succcurlyeq \mathbf{0},  l=1,2$. Interneurons with activations $q^{(1)}$ and $q^{(2)}$ are to impose $\ell_1-$norm constraints for both layers.} 
\label{fig:twolayercorinfomaxNNSparse}
\end{figure}

Figure \ref{fig:twolayercorinfomaxNNSparse} provides a depiction of a CorInfoMax network with a single hidden layer. In this instance, both the hidden and output layers have the same constraint set $\mathcal{P}=\mathcal{B}_{1,+}$. Relative to the CorInfoMax-$\mathcal{B}_{\infty,+}$ network structure in Figure \ref{fig:twolayercorinfomaxNNAntiSparse}, this network contains additional interneurons, namely $q^{(1)}$ and $q^{(2)}$ to impose sparsity constraint on hidden and output layer networks.

\subsubsection{Hyperparameters}

\begin{table}[h!]
  \caption{Hyperparameters used to train two layers CorInfoMax-$\mathcal{B}_{1,+}$ network. (In the row stating the lr decay, ep. and O/W means \textit{epoch} and \textit{otherwise}, respectively.)\newline}
  \label{CorInfoSparseOneHL_Hyperparams}
  \centering
  \begin{tabular}{llll}
    \toprule
    Hyperparameter     & MNIST   & FashionMNIST & CIFAR10 \\
    \midrule
    Batch size & $20$ & $20$ & $20$ \\
    Architecture & $[784, 500, 10]$ & $[784, 500, 10]$  & $[3072, 1000, 10]$\\
    $T_{\text{free}}$ & $20$  & $20$ & $30$ \\
    $T_{\text{nudged}}$ & $4$ & $10$ & $10$\\
    $\mu_{\text{ff}}$ & $[1.0, 0.7]$ & $[0.35, 0.23]$ & $[0.095, 0.075]$ \\
    $\mu_{\text{fb}}$ & $[-, 0.12]$ & $[-, 0.06]$ & $[ -, 0.05]$\\
    $\lambda_\rvr$ & $1 - 10^{-5}$ & $1 - 10^{-5}$ & $1 - 10^{-5}$\\
    $\epsilon_k$ & $0.15 \quad \forall k$ & $0.15 \quad \forall k$ & $0.15 \quad \forall k$\\
    $\beta'$ & $1.0$ & $1.0$ & $1.0$\\
    $g_{lk}$ & {$0.5$}  & 0.2& $0.1$\\
    $\mu_{\rvu}$ & {$0.05$} & $\text{max}\{\frac{0.045}{s \times 10^{-2} + 1}, 10^{-3} \}$ & $\text{max}\{\frac{0.025}{s \times 10^{-2} + 1}, 10^{-3} \}$ \\
    $\mu_{a}$ & $[1e-6, 0.01]$ & $[1e-6, 0.01]$ & $[1e-5, 0.01]$\\
    lr decay & $\left\{\begin{array}{cc}
                        0.95 & \text{ep.}< 15 \\
                        0.9 & \mbox{O/W}. \end{array}\right.$ & $\left\{\begin{array}{cc}
                        0.95 & \text{ep.}< 11 \\
                        0.9 & \mbox{O/W}. \end{array}\right.$ & $\left\{\begin{array}{cc}
                        0.95 & \text{ep.}< 15 \\
                        0.9 & \mbox{O/W}. \end{array}\right.$\\
    \bottomrule
  \end{tabular}
  
\end{table}

\newpage

\subsubsection{Test accuracy results}
\begin{figure}[h!]

\subfloat[]{
\includegraphics[trim = {0cm 0cm 0cm 1.0cm},clip,width=0.325\textwidth]{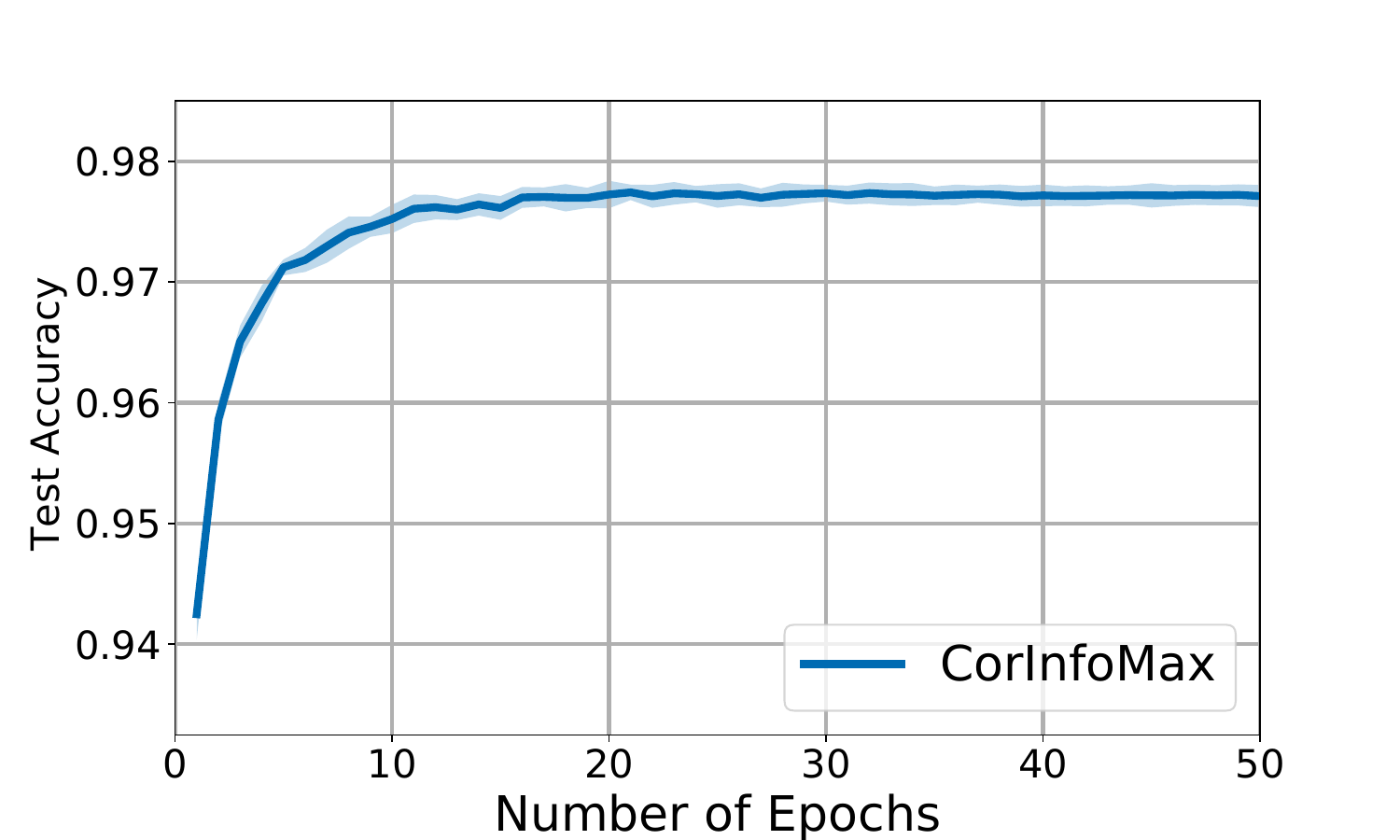}}
\subfloat[]{
\includegraphics[trim = {0cm 0cm 0cm 1.0cm},clip,width=0.325\textwidth]{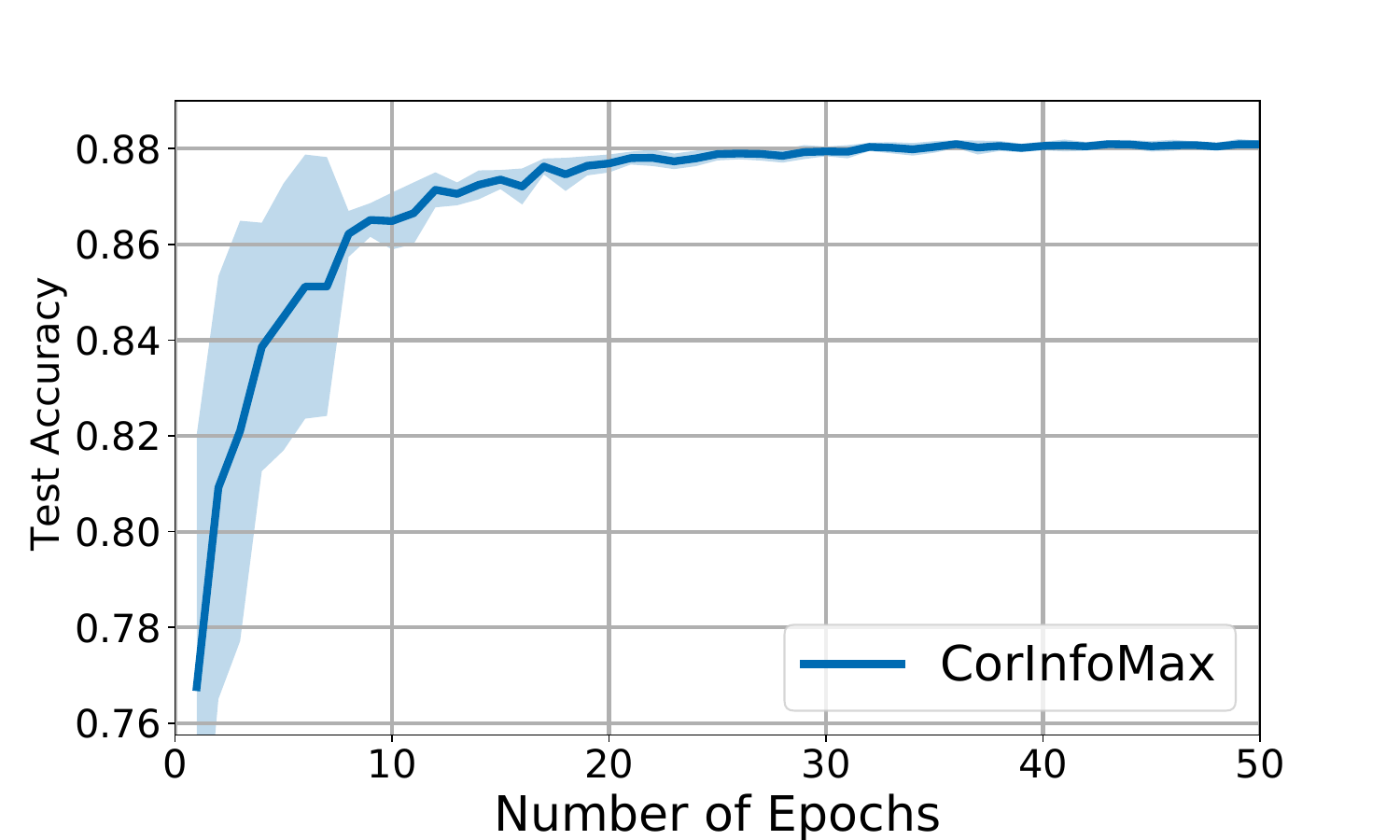}}
\subfloat[]{
\includegraphics[trim = {0cm 0cm 0cm 1.0cm},clip,width=0.325\textwidth]{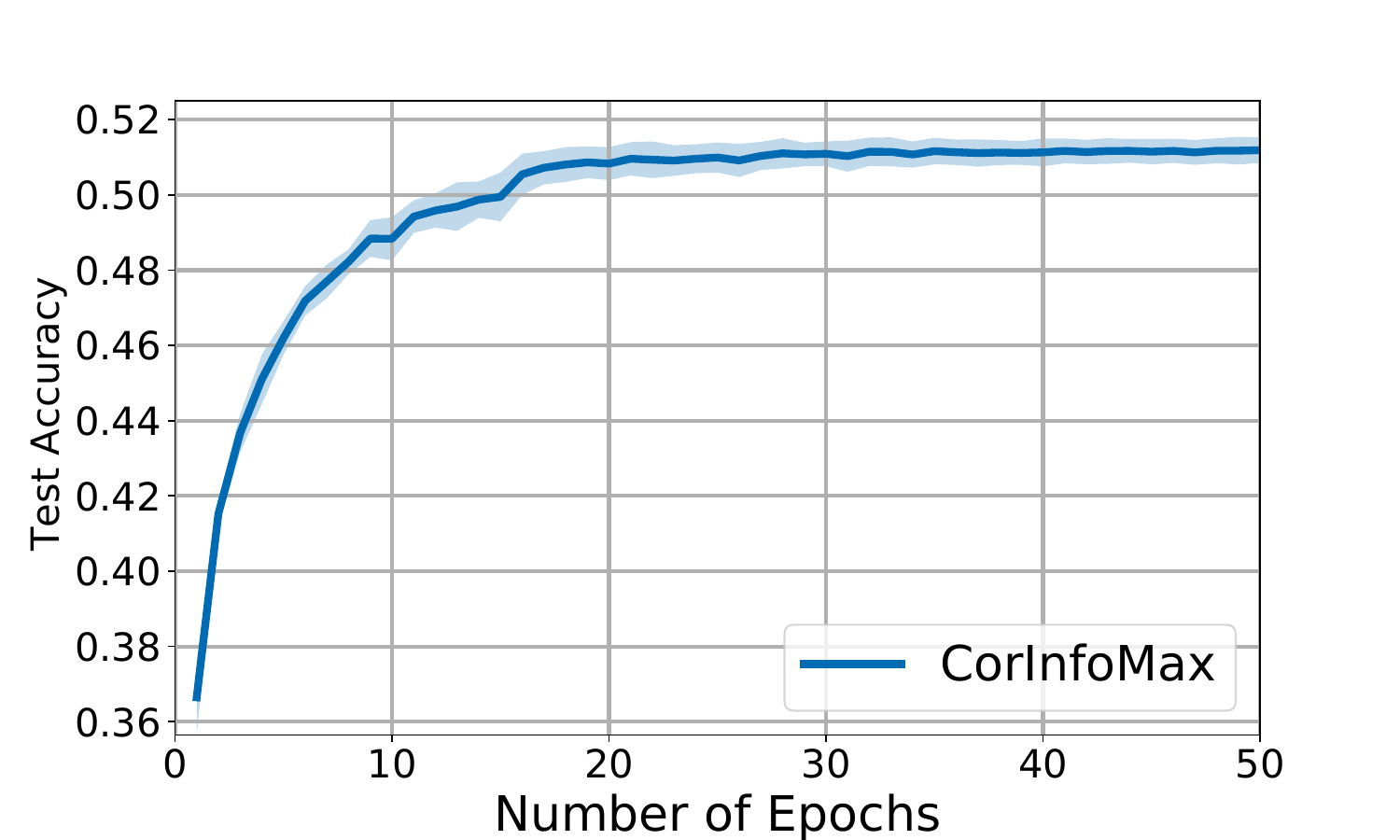}}
\newline\caption{Test accuracy convergence of CorInfoMax-$\mathcal{B}_{1,+}$ network with single hidden layer as a function of epochs (averaged over $n=10$ runs associated with the corresponding $\pm$ std envelopes) for the (a) MNIST dataset, (b) FashionMNIST dataset, and (c) CIFAR10 dataset.}
\label{fig:MNIST_Fashion_CIFAR10SparseConvergence}
\end{figure}

Figure \ref{fig:MNIST_Fashion_CIFAR10SparseConvergence} reports the test accuracy performance of  CorInfoMax-$\mathcal{B}_{1,+}$ network with single hidden layer as a function of training epochs for the MNIST, FashionMNIST, and CIFAR10 image classification tasks.

\subsubsection{Angle measurement results}
\begin{figure}[h!]
\hspace*{1.45cm}
\includegraphics[trim = {0cm 0cm 0cm 1.0cm},clip,width=0.6\textwidth]{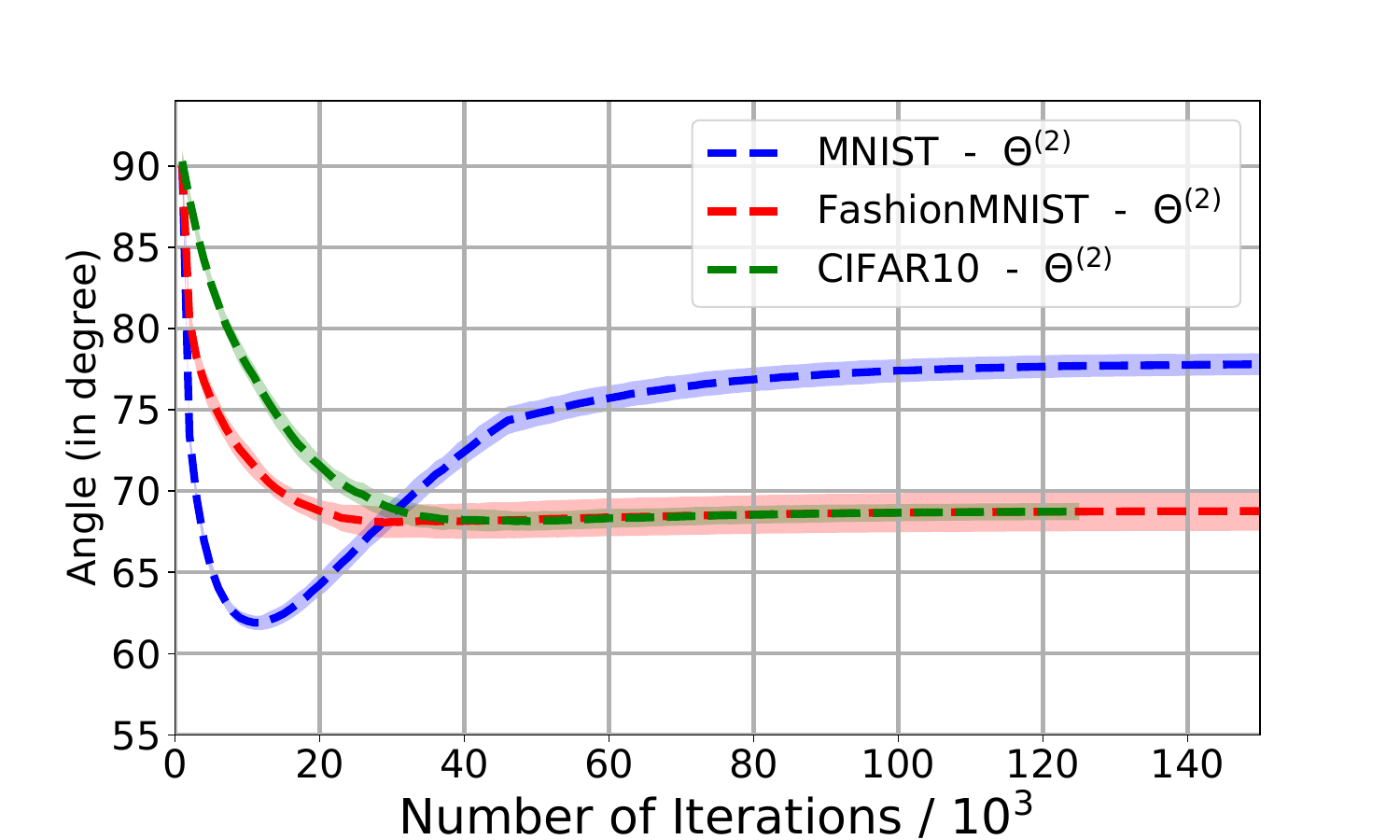}
	\caption{The angle between the feedforward and the transpose of the feedback weights between hidden and output layers (averaged over $n=10$ runs associated with the corresponding $\pm$ std envelopes) as a function of weight update iterations for CorInfoMax-$\mathcal{B}_{1,+}$. }
\label{fig:AngleSparseNetwork}
\end{figure}

The angle measurements between the feedforward and feedback weights demonstrated in Figure \ref{fig:AngleSparseNetwork} confirm the typical asymmetry between these weights.

\subsection{Compute sources}

All the simulations are carried out in an HPC cluster using either a single Tesla T4 or a single Tesla V100 GPU. Although the simulation times are dependent on the memory utilization and number of CPUs, the following list declares the approximate training periods for our experiments:
\begin{itemize}
    \item One epoch training of CorInfoMax-$\mathcal{B}_{\infty, +}$ networks with single hidden layer on MNIST, Fashion MNIST, and CIFAR10 datasets using batch size 20 takes approximately 1-2 minutes,
    \item One epoch training of CorInfoMax-$\mathcal{B}_{\infty, +}$ network with two hidden layers on CIFAR10 dataset using batch size 20 takes approximately 2-3 minutes,
    \item One epoch training of CorInfoMax-$\mathcal{B}_{\infty, +}$ network with two hidden layers on CIFAR100 dataset using batch size 20 takes approximately 3-4 minutes,
    \item One epoch training of CorInfoMax-$\mathcal{B}_{1, +}$ network with single hidden layer on MNIST, Fashion MNIST, and CIFAR10 datasets using batch size 20 takes approximately 1-2 minutes.
\end{itemize}

\section{Ablation studies: the effect of hyperparameters}
\label{sec:ablation}
In this section, we examine the influence of various hyperparameters on the performance of our proposed framework. To achieve this, we conducted a series of simulations over a selected grid of parameters. The outcomes, in terms of training and test accuracy, from these grid-based experiments are detailed here.

\subsection{Neural dynamic's learning rate and leakage conductance \texorpdfstring{$g_{lk}$}{Lg}}
Table \ref{tab:HyperparamGridMNIST_V1} presents the effect of the neural dynamic's learning rate and the value of $g_{lk}$ on the training and test accuracies for CorInfoMax-$\mathcal{B}_{\infty, +}$ network with single hidden layer on the MNIST image classification task. We observe fairly stable performance over the selected range of values.

\begin{table}[h!]
  \caption{Train and test accuracies (mean $\pm$ standard deviation of $n=10$ runs) with respect to the combination of neural dynamic's learning rate and $g_{lk}$ for the MNIST simulations with $2$-layer CorInfoMax-$\mathcal{B}_{\infty, +}$ network. The other hyperparameters are as specified in Table \ref{CorInfoOneHL_Hyperparams}.\newline}
  \centering
  \begin{tabular}{lllll}
    \toprule
        \hspace{1.5cm}$\mu_{\rvu}$ & $g_{lk}$     & Train accuracy & Test accuracy\\
    \midrule
     $\text{max}\{\frac{0.05}{s \times 10^{-2} + 1}, 10^{-3} \}$ & $0.5$ &$98.915\pm0.06$ & $97.613\pm0.11$ \\
     $\text{max}\{\frac{0.05}{s \times 10^{-2} + 1}, 10^{-3} \}$ & $0.2$ &$98.919\pm0.04$ & $97.610\pm0.10$  \\
     $0.05$ & $0.5$ & $98.930\pm0.05$& $97.622\pm0.12$\\
     $0.05$ & $0.2$ & $98.929\pm0.04$& $97.622\pm0.10$ \\

    \bottomrule
    \label{tab:HyperparamGridMNIST_V1}
  \end{tabular}
\end{table}

Similarly, Table \ref{tab:HyperparamGridMNISTSparse_V1} demonstrates the impression of the learning rate for neural dynamics and the leakage conductance $g_{lk}$ for CorInfoMax-$\mathcal{B}_{1, +}$ with single hidden layer. We note that the training and test accuracy results in the first row of Table \ref{tab:HyperparamGridMNISTSparse_V1} is low with high variance. This is due to the fact that one out of ten runs has diverged after epoch $35$ for this hyperparameter combination.

\begin{table}[h!]
  \caption{Train and test accuracies (mean $\pm$ standard deviation of $n=10$ runs) with respect to the combination of neural dynamic's learning rate and $g_{lk}$ for the MNIST simulations with $2$-layer CorInfoMax-$\mathcal{B}_{1, +}$ network. The other hyperparameters are as specified in Table \ref{CorInfoSparseOneHL_Hyperparams}.\newline}
  \centering
  \begin{tabular}{lllll}
    \toprule
        \hspace{1cm}$\mu_{\rvu}$ & $g_{lk}$     & Train accuracy & Test accuracy\\
    \midrule
     $\text{max}\{\frac{0.05}{s \times 10^{-2} + 1}, 10^{-3} \}$ & $0.5$ &$89.793\pm28.1$ & $88.891\pm27.79$ \\
     $\text{max}\{\frac{0.05}{s \times 10^{-2} + 1}, 10^{-3} \}$ & $0.2$ &$98.634\pm0.13$ & $97.634\pm0.12$  \\
     $0.05$ & $0.5$ & $98.698\pm0.06$& $97.711\pm0.10$\\
     $0.05$ & $0.2$ & $98.695\pm0.05$& $97.724\pm0.08$ \\

    \bottomrule
    \label{tab:HyperparamGridMNISTSparse_V1}
  \end{tabular}
\end{table}

\newpage
\subsection{Learning rates for synapses and network dynamics}
Table \ref{tab:HyperparamGridFashionMNIST_V1} reports performance variation for different choices of neural dynamic and synaptic learning rates for CorInfoMax-$\mathcal{B}_{\infty, +}$ with single hidden layer on the Fashion MNIST classification task. Better performance is achieved for relatively higher feedforward synapse learning rates and relatively smaller feedback synapse learning rates.

\begin{table}[h!]
  \caption{Train and test accuracies (mean $\pm$ standard deviation of $n=10$ runs) with respect to the combination of feedforward and feedback learning rates, and neural dynamic's learning rate for the Fashion MNIST simulations with $2$-layer CorInfoMax-$\mathcal{B}_{\infty, +}$ network. The other hyperparameters are as specified in Table \ref{CorInfoOneHL_Hyperparams}.\newline}
  \centering
  \begin{tabular}{lllll}
    \toprule
        \hspace{0.5cm}$\mu_{\text{ff}}$ & \hspace{0.5cm}$\mu_{\text{fb}}$ & \hspace{1.5cm}$\mu_{\rvu}$ & Train accuracy & Test accuracy\\
    \midrule
     $[0.3 , 0.22]$ & $[ -, 0.07]$ &$\text{max}\{\frac{0.07}{s \times 10^{-2} + 1}, 10^{-3} \}$ & $91.468\pm0.22$ & $88.138\pm0.28$ \\
     $[0.3 , 0.22]$ & $[ -, 0.07]$ &$\text{max}\{\frac{0.05}{s \times 10^{-2} + 1}, 10^{-3} \}$ & $91.230\pm0.11$ & $88.140\pm0.16$ \\
     $[0.25, 0.15]$ & $[ -, 0.09]$ &$\text{max}\{\frac{0.07}{s \times 10^{-2} + 1}, 10^{-3} \}$ & $89.961\pm2.79$ & $87.215\pm2.37$ \\
     $[0.25, 0.15]$ & $[ -, 0.09]$ &$\text{max}\{\frac{0.05}{s \times 10^{-2} + 1}, 10^{-3} \}$ & $90.530\pm0.12$ & $87.770\pm0.23$ \\

    \bottomrule
    \label{tab:HyperparamGridFashionMNIST_V1}
  \end{tabular}
\end{table}
\subsection{Forgetting factor and learning rates}
Table \ref{tab:HyperparamGridCIFAR10_V1} shows  the impact of forgetting factor $\lambda_{\rvr}$ together with synaptic learning rates for CorInfoMax-$\mathcal{B}_{\infty, +}$ with single hidden layer on the CIFAR10 classification accuracy. Overall, a slightly improved performance is observed when using relatively higher values of $\mu_{\text{ff}}$ and relatively smaller values of $\mu_{\text{fb}}$.

\begin{table}[h!]
  \caption{Train and test accuracies (mean $\pm$ standard deviation of $n=10$ runs) with respect to the combination of feedforward and feedback learning rates, and $\lambda_{\rvr}$ for the CIFAR10 simulations with $2$-layer CorInfoMax-$\mathcal{B}_{\infty, +}$ network. The other hyperparameters are as specified in Table \ref{CorInfoOneHL_Hyperparams}.\newline}
  \centering
  \begin{tabular}{lllll}
    \toprule
        \hspace{0.5cm}$\mu_{\text{ff}}$ & \hspace{0.5cm}$\mu_{\text{fb}}$  & \hspace{0.4cm}$\lambda_{\rvr}$ & Train accuracy & Test accuracy\\
    \midrule
     $[0.08, 0.04]$ & $[ -, 0.04]$ &$1 - 10^{-5}$ & $64.841\pm0.17$ & $51.732\pm0.34$ \\
     $[0.07, 0.03]$ & $[ -, 0.05]$ &$1 - 10^{-5}$ & $62.427\pm0.11$ & $51.065\pm0.37$ \\
     $[0.08, 0.04]$ & $[ -, 0.04]$ &$1 - 5 \times 10^{-5}$ & $64.848\pm0.16$ & $51.856\pm0.33$ \\
     $[0.07, 0.03]$ & $[ -, 0.05]$ &$1 - 5 \times 10^{-5}$ & $62.456\pm0.18$ & $51.066\pm0.32$ \\
    \bottomrule
    \label{tab:HyperparamGridCIFAR10_V1}
  \end{tabular}
\end{table}

Likewise, Table \ref{tab:HyperparamGridSparseCIFAR10_V1} illustrates the effect of the variations on the same set of hyperparameters for CorInfoMax-$\mathcal{B}_{1, +}$ with a single hidden layer for CIFAR10 classification task. Comparatively, the resulting performances exhibit a moderate level of stability when compared to the CorInfoMax-$\mathcal{B}_{\infty, +}$ results presented in Table \ref{tab:HyperparamGridCIFAR10_V1}.

\begin{table}[h!]
  \caption{Train and test accuracies (mean $\pm$ standard deviation of $n=10$ runs) with respect to the combination of feedforward and feedback learning rates, and $\lambda_{\rvr}$ for the CIFAR10 simulations with $2$-layer CorInfoMax-$\mathcal{B}_{1, +}$ network. The other hyperparameters are as specified in Table \ref{CorInfoSparseOneHL_Hyperparams}.\newline}
  \centering
  \begin{tabular}{lllll}
    \toprule
        \hspace{0.5cm}$\mu_{\text{ff}}$ & \hspace{0.5cm}$\mu_{\text{fb}}$  & $\lambda_{\rvr}$ & Train accuracy & Test accuracy\\
    \midrule
     $[0.09, 0.07]$ & $[ -, 0.045]$ &$1 - 10^{-5}$ & $63.005\pm0.48$ & $51.047\pm0.40$ \\
     $[0.095, 0.075]$ & $[ -, 0.05]$ &$1 - 10^{-5}$ & $63.719\pm0.56$ & $51.188\pm0.36$ \\
     $[0.09, 0.07]$ & $[ -, 0.045]$ &$1 - 5 \times 10^{-5}$ & $63.003\pm0.54$ & $51.062\pm0.32$ \\
     $[0.095, 0.075]$ & $[ -, 0.05]$ &$1 - 5 \times 10^{-5}$ & $63.716\pm0.53$ & $51.106\pm0.38$\\
    \bottomrule
    \label{tab:HyperparamGridSparseCIFAR10_V1}
  \end{tabular}
\end{table}

\vspace{0.5in}
\newpage
\subsection{Learning rate decay, free phase iterations, synaptic learning rates}

Table \ref{tab:HyperparamGridCIFAR100_V1} investigates the influence of learning rate decay, the number of free phase iterations, and synaptic learning rates on classification accuracy for CorInfoMax-$\mathcal{B}_{\infty, +}$ with two hidden layers on the CIFAR100 classification task. The best train and test results are acquired by utilizing higher values of $\mu_{\text{ff}}$ and $\mu_{\text{fb}}$, along with implementing a learning rate decay during the initial epochs.

\begin{table}[h!]
  \caption{Train and test accuracies (mean $\pm$ standard deviation of $n=10$ runs) with respect to the combination of feedforward and feedback learning rates, learning rate decay, and number of iterations for the free phase ($T_{\text{free}}$) for the CIFAR100 simulations with $3$-layer CorInfoMax-$\mathcal{B}_{\infty, +}$ network. The other hyperparameters are as specified in Table \ref{CorInfoTwoHL_Hyperparams}. (In the row stating the lr decay, ep. and O/W means \textit{epoch} and \textit{otherwise}, respectively.)\newline}
\begin{tabular}{llllll}
    \toprule
        \hspace{1cm}$\mu_{\text{ff}}$ & \hspace{1cm}$\mu_{\text{fb}}$  & \hspace{1cm}lr decay & $T_{\text{free}}$ & Train acc. & Test acc.\\
        \cmidrule(r){5-6}
        &&&& Top-1/5  & Top-1/5\\
    \midrule
     $[0.16, 0.13, 0.08]$ & $[ -, 0.06, 0.04]$ &$\left\{\begin{array}{cc}
                        1.0 & \text{ep.}< 15 \\
                        0.9 & \mbox{O/W}. \end{array}\right.$ & $50$ & $23.9 / 40.9$ & $19.1 / 36.4$\\
     $[0.16, 0.13, 0.08]$ & $[ -, 0.06, 0.04]$ &$\left\{\begin{array}{cc}
                        0.99 & \text{ep.}< 20 \\
                        0.9 & \mbox{O/W}. \end{array}\right.$ & $50$ & $26.0 / 42.8$ & $20.2 / 37.6$\\
     $[0.16, 0.13, 0.08]$ & $[ -, 0.06, 0.04]$ & $\left\{\begin{array}{cc}
                        1.0 & \text{ep.}< 15 \\
                        0.9 & \mbox{O/W}. \end{array}\right.$ & $60$ & $23.9 / 41.0$ & $19.1 / 36.5$\\
     $[0.18, 0.15, 0.09]$ & $[ -, 0.08, 0.06]$ &$\left\{\begin{array}{cc}
                        0.99 & \text{ep.}< 20 \\
                        0.9 & \mbox{O/W}. \end{array}\right.$ & $50$ & $27.7 / 43.5$ & $20.8 / 37.9$\\
    \bottomrule
    \label{tab:HyperparamGridCIFAR100_V1}
  \end{tabular}
\end{table}

\subsection{Larger variations on forward mapping learning rate and neural dynamic learning rate}




To assess the impact of larger variations in certain hyperparameters on performance, we conducted two additional ablation studies using both the MNIST and CIFAR10 datasets. These experiments were conducted using a two-layered CorInfoMax-$\mathcal{B}_{\infty, +}$ network. Specifically, we explored variations in the selection of the learning rate for the forward mapping ($\mu_{ff}$) and the initial learning rate for neural dynamics ($\mu_{\mathbf{u}}[1]$).

\begin{figure}[h!]
\subfloat[\label{fig:MNIST_CIFAR_muForward_Ablation_a}]{
\hspace{-0.35cm}\includegraphics[trim = {0cm 0cm 0cm 0cm},clip,width=0.495\textwidth]{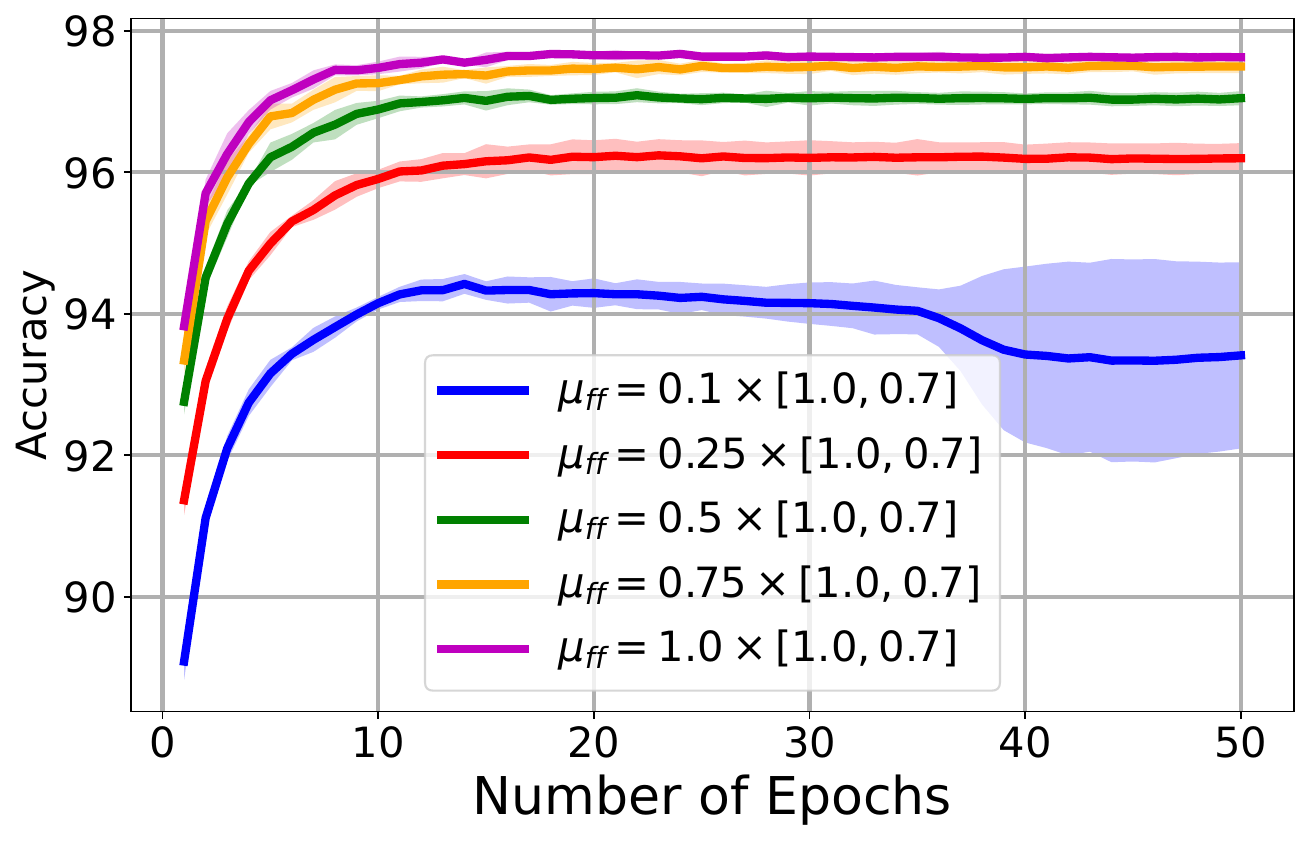}}
\subfloat[\label{fig:MNIST_CIFAR_muForward_Ablation_b}]{
\includegraphics[trim = {0cm 0cm 0cm 0cm},clip,width=0.495\textwidth]{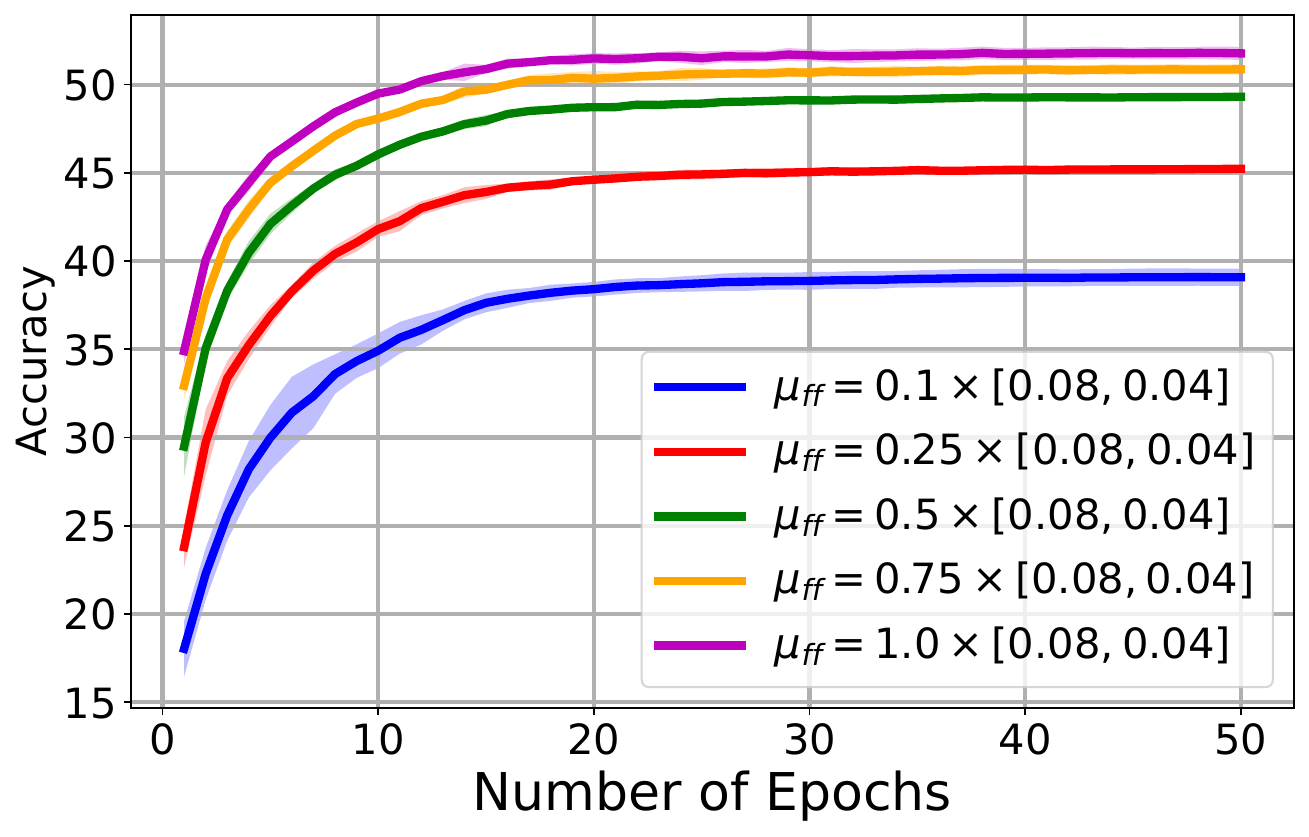}} 
\newline\caption{Ablation study on $\mu_{ff}$ variations for two-layered CorInfoMax-$\mathcal{B}_{\infty, +}$ networks. (a) Convergence of test accuracy across different selections of $\mu_{ff}$ for the MNIST classification task (averaged over $n = 5$ runs associated with the corresponding $\pm$ std envelopes). (b) Convergence of test accuracy across different selections of $\mu_{ff}$ for the CIFAR-10 classification task (averaged over $n = 5$ runs associated with the corresponding $\pm$ std envelopes). }
\label{fig:MNIST_CIFAR_muForward_Ablation}
\end{figure}

Figure \ref{fig:MNIST_CIFAR_muForward_Ablation_a} presents the test accuracy curves for the MNIST classification task, showcasing different selections of $\mu_{ff}$ averaged over five runs, with standard deviation envelopes ($\pm$std). The remaining hyperparameters of the CorInfoMax-$\mathcal{B}_{\infty, +}$ network were kept constant at the values reported in Table \ref{CorInfoOneHL_Hyperparams}, including the learning rate decay. It is important to note that the specific value of $\mu_{ff} = [1.0, 0.7]$ corresponds to the value reported in Table \ref{CorInfoOneHL_Hyperparams} for our experiments. Our observations indicate that as the forward mapping learning rate approaches this reported value, the accuracy curve exhibits gradual improvement. Conversely, for relatively small values of $\mu_{ff}$, the resulting accuracy is lower and exhibits higher variance.

Figure \ref{fig:MNIST_CIFAR_muForward_Ablation_b} presents the same experiment conducted for the CIFAR10 classification task. Once again, we observe that as $\mu_{ff}$ approaches the value reported in Table \ref{CorInfoOneHL_Hyperparams}, the accuracy improves.

\begin{figure}[h!]
\centering
\includegraphics[trim = {0cm 0cm 0cm 0cm},clip,width=0.9\textwidth]{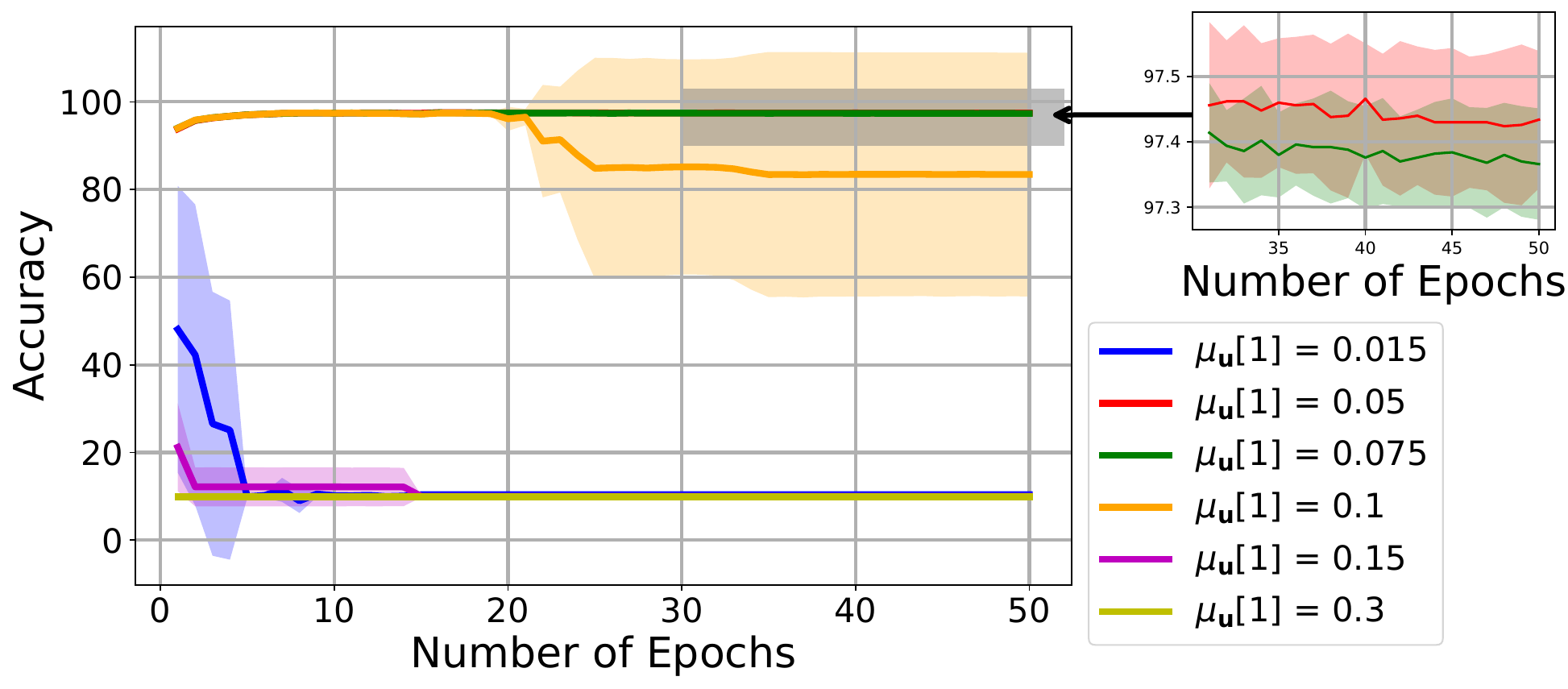}
\newline\caption{
Ablation study on $\mu_{\mathbf{u}}[1]$ variations: Test accuracy curves in a two-layered CorInfoMax-$\mathcal{B}_{\infty, +}$ network for MNIST classification (averaged over $n = 5$ runs associated with the corresponding $\pm$ std envelopes). Convergence to approximately $97.5\%$ accuracy is observed for $\mu_{\mathbf{u}}[1]$ values of $0.05$ and $0.075$.
}
\label{fig:MNIST_neuralLR_Ablation}
\end{figure}

In addition, we conducted experiments to investigate the impact of varying the initial learning rate for neural dynamics, denoted as $\mu_{\mathbf{u}}[1]$. Figure \ref{fig:MNIST_neuralLR_Ablation} presents test accuracy curves for the MNIST classification task with two-layered CorInfoMax-$\mathcal{B}_{\infty, +}$, which were obtained by averaging results from five runs and displaying standard deviation envelopes ($\pm $std). The remaining hyperparameters were held constant at the values specified in Table \ref{CorInfoOneHL_Hyperparams}. In Table \ref{CorInfoOneHL_Hyperparams}, we define $\mu_{\mathbf{u}}[s]$ as $\text{max}\{\frac{\mu_{\mathbf{u}}[1]}{s \times 10^{-2} + 1}, 10^{-3} \}$. Consequently, for this experiment, we applied the same decay rule to all selected values of $\mu_{\mathbf{u}}[1]$. Notably, our results indicate that $\mu_{\mathbf{u}}[1]$ values of $0.05$ or $0.075$ consistently yield reliable accuracy outcomes, while other values fail to provide dependable accuracy estimates.

\begin{figure}[h!]
\centering
\includegraphics[trim = {0cm 0cm 0cm 0cm},clip,width=0.9\textwidth]{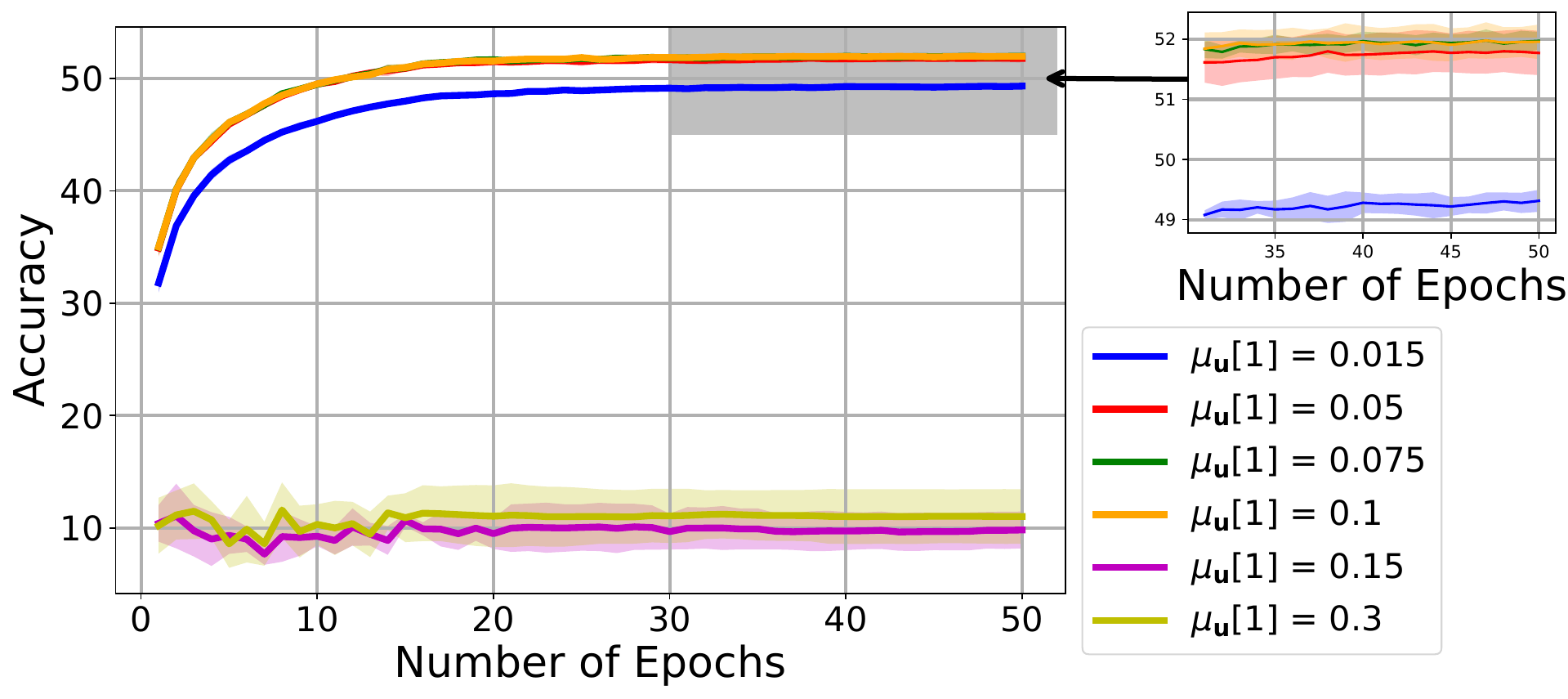}
\newline\caption{Ablation study on $\mu_{\mathbf{u}}[1]$ variations: Test accuracy curves in a two-layered CorInfoMax-$\mathcal{B}_{\infty, +}$ network for CIFAR10 classification (averaged over $n = 5$ runs associated with the corresponding $\pm$ std envelopes). Convergence to approximately $52\%$ accuracy is observed for $\mu_{\mathbf{u}}[1]$ values between $0.05$ and $0.1$.}
\label{fig:CIFAR_neuralLR_Ablation}
\end{figure}

\newpage
Similarly, we conducted the same experiment using a two-layered CorInfoMax-$\mathcal{B}_{\infty, +}$ network for the CIFAR10 classification task. Figure \ref{fig:CIFAR_neuralLR_Ablation} illustrates the test accuracy curves obtained for the CIFAR10 classification task, representing the average results of five runs with standard deviation envelopes ($\pm $std). The remaining hyperparameters were held constant at the values specified in Table \ref{CorInfoOneHL_Hyperparams}. It is worth noting that, for the CIFAR10 task, we did not implement decay rule for the neural dynamics learning rate, i.e., $\mu_{\mathbf{u}}[s] = \mu_{\mathbf{u}}[1] \quad \forall s$. Figure \ref{fig:CIFAR_neuralLR_Ablation} demonstrates that our method achieves convergence to approximately $52\%$ accuracy when $\mu_{\mathbf{u}}[1]$ falls within the range of $0.05$ to $0.1$. However, when $\mu_{\mathbf{u}}[1]$ is set to $0.015$, the accuracy slightly decreases, converging to around $49.5\%$. Notably, relatively higher values of $\mu_{\mathbf{u}}[1]$, such as $0.15$ and $0.3$, lead to divergence

\newpage

\end{document}

%% file: math_commands.tex
\usepackage{amsmath,amsfonts,bm}

\def\eqref#1{equation~\ref{#1}}

\def\1{\bm{1}}

\def\rve{{\mathbf{e}}}

\def\rvr{{\mathbf{r}}}

\def\rvu{{\mathbf{u}}}
\def\rvv{{\mathbf{v}}}

\def\rvx{{\mathbf{x}}}
\def\rvy{{\mathbf{y}}}
\def\rvz{{\mathbf{z}}}

\def\rmD{{\mathbf{D}}}

\def\rmM{{\mathbf{M}}}

\def\rmO{{\mathbf{O}}}

\def\rmR{{\mathbf{R}}}

\def\vzero{{\bm{0}}}
\def\vone{{\bm{1}}}

\def\vr{{\bm{r}}}
\def\vs{{\bm{s}}}

\def\vx{{\bm{x}}}
\def\vy{{\bm{y}}}

\def\evs{{s}}

\def\mB{{\bm{B}}}

\def\mD{{\bm{D}}}

\def\mI{{\bm{I}}}

\def\mM{{\bm{M}}}

\def\mO{{\bm{O}}}

\def\mR{{\bm{R}}}

\def\mW{{\bm{W}}}

\DeclareMathAlphabet{\mathsfit}{\encodingdefault}{\sfdefault}{m}{sl}
\SetMathAlphabet{\mathsfit}{bold}{\encodingdefault}{\sfdefault}{bx}{n}

\def\sZ{\mathbb{Z}^+}

\newcommand{\R}{\mathbb{R}}

\newcommand{\euler}{e}

\newcommand{\Pcal}{\mathcal{P}}

\DeclareMathOperator{\Tr}{Tr}

%% file: main_camera_ready.bbl
\begin{thebibliography}{10}

\bibitem{rumelhart1986learning}
David~E Rumelhart, Geoffrey~E Hinton, and Ronald~J Williams.
\newblock Learning representations by back-propagating errors.
\newblock {\em Nature}, 323(6088):533--536, 1986.

\bibitem{whittington2019theories}
James~CR Whittington and Rafal Bogacz.
\newblock Theories of error back-propagation in the brain.
\newblock {\em Trends in cognitive sciences}, 23(3):235--250, 2019.

\bibitem{crick1989recent}
Francis Crick.
\newblock The recent excitement about neural networks.
\newblock {\em Nature}, 337:129--132, 1989.

\bibitem{grossberg1987competitive}
Stephen Grossberg.
\newblock Competitive learning: From interactive activation to adaptive
  resonance.
\newblock {\em Cognitive science}, 11(1):23--63, 1987.

\bibitem{xie2003equivalence}
Xiaohui Xie and H~Sebastian Seung.
\newblock Equivalence of backpropagation and contrastive hebbian learning in a
  layered network.
\newblock {\em Neural computation}, 15(2):441--454, 2003.

\bibitem{scellier2017equilibrium}
Benjamin Scellier and Yoshua Bengio.
\newblock Equilibrium propagation: Bridging the gap between energy-based models
  and backpropagation.
\newblock {\em Frontiers in computational neuroscience}, 11:24, 2017.

\bibitem{larkum2013cellular}
Matthew Larkum.
\newblock A cellular mechanism for cortical associations: an organizing
  principle for the cerebral cortex.
\newblock {\em Trends in neurosciences}, 36(3):141--151, 2013.

\bibitem{urbanczik2014learning}
Robert Urbanczik and Walter Senn.
\newblock Learning by the dendritic prediction of somatic spiking.
\newblock {\em Neuron}, 81(3):521--528, 2014.

\bibitem{sacramento2018dendritic}
Jo{\~a}o Sacramento, Rui Ponte~Costa, Yoshua Bengio, and Walter Senn.
\newblock Dendritic cortical microcircuits approximate the backpropagation
  algorithm.
\newblock {\em Advances in neural information processing systems}, 31, 2018.

\bibitem{golkar2022constrained}
Siavash Golkar, Tiberiu Tesileanu, Yanis Bahroun, Anirvan Sengupta, and Dmitri
  Chklovskii.
\newblock Constrained predictive coding as a biologically plausible model of
  the cortical hierarchy.
\newblock {\em Advances in Neural Information Processing Systems},
  35:14155--14169, 2022.

\bibitem{bozkurt2023correlative}
Bariscan Bozkurt, Ate{\c{s}} {\.I}sfendiyaro{\u{g}}lu, Cengiz Pehlevan, and
  Alper~Tunga Erdogan.
\newblock Correlative information maximization based biologically plausible
  neural networks for correlated source separation.
\newblock In {\em The Eleventh International Conference on Learning
  Representations}, 2023.

\bibitem{petreanu2009subcellular}
Leopoldo Petreanu, Tianyi Mao, Scott~M Sternson, and Karel Svoboda.
\newblock The subcellular organization of neocortical excitatory connections.
\newblock {\em Nature}, 457(7233):1142--1145, 2009.

\bibitem{richards2019dendritic}
Blake~A Richards and Timothy~P Lillicrap.
\newblock Dendritic solutions to the credit assignment problem.
\newblock {\em Current opinion in neurobiology}, 54:28--36, 2019.

\bibitem{urban2016somatostatin}
Joanna Urban-Ciecko and Alison~L Barth.
\newblock Somatostatin-expressing neurons in cortical networks.
\newblock {\em Nature Reviews Neuroscience}, 17(7):401--409, 2016.

\bibitem{guerguiev2017towards}
Jordan Guerguiev, Timothy~P Lillicrap, and Blake~A Richards.
\newblock Towards deep learning with segregated dendrites.
\newblock {\em Elife}, 6:e22901, 2017.

\bibitem{Grossberg1987CompetitiveLF}
Stephen Grossberg.
\newblock Competitive learning: From interactive activation to adaptive
  resonance.
\newblock {\em Cogn. Sci.}, 11:23--63, 1987.

\bibitem{lillicrap2016random}
Timothy~P Lillicrap, Daniel Cownden, Douglas~B Tweed, and Colin~J Akerman.
\newblock Random synaptic feedback weights support error backpropagation for
  deep learning.
\newblock {\em Nature communications}, 7(1):13276, 2016.

\bibitem{Akrout2019Neurips}
Mohamed Akrout, Collin Wilson, Peter Humphreys, Timothy Lillicrap, and
  Douglas~B Tweed.
\newblock Deep learning without weight transport.
\newblock In H.~Wallach, H.~Larochelle, A.~Beygelzimer, F.~d\textquotesingle
  Alch\'{e}-Buc, E.~Fox, and R.~Garnett, editors, {\em Advances in Neural
  Information Processing Systems}, volume~32. Curran Associates, Inc., 2019.

\bibitem{amit2019deep}
Yali Amit.
\newblock Deep learning with asymmetric connections and hebbian updates.
\newblock {\em Frontiers in computational neuroscience}, 13:18, 2019.

\bibitem{HowImportantWeightSymmetry}
Qianli Liao, Joel Leibo, and Tomaso Poggio.
\newblock How important is weight symmetry in backpropagation?
\newblock {\em Proceedings of the AAAI Conference on Artificial Intelligence},
  30, 10 2015.

\bibitem{rao1999predictive}
Rajesh~PN Rao and Dana~H Ballard.
\newblock Predictive coding in the visual cortex: a functional interpretation
  of some extra-classical receptive-field effects.
\newblock {\em Nature neuroscience}, 2(1):79--87, 1999.

\bibitem{whittington2017approximation}
James~CR Whittington and Rafal Bogacz.
\newblock An approximation of the error backpropagation algorithm in a
  predictive coding network with local hebbian synaptic plasticity.
\newblock {\em Neural computation}, 29(5):1229--1262, 2017.

\bibitem{song2020can}
Yuhang Song, Thomas Lukasiewicz, Zhenghua Xu, and Rafal Bogacz.
\newblock Can the brain do backpropagation?---exact implementation of
  backpropagation in predictive coding networks.
\newblock {\em Advances in neural information processing systems},
  33:22566--22579, 2020.

\bibitem{EqProp}
Benjamin Scellier and Yoshua Bengio.
\newblock Equilibrium propagation: Bridging the gap between energy-based models
  and backpropagation.
\newblock {\em Frontiers in Computational Neuroscience}, 11, 2017.

\bibitem{ScalingEP}
Axel Laborieux, Maxence Ernoult, Benjamin Scellier, Yoshua Bengio, Julie
  Grollier, and Damien Querlioz.
\newblock Scaling equilibrium propagation to deep convnets by drastically
  reducing its gradient estimator bias.
\newblock {\em Frontiers in Neuroscience}, 15:633674, 02 2021.

\bibitem{laborieux2022EPholomorphic}
Axel Laborieux and Friedemann Zenke.
\newblock Holomorphic equilibrium propagation computes exact gradients through
  finite size oscillations.
\newblock In Alice~H. Oh, Alekh Agarwal, Danielle Belgrave, and Kyunghyun Cho,
  editors, {\em Advances in Neural Information Processing Systems}, 2022.

\bibitem{qin2021contrastive}
Shanshan Qin, Nayantara Mudur, and Cengiz Pehlevan.
\newblock Contrastive similarity matching for supervised learning.
\newblock {\em Neural computation}, 33(5):1300--1328, 2021.

\bibitem{scellier2018generalization}
Benjamin Scellier, Anirudh Goyal, Jonathan Binas, Thomas Mesnard, and Yoshua
  Bengio.
\newblock Generalization of equilibrium propagation to vector field dynamics,
  2018.

\bibitem{BP_without_weight_transport}
J.F. Kolen and J.B. Pollack.
\newblock Backpropagation without weight transport.
\newblock In {\em Proceedings of 1994 IEEE International Conference on Neural
  Networks (ICNN'94)}, volume~3, pages 1375--1380 vol.3, 1994.

\bibitem{linsker1988self}
Ralph Linsker.
\newblock Self-organization in a perceptual network.
\newblock {\em Computer}, 21(3):105--117, 1988.

\bibitem{bell1995information}
Anthony~J Bell and Terrence~J Sejnowski.
\newblock An information-maximization approach to blind separation and blind
  deconvolution.
\newblock {\em Neural computation}, 7(6):1129--1159, 1995.

\bibitem{HubelReceptive}
D.H. Hubel and T.N. Wiesel.
\newblock Receptive fields of single neurones in the cat's striate cortex.
\newblock {\em The Journal of physiology}, 148:574--591, 1959.

\bibitem{bell1997independent}
Anthony~J Bell and Terrence~J Sejnowski.
\newblock The “independent components” of natural scenes are edge filters.
\newblock {\em Vision research}, 37(23):3327--3338, 1997.

\bibitem{hjelm2018learning}
R~Devon Hjelm, Alex Fedorov, Samuel Lavoie-Marchildon, Karan Grewal, Phil
  Bachman, Adam Trischler, and Yoshua Bengio.
\newblock Learning deep representations by mutual information estimation and
  maximization.
\newblock {\em International Conference on Learning Representations (ICLR)},
  2019.

\bibitem{becker1992self}
Suzanna Becker and Geoffrey~E Hinton.
\newblock Self-organizing neural network that discovers surfaces in random-dot
  stereograms.
\newblock {\em Nature}, 355(6356):161--163, 1992.

\bibitem{erdogan2022icassp}
Alper~T Erdogan.
\newblock An information maximization based blind source separation approach
  for dependent and independent sources.
\newblock In {\em ICASSP 2022 - 2022 IEEE International Conference on
  Acoustics, Speech and Signal Processing (ICASSP)}, pages 4378--4382, 2022.

\bibitem{ozsoy2022selfsupervised}
Serdar Ozsoy, Shadi Hamdan, Sercan~O Arik, Deniz Yuret, and Alper~Tunga
  Erdogan.
\newblock Self-supervised learning with an information maximization criterion.
\newblock In Alice~H. Oh, Alekh Agarwal, Danielle Belgrave, and Kyunghyun Cho,
  editors, {\em Advances in Neural Information Processing Systems}, 2022.

\bibitem{oja1982simplified}
Erkki Oja.
\newblock Simplified neuron model as a principal component analyzer.
\newblock {\em Journal of mathematical biology}, 15:267--273, 1982.

\bibitem{lipshutz2021biologically}
David Lipshutz, Yanis Bahroun, Siavash Golkar, Anirvan~M Sengupta, and Dmitri~B
  Chklovskii.
\newblock A biologically plausible neural network for multichannel canonical
  correlation analysis.
\newblock {\em Neural Computation}, 33(9):2309--2352, 2021.

\bibitem{tatli2021tsp}
Gokcan Tatli and Alper~T Erdogan.
\newblock Polytopic matrix factorization: Determinant maximization based
  criterion and identifiability.
\newblock {\em IEEE Transactions on Signal Processing}, 69:5431--5447, 2021.

\bibitem{plumbley2003algorithms}
Mark~D Plumbley.
\newblock Algorithms for nonnegative independent component analysis.
\newblock {\em IEEE Transactions on Neural Networks}, 14(3):534--543, 2003.

\bibitem{pehlevan2017blind}
Cengiz Pehlevan, Sreyas Mohan, and Dmitri~B Chklovskii.
\newblock Blind nonnegative source separation using biological neural networks.
\newblock {\em Neural computation}, 29(11):2925--2954, 2017.

\bibitem{whittington2023disentanglement}
James C.~R. Whittington, Will Dorrell, Surya Ganguli, and Timothy Behrens.
\newblock Disentanglement with biological constraints: A theory of functional
  cell types.
\newblock In {\em The Eleventh International Conference on Learning
  Representations}, 2023.

\bibitem{lubke1996frequency}
Joachim L{\"u}bke, Henry Markram, Michael Frotscher, and Bert Sakmann.
\newblock Frequency and dendritic distribution of autapses established by layer
  5 pyramidal neurons in the developing rat neocortex: comparison with synaptic
  innervation of adjacent neurons of the same class.
\newblock {\em Journal of Neuroscience}, 16(10):3209--3218, 1996.

\bibitem{o1996biologically}
Randall~C O'Reilly.
\newblock Biologically plausible error-driven learning using local activation
  differences: The generalized recirculation algorithm.
\newblock {\em Neural computation}, 8(5):895--938, 1996.

\bibitem{PierreContrastive}
Pierre Baldi and Fernando Pineda.
\newblock Contrastive learning and neural oscillations.
\newblock {\em Neural Computation}, 3(4):526--545, 1991.

\bibitem{ketz2013theta}
Nicholas Ketz, Srinimisha~G Morkonda, and Randall~C O'Reilly.
\newblock Theta coordinated error-driven learning in the hippocampus.
\newblock {\em PLoS computational biology}, 9(6):e1003067, 2013.

\bibitem{Fellsynchronization}
Juergen Fell and Nikolai Axmacher.
\newblock The role of phase synchronization in memory processes.
\newblock {\em Nature reviews. Neuroscience}, 12:105--18, 02 2011.

\bibitem{EngelDynamic}
Andreas Engel, Pascal Fries, and Wolf Singer.
\newblock Dynamic predictions: oscillations and synchrony in top-down
  processing.
\newblock {\em Nature reviews. Neuroscience}, 2:704--716, 11 2001.

\bibitem{lecunMNIST}
Yann LeCun and Corinna Cortes.
\newblock {MNIST} handwritten digit database.
\newblock http://yann.lecun.com/exdb/mnist/, 2010.

\bibitem{xiao2017fashionmnist}
Han Xiao, Kashif Rasul, and Roland Vollgraf.
\newblock Fashion-mnist: a novel image dataset for benchmarking machine
  learning algorithms, 2017.

\bibitem{krizhevsky2009learning}
Alex Krizhevsky, Geoffrey Hinton, et~al.
\newblock Learning multiple layers of features from tiny images, 2009.

\bibitem{millidge2023backpropagation}
Beren Millidge, Yuhang Song, Tommaso Salvatori, Thomas Lukasiewicz, and Rafal
  Bogacz.
\newblock Backpropagation at the infinitesimal inference limit of energy-based
  models: Unifying predictive coding, equilibrium propagation, and contrastive
  hebbian learning.
\newblock In {\em The Eleventh International Conference on Learning
  Representations}, 2023.

\bibitem{zhouyin:21}
{Zhanghao Zhouyin} and {Ding Liu}.
\newblock Understanding neural networks with logarithm determinant entropy
  estimator.
\newblock {\em arXiv preprint arXiv:1401.3420}, 2021.

\bibitem{kailath2000linear}
Thomas Kailath, Ali~H Sayed, and Babak Hassibi.
\newblock {\em Linear estimation}.
\newblock Prentice-Hall information and system sciences series. Prentice Hall,
  2000.

\bibitem{fu:2016}
Xiao Fu, Kejun Huang, Bo~Yang, Wing-Kin Ma, and Nicholas~D Sidiropoulos.
\newblock Robust volume minimization-based matrix factorization for remote
  sensing and document clustering.
\newblock {\em IEEE Transactions on Signal Processing}, 64(23):6254--6268,
  2016.

\bibitem{brondsted2012introduction}
Arne Brondsted.
\newblock {\em An Introduction to Convex Polytopes}, volume~90.
\newblock Springer Science \& Business Media, 2012.

\bibitem{donoho2006most}
David~L Donoho.
\newblock For most large underdetermined systems of equations, the minimal
  $\ell_1$-norm near-solution approximates the sparsest near-solution.
\newblock {\em Communications on Pure and Applied Mathematics}, 59(7):907--934,
  July 2006.

\bibitem{elad2010sparse}
Michael Elad.
\newblock {\em Sparse and Redundant Representations: From Theory to
  Applications in Signal and Image Processing}.
\newblock Springer Science \& Business Media, 2010.

\bibitem{duchi2008efficient}
John Duchi, Shai Shalev-Shwartz, Yoram Singer, and Tushar Chandra.
\newblock Efficient projections onto the $\ell_1$-ball for learning in high
  dimensions.
\newblock In {\em Proceedings of the 25th International Conference on Machine
  learning}, pages 272--279, July 2008.

\bibitem{babatas2018algorithmic}
Eren Babatas and Alper~T Erdogan.
\newblock An algorithmic framework for sparse bounded component analysis.
\newblock {\em IEEE Transactions on Signal Processing}, 66(19):5194--5205,
  August 2018.

\bibitem{studer2014democratic}
Christoph Studer, Tom Goldstein, Wotao Yin, and Richard~G Baraniuk.
\newblock Democratic representations.
\newblock {\em arXiv preprint arXiv:1401.3420}, 2014.

\bibitem{erdogan2013class}
Alper~T Erdogan.
\newblock A class of bounded component analysis algorithms for the separation
  of both independent and dependent sources.
\newblock {\em IEEE Transactions on Signal Processing}, 61(22):5730--5743,
  August 2013.

\bibitem{inan2014convolutive}
Huseyin~A Inan and Alper~T Erdogan.
\newblock A convolutive bounded component analysis framework for potentially
  nonstationary independent and/or dependent sources.
\newblock {\em IEEE Transactions on Signal Processing}, 63(1):18--30, November
  2014.

\bibitem{bozkurt2022biologicallyplausible}
Bariscan Bozkurt, Cengiz Pehlevan, and Alper~Tunga Erdogan.
\newblock Biologically-plausible determinant maximization neural networks for
  blind separation of correlated sources.
\newblock In Alice~H. Oh, Alekh Agarwal, Danielle Belgrave, and Kyunghyun Cho,
  editors, {\em Advances in Neural Information Processing Systems}, 2022.

\bibitem{rozell:2008}
Christopher~J Rozell, Don~H Johnson, Richard~G Baraniuk, and Bruno~A Olshausen.
\newblock Sparse coding via thresholding and local competition in neural
  circuits.
\newblock {\em Neural computation}, 20(10):2526--2563, 2008.

\bibitem{boyd2004convex}
Stephen Boyd and Lieven Vandenberghe.
\newblock {\em Convex optimization}.
\newblock Cambridge university press, 2004.

\bibitem{parikh2014proximal}
Neal Parikh, Stephen Boyd, et~al.
\newblock Proximal algorithms.
\newblock {\em Foundations and trends{\textregistered} in Optimization},
  1(3):127--239, 2014.

\bibitem{paszke2017automatic}
Adam Paszke, Sam Gross, Francisco Massa, Adam Lerer, James Bradbury, Gregory
  Chanan, Trevor Killeen, Zeming Lin, Natalia Gimelshein, Luca Antiga, Alban
  Desmaison, Andreas Kopf, Edward Yang, Zachary DeVito, Martin Raison, Alykhan
  Tejani, Sasank Chilamkurthy, Benoit Steiner, Lu~Fang, Junjie Bai, and Soumith
  Chintala.
\newblock Pytorch: An imperative style, high-performance deep learning library.
\newblock In {\em Advances in Neural Information Processing Systems 32}, pages
  8024--8035. Curran Associates, Inc., 2019.

\end{thebibliography}
